\documentclass[final,3p]{elsarticle}

\usepackage{lineno,hyperref}
\modulolinenumbers[10]

\usepackage{ifpdf}
\usepackage{ifthen}

\usepackage{array}
\usepackage{tabularx}
\usepackage{multirow}
\usepackage{booktabs}

\usepackage{graphicx}
\usepackage{grffile}
\usepackage{color}
\usepackage[dvipsnames]{xcolor}
\usepackage{float}
\usepackage{caption}
\usepackage{subcaption}
\usepackage{rotating}

\usepackage{tablefootnote}

\usepackage{pgf,pgfplots,pgfplotstable}
\pgfplotsset{compat=newest}
\pgfplotsset{
xlabel near ticks,
ylabel near ticks,
label style={font=\footnotesize},
tick label style={font=\footnotesize},
legend style={font=\scriptsize},
xticklabel style={/pgf/number format/set thousands separator={\,}},
yticklabel style={/pgf/number format/set thousands separator={\,}},
tick scale binop=\times,
try min ticks=6,
legend pos=outer north east
}
\usepgfplotslibrary{external} 
\usepgfplotslibrary{units}
\usepgfplotslibrary{patchplots}
\usepgfplotslibrary{groupplots}
\usepgfplotslibrary{fillbetween}

\usepackage{tikz,tkz-base,xstring}
\usetikzlibrary{shapes,decorations,shadows}
\usetikzlibrary{decorations.pathreplacing}
\usetikzlibrary{decorations.pathmorphing}
\usetikzlibrary{decorations.shapes}
\usetikzlibrary{decorations.text}
\usetikzlibrary{decorations.footprints}
\usetikzlibrary{decorations.fractals}
\usetikzlibrary{fadings}
\usetikzlibrary{patterns}
\usetikzlibrary{calc}
\usetikzlibrary{shapes.geometric}
\usetikzlibrary{shapes.gates.logic.IEC}
\usetikzlibrary{shapes.gates.logic.US}
\usetikzlibrary{fit,chains}
\usetikzlibrary{positioning}
\usepgflibrary{shapes}
\usetikzlibrary{scopes}
\usetikzlibrary{arrows}
\usetikzlibrary{arrows.meta}
\usetikzlibrary{backgrounds}
\usetikzlibrary{intersections}
\usetikzlibrary{matrix}
\usetikzlibrary{plotmarks}
\usetikzlibrary{pgfplots.units}
\usetikzlibrary{trees}
\usetikzlibrary{hobby}

\newlength\figureheight
\newlength\figurewidth
\usepgfplotslibrary{units}

\pgfdeclaredecoration{penciline}{initial}{
\state{initial}[width=+\pgfdecoratedinputsegmentremainingdistance,auto corner on length=1mm]{
\pgfpathcurveto
{
\pgfqpoint{\pgfdecoratedinputsegmentremainingdistance}{\pgfdecorationsegmentamplitude}}{
\pgfpointadd{\pgfqpoint{\pgfdecoratedinputsegmentremainingdistance}{0pt}}{
\pgfqpoint{-\pgfdecorationsegmentaspect\pgfdecoratedinputsegmentremainingdistance}{\pgfmathresult\pgfdecorationsegmentamplitude}}}
{
\pgfpointadd{\pgfpointdecoratedinputsegmentlast}{\pgfpoint{1pt}{1pt}}}
}
\state{final}{}
}

\usepackage{stanli}

\usepackage[all]{xy}

\usepackage{amsmath,amssymb,amsfonts,amsthm,bm}
\usepackage{textcomp,wasysym,eurosym,stmaryrd}
\usepackage{mathrsfs}
\usepackage{mathtools}
\usepackage{pifont}
\usepackage{upgreek}
\usepackage{centernot}
\usepackage{algorithm,algorithmic}

\usepackage{xspace}
\usepackage{enumitem}
\setlist[itemize]{label={--}}

\newcommand{\fb}{\boldsymbol{f}}

\newcommand{\hb}{\boldsymbol{h}}

\newcommand{\qb}{\boldsymbol{q}}

\newcommand{\ssb}{\boldsymbol{s}}

\newcommand{\xb}{\boldsymbol{x}}

\newcommand{\epsilonb}{\boldsymbol{\varepsilon}}

\newcommand{\ellb}{\boldsymbol{\ell}}

\newcommand{\Eb}{\boldsymbol{E}}

\newcommand{\Hb}{\boldsymbol{H}}

\newcommand{\Lb}{\boldsymbol{L}}

\newcommand{\Qb}{\boldsymbol{Q}}

\newcommand{\Sb}{\boldsymbol{S}}

\newcommand{\Ub}{\boldsymbol{U}}

\newcommand{\Xb}{\boldsymbol{X}}

\newcommand{\Gc}{\mathcal{G}}
\newcommand{\Hc}{\mathcal{H}}

\newcommand{\Kc}{\mathcal{K}}

\newcommand{\Mc}{\mathcal{M}}
\newcommand{\Nc}{\mathcal{N}}

\newcommand{\Uc}{\mathcal{U}}

\newcommand{\Gcb}{\boldsymbol{\Gc}}

\newcommand{\Mcb}{\boldsymbol{\Mc}}
\newcommand{\Ncb}{\boldsymbol{\Nc}}

\newcommand{\Ucb}{\boldsymbol{\Uc}}

\newcommand{\Ebb}{\mathbb{E}}

\newcommand{\Mbb}{\mathbb{M}}

\newcommand{\Rbb}{\mathbb{R}}

\DeclareFontFamily{U}{bbold}{}
\DeclareFontShape{U}{bbold}{m}{n}{<-5.5> bbold5 <5.5-7.5> bbold7 <7.5-> bbold10}{}

\newcommand{\abs}[1]{\lvert#1\rvert}

\newcommand{\norm}[1]{\lVert#1\rVert}

\newcommand{\set}[1]{\{#1\}}

\DeclareMathOperator{\gradd}{grad\kern-.5em{grad}}

\newcommand{\interval}[4]{\mathopen{#1}#2 \mathclose{}\mathpunct{},#3 \mathclose{#4}}
\newcommand{\intervalcc}[2]{\interval{[}{#1}{#2}{]}}

\newcommand{\intervaloo}[2]{\interval{]}{#1}{#2}{[}}

\renewcommand{\(}{\left(}
\renewcommand{\)}{\right)}

\renewcommand{\leq}{\leqslant}
\let\oldtimes\times
\renewcommand{\times}{\!\oldtimes\!}

\newcommand{\ie}{\emph{i.e.}\xspace}
\newcommand{\eg}{\emph{e.g.}\xspace}

\journal{Computer Methods in Applied Mechanics and Engineering}

\begin{document}

\begin{frontmatter}

\title{A robust solution of a statistical inverse problem in multiscale computational mechanics using an artificial neural network}

\author[label1]{Florent Pled\corref{cor1}}
\ead{florent.pled@univ-eiffel.fr}
\author[label1]{Christophe Desceliers}
\ead{christophe.desceliers@univ-eiffel.fr}
\author[label1]{Tianyu Zhang}
\ead{tianyu.zhang@univ-eiffel.fr}
\cortext[cor1]{Corresponding author}

\address[label1]{Univ Gustave Eiffel, MSME UMR 8208, F-77454, Marne-la-Vall\'ee, France}

\begin{abstract}
This work addresses the inverse identification of apparent elastic properties of random heterogeneous materials using machine learning based on artificial neural networks. The proposed neural network-based identification method requires the construction of a database from which an artificial neural network can be trained to learn the nonlinear relationship between the hyperparameters of a prior stochastic model of the random compliance field and some relevant quantities of interest of an \emph{ad hoc} multiscale computational model. An initial database made up with input and target data is first generated from the computational model, from which a processed database is deduced by conditioning the input data with respect to the target data using the nonparametric statistics. Two- and three-layer feedforward artificial neural networks are then trained from each of the initial and processed databases to construct an algebraic representation of the nonlinear mapping between the hyperparameters (network outputs) and the quantities of interest (network inputs). The performances of the trained artificial neural networks are analyzed in terms of mean squared error, linear regression fit and probability distribution between network outputs and targets for both databases. An \emph{ad hoc} probabilistic model of the input random vector is finally proposed in order to take into account uncertainties on the network input and to perform a robustness analysis of the network output with respect to the input uncertainties level. The capability of the proposed neural network-based identification method to efficiently solve the underlying statistical inverse problem is illustrated through two numerical examples developed within the framework of 2D plane stress linear elasticity, namely a first validation example on synthetic data obtained through computational simulations and a second application example on real experimental data obtained through a physical experiment monitored by digital image correlation on a real heterogeneous biological material (beef cortical bone).
\end{abstract}




\end{frontmatter}


\section{Introduction}
\label{sec:introduction}

\subsection{\emph{Multiscale statistical inverse problem}}

The present paper concerns the mechanical characterization and identification of elastic properties for heterogeneous materials with a complex microstructure that may be considered as a random linear elastic medium.
The high complexity level and multiphase nature of such microstructures do not allow for a proper description and modeling of the morphological and mechanical properties of their constituents at microscale.
For such kind of materials, such as rock-like materials, concretes and cementitious materials, natural or synthetic composites and biological materials, a stochastic modeling of the apparent elastic properties of the microstructure can be constructed at a given mesoscale corresponding to the scale of the spatial correlation length of the microstructure.
The uncertainties on the mechanical properties of such random heterogeneous materials are modeled by a non-Gaussian random elasticity (or compliance) field \cite{Soi06,Soi08a} whose prior stochastic model is constructed within the framework of probability theory and information theory, that are among the most robust and well-established theories based on a solid mathematical background for several centuries.
Such stochastic models of uncertainties are classically implemented into deterministic computational models yielding stochastic computational models that require parallel and high-performance computing (HPC) for propagating the uncertainties in high stochastic dimension.
A major and still open challenge concerns the statistical inverse identification of stochastic models in using available data coming from either forward numerical simulations performed with computational models or experimental measurements obtained by means of physical tests.
The statistical inverse problem under consideration consists in finding the values of the hyperparameters of a prior stochastic model of the random compliance field corresponding to highly probable values for some given observed quantities of interest of an \emph{ad hoc} computational model.
Such a statistical inverse problem has been formulated in \cite{Ngu15,Zha20} as a multi-objective optimization problem and solved by using a global optimization algorithm (genetic algorithm) in \cite{Ngu15} and a fixed-point iterative algorithm in \cite{Zha20}, both requiring many calls to the computational model, which may be time consuming in practice especially for real-time applications such as in biomechanics.
During the last decade, many identification methodologies and numerical developments have been proposed for addressing the problem related to the statistical inverse identification of stochastic models of the random elasticity (or compliance) field in low or high stochastic dimension at macroscale and/or mesoscale for complex microstructures modeled by random heterogeneous isotropic or anisotropic linear elastic media \cite{Des06,Gha06,Des07,Mar07,Arn08,Das08,Das09b,Des09,Gui09,Ma09,Mar09a,Arn10,Das10,Ta10,Soi10,Soi11a,Cot11,Des12,Per12,Clo13,Ngu15,Zha20}.
The proposed identification methods require solving a statistical inverse problem classically formulated as a stochastic optimization problem, which may be computationally expensive even using modern computing hardware with powerful multicores processors and tricky to implement into commercial softwares dedicated to real-time identification or digital twin applications.
In addition, the data required for performing the identification has to be stored in memory on the computing device and always accessible, which may be difficult to manage depending on the available memory storage capacity.

\subsection{\emph{Improvements of the multiscale identification method and novelty of the paper}}

In the present work, we propose an appealing alternative for addressing the aforementioned drawbacks and solving the statistical inverse problem related to the identification of an \emph{ad hoc} stochastic model of the random compliance field within the framework of 2D plane stress linear elasticity theory by using Machine Learning (ML) approaches based on Artificial Neural Networks (ANNs) \cite{Hay94,Hag96,Dem14}.
ANNs are among the most widely used algorithms in supervised ML techniques to construct and train predictive models that map inputs to outputs for feature or pattern recognition/detection/selection/extraction, clustering, classification, compression/filtering, fitting/regression, identification and/or prediction/forecasting purposes.
ML algorithms, such as ANNs, use advanced computational methods to learn information directly from data (without relying on any analytical or numerical model describing the input-output relationship).
Since the training algorithms based on gradient computations for the design of ANNs are well adapted to parallelization, the training can be performed in parallel and distributed across multicores central processing units (CPUs), graphics processing units (GPUs), or even scaled up to clusters of computers and clouds with multiple CPUs and/or GPUs for a better computational efficiency.
With the recent development in HPC and massively parallel processing systems, modern GPUs are more efficient than general-purpose CPUs for manipulating a huge amount of data due to their highly parallel structure.
Consequently, they turn out to be particularly well adapted to machine learning for accelerating the network training process for very large datasets (often referred to as big data).
Lastly, the use of ML algorithms has surged in popularity over the last years primarily due to their high accuracy, their short training time (thanks to the use of GPUs) and the storage, accessibility and processing of lots of data.

In the present paper, the statistical inverse problem is formulated as a function approximation problem and solved by using an ANN trained from a numerical database constructed from the computational model.

\subsection{\emph{Methodology proposed in the paper}}

The proposed neural network-based identification method consists in the following steps.
\begin{enumerate}
\item
  A (deterministic) forward computational model is constructed and parameterized by the compliance field at mesoscale of the material.
  The quantities of interest that are computed by the forward computational model are gathered into the deterministic vector $\qb$.
\item
  The uncertainty quantification on the values of $\qb$ is carried out by modeling the quantities of interest as a random vector $\Qb$.
  The random quantities of interest are defined as the outputs of the stochastic forward computational model that is constructed by introducing the prior stochastic model of the random compliance field.
  Let $\hb$ be the vector of the hyperparameters of this prior stochastic model that has to be identified by the statistical inverse problem given an observation of $\Qb$.
\item
  Since the value of $\hb$ is uncertain, then the hyperparameters are modeled as a random vector $\Hb$. 
\item
  For each of the $N_d$ independent realizations $\hb^{(1)},\dots,\hb^{(N_d)}$ of $\Hb$, the stochastic forward computational model is used for computing one realization of $\Qb$, yielding $N_d$ independent realizations $\qb^{(1)},\dots,\qb^{(N_d)}$ of the quantities of interest (see Figure~\ref{fig:flowchart_generate_database_initial}). 
  An initial database is then obtained for which the $i$-th element is the vector $\xb^{(i)} = (\qb^{(i)}, \hb^{(i)})$.
\item
  It should be noted that the mapping between $\Qb$ and $\Hb$ is random by construction.
  As a consequence, the supervised training of an ANN with the initial database cannot be efficient since a trained ANN is a deterministic mapping between its inputs and outputs.
  This is the reason why the initial database is then processed in substituting $\Qb$ by another network input random vector $\widetilde{\Qb}$ such that the mapping between $\widetilde{\Qb}$ and $\Hb$ is (almost) deterministic.
  It would then make it possible to efficiently train an artificial neural network.
  In this paper, it is proposed to obtain the processed database by conditioning the initial database.
  The $N_d$ vectors $\qb^{(1)},\dots,\qb^{(N_d)}$ are then replaced by the $N_d$ vectors $\tilde{\qb}^{(1)},\dots,\tilde{\qb}^{(N_d)}$, respectively (see Figure~\ref{fig:flowchart_generate_database_processed}).
  Additional details on the construction of vectors $\tilde{\qb}^{(1)},\ldots,\tilde{\qb}^{(N_d)}$ are given later in the paper.
  Such a data conditioning is performed by using classical kernel smoothing techniques in nonparametric statistics \cite{Bow97,Hor12,Giv13,Sco15,Soi17a} for computing conditional mathematical expectations.
\item
  A multilayer ANN can then be designed to learn the nonlinear relationship between the hyperparameters (network outputs) $\hb^{(1)},\dots,\hb^{(N_d)}$ and the quantities of interest (network inputs) $\tilde{\qb}^{(1)},\dots,\tilde{\qb}^{(N_d)}$ and trained using the processed database in a potentially computationally expensive offline phase (preliminary learning phase).
  The (best) trained ANN can then be used to identify the value $\hb^{\ast}$ of the output vector $\hb$ of hyperparameters for a given observed input vector $\qb^{\text{obs}}$ of quantities of interest in a computationally cheap online phase (real-time computing phase) (see Figure~\ref{fig:flowchart_compute_solution}).
\item
  Finally, the robustness of the proposed identification method can be further assessed by considering the observed vector of quantities of interest as an input random vector for which the probabilistic model can be constructed by using the maximum entropy (MaxEnt) principle \cite{Jay57a,Jay57b,Sob90,Kap92,Jum00,Jay03,Cov06,Soi17a}, thus allowing the experimental errors on the observed quantities of interest (induced by the measurement noise and/or the variabilities in the experimental configuration) to be taken into account.
  %
  %
\end{enumerate}
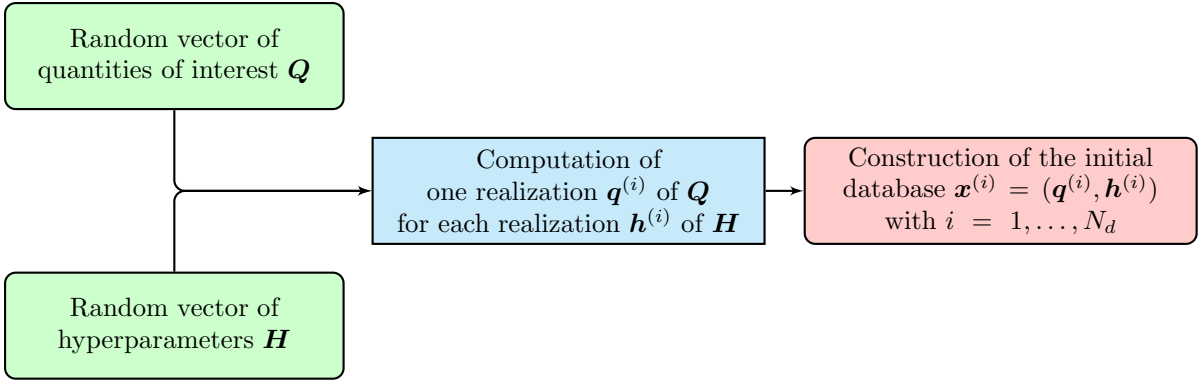
\begin{figure}[h!]
\centering
\tikzsetnextfilename{flowchart_generate_database_initial}
\tikzstyle{line} = [draw, -latex', thick, rounded corners=4pt]
\tikzstyle{block} = [rectangle, draw, thick, fill=cyan!20, text width=14em, text centered, node distance=0.5cm, minimum height=4em]
\tikzstyle{inputcloud} = [rectangle, draw, thick, fill=green!20, text width=12em, text centered, rounded corners, node distance=0.5cm, minimum height=4em]
\tikzstyle{outputcloud} = [rectangle, draw, thick, fill=red!20, text width=14em, text centered, rounded corners, node distance=0.5cm, minimum height=4em]

\begin{tikzpicture}
\centering
\node [block] (2) {Computation of\\ one realization $\qb^{(i)}$ of $\Qb$\\ for each realization $\hb^{(i)}$ of $\Hb$};
\node [inputcloud, above left= of 2] (0) {Random vector of\\ quantities of interest $\Qb$};
\node [inputcloud, below left= of 2] (1) {Random vector of\\ hyperparameters $\Hb$};
\node [outputcloud, right= of 2] (3) {Construction of the initial\\ database $\xb^{(i)} = (\qb^{(i)},\hb^{(i)})$\\ with $i=1,\dots,N_d$};

\path [line] (0) |- (2);
\path [line] (1) |- (2);
\path [line] (2) -- (3);
\end{tikzpicture}
\caption{Flowchart for generating the initial database}\label{fig:flowchart_generate_database_initial}
\end{figure}

\begin{figure}[h!]
\centering
\tikzsetnextfilename{flowchart_generate_database_processed}
\tikzstyle{line} = [draw, -latex', thick, rounded corners=4pt]
\tikzstyle{block} = [rectangle, draw, thick, fill=cyan!20, text width=14em, text centered, node distance=0.5cm, minimum height=4em]
\tikzstyle{inputcloud} = [rectangle, draw, thick, fill=green!20, text width=12em, text centered, rounded corners, node distance=0.5cm, minimum height=4em]
\tikzstyle{outputcloud} = [rectangle, draw, thick, fill=red!20, text width=14em, text centered, rounded corners, node distance=0.5cm, minimum height=4em]

\begin{tikzpicture}
\centering
\node [inputcloud] (0) {Initial database\\ $\xb^{(i)} = (\qb^{(i)},\hb^{(i)})$\\ with $i=1,\dots,N_d$};
\node [block, right= of 0] (1) {Computation of the conditional\\ mathematical expectation\\ $\tilde{q}_k^{(i)}$ of $Q_k$ given $\Hb = \hb^{(i)}$\\ with $k=1,\dots,n$};
\node [outputcloud, right= of 1] (2) {Construction of the processed\\ database $\tilde{\xb}^{(i)} = (\tilde{\qb}^{(i)},\hb^{(i)})$\\ with $\tilde{\qb}^{(i)} = (\tilde{q}^{(i)}_1,\dots,\tilde{q}^{(i)}_n)$ and $i=1,\dots,N_d$};

\path [line] (0) -- (1);
\path [line] (1) -- (2);
\end{tikzpicture}
\caption{Flowchart for generating the processed database}\label{fig:flowchart_generate_database_processed}
\end{figure}
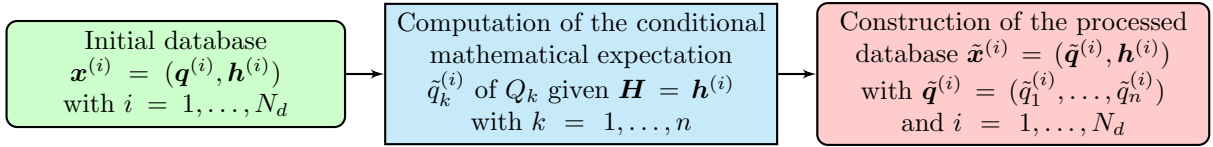
\begin{figure}[h!]
\centering
\tikzsetnextfilename{flowchart_compute_solution}
\tikzstyle{line} = [draw, -latex', thick, rounded corners=4pt]
\tikzstyle{block} = [rectangle, draw, thick, fill=cyan!20, text width=14em, text centered, node distance=0.5cm, minimum height=4em]
\tikzstyle{inputcloud} = [rectangle, draw, thick, fill=green!20, text width=14em, text centered, rounded corners, node distance=0.5cm, minimum height=4em]
\tikzstyle{outputcloud} = [rectangle, draw, thick, fill=red!20, text width=14em, text centered, rounded corners, node distance=0.5cm, minimum height=4em]

\begin{tikzpicture}
\centering
\node [inputcloud] (0) {Observed vector of\\ quantities of interest $\qb^{\text{obs}}$\\ as input of the trained ANN};
\node [block, right= of 0] (1) {ANN trained using\\ the processed database $\tilde{\xb}^{(i)} = (\tilde{\qb}^{(i)},\hb^{(i)})$\\ with $i=1,\dots,N_d$};
\node [outputcloud, right= of 1] (2) {Computation of\\ the solution vector $\hb^{\ast}$ of\\ the statistical inverse problem\\ as output of the trained ANN};

\path [line] (0) -- (1);
\path [line] (1) -- (2);
\end{tikzpicture}
\caption{Flowchart for computing the solution of the statistical inverse problem}\label{fig:flowchart_compute_solution}
\end{figure}
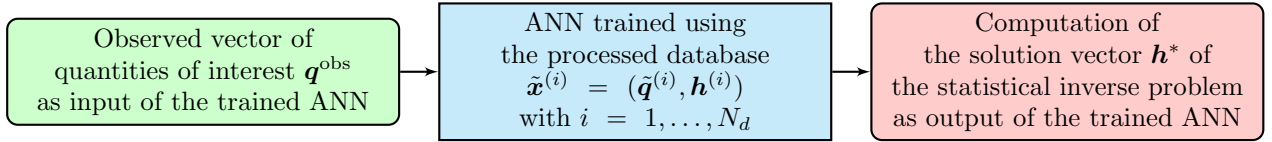
It should be pointed out that such an identification procedure can be performed directly with no call to the computational model (during the online computing phase), this latter being used only for the generation of the database required to design the multilayer ANN (during the offline learning phase).
As a consequence, the proposed neural network-based identification strategy is computationally cheap, easy to implement and use.

\subsection{\emph{Outline of the paper}}

The remainder of the paper is structured as follows.
Section~\ref{sec:definition_HFCMM} presents the forward computational models, namely the High Fidelity Computational Mechanical Model and the Homogenization Computational Model, introduced within the framework of linear elasticity theory and used to 
compute relevant quantities of interest for the considered inverse identification problem to be solved.
In Section~\ref{sec:stochastic_model}, the prior stochastic model of the random compliance field that characterizes the apparent elastic properties of the random heterogeneous linear elastic medium under consideration is described and the associated hyperparameters to be identified are introduced.
Section~\ref{sec:construction_database} is devoted to the construction of the initial database containing the network input and target data.
Then, the statistical inverse problem is introduced and formulated as a function approximation problem in Section~\ref{sec:inverse_problem}.
Section~\ref{sec:conditioning_database} is devoted to the construction of the processed database obtained by conditioning the initial database and allowing for a robust identification of the solution of the statistical inverse problem.
A statistical analysis of the initial and processed databases is then carried out in Section~\ref{sec:sensitivity_analysis_databases} for studying the sensitivity of the network target data with respect to the network input data.
Section~\ref{sec:design_network} deals with the design of the artificial neural network including the neural network architecture, the data partitioning and the training algorithm used to find the best predictive model.
The performances of the multilayer neural networks trained with each of the initial and processed databases are then evaluated in terms of normalized mean squared error, linear regression fit and estimation of probability density function between network outputs and targets.
An \emph{ad hoc} probabilistic model for the input random vector of quantities of interest is presented in Section~\ref{sec:robustness} in order to perform a robustness analysis of the network output with respect to the uncertainties on a given input.
The capability of the proposed neural network-based identification method to efficiently solve the considered statistical inverse problem is shown though two numerical examples presented in Sections~\ref{sec:results_synthetic_data} and \ref{sec:results_real_data}.
The proposed approach implemented within the framework of 2D plane stress linear elasticity is first validated on synthetic data obtained by numerical simulations in Section~\ref{sec:results_synthetic_data} and then applied to real data obtained by means of experimental measurements on a real heterogeneous biological tissue (bovine cortical bone) in Section~\ref{sec:results_real_data}.
Finally, Section~\ref{sec:conclusion} draws some conclusions and suggests potentially beneficial directions for future research works.

\section{Construction of a High Fidelity Computational Mechanical Model and a Homogenization Computational Model}
\label{sec:definition_HFCMM}

\begin{figure}[h!]
\centering
\tikzsetnextfilename{boundary_value_problem}
\begin{tikzpicture}[every node/.style={minimum size=1cm},on grid]
\def \L {5}
\def \l {0.5}

\coordinate (A) at (0,0);
\coordinate (B) at (\L,\L);
\coordinate (C) at (0,\L);
\coordinate (D) at (\L,0);
\draw[black,thick,fill=green!20
] (A) rectangle (B);
\coordinate (lineloadVarA1) at ($ (C)!0!90:(B) $);
\coordinate (lineloadVarB1) at ($ (B)!0!-90:(C) $);
\coordinate (lineloadVarA2) at ($ (C)!{0.75cm}!90:(B) $);
\coordinate (lineloadVarB2) at ($ (B)!{0.75cm}!-90:(C) $);
\draw[green!60!black,force,->] (lineloadVarA2) -- (lineloadVarA1);
\draw[green!60!black,force,->] (lineloadVarB2) -- (lineloadVarB1);
\pgfmathsetmacro{\lineloadIntervalBegin}{\lineloadInterval/2/\scalingParameter}
\pgfmathsetmacro{\lineloadIntervalStep}{\lineloadInterval/2/\scalingParameter*2}
\pgfmathsetmacro{\lineloadIntervalEnd}{1-\lineloadInterval/2/\scalingParameter}
\foreach \i in {\lineloadIntervalBegin,\lineloadIntervalStep,...,\lineloadIntervalEnd}
\draw[green!60!black,force,->] ($(lineloadVarA2)!\i!(lineloadVarB2)$)-- ($(lineloadVarA1)!\i!(lineloadVarB1)$);
\node[green!60!black,above=0.6] at ($(B)!0.5!(C)$) {$\fb$};
\begin{scope}
	\clip (0,-\supportBasicHeight) rectangle (\L,0);
	\draw[hatching](\L+1,0) -- ++(-\L-1,0);
\end{scope}
\foreach \i in {3.25}{
	\foreach \j in {3.25}{
		\coordinate (E) at (\j*\L/5-\l/2,\i*\L/5-\l/2);
		\coordinate (F) at (\j*\L/5+\l/2,\i*\L/5+\l/2);
		\coordinate (G) at (\j*\L/5-\l/2,\i*\L/5+\l/2);
		\coordinate (H) at (\j*\L/5+\l/2,\i*\L/5-\l/2);
		\draw[black,thick,dashed
		] (E) rectangle (F);
	}
}
\node[right=-0.2] at ($(A)!0.07!(B)$) {$\Omega^{\text{macro}}$};
\coordinate (GN) at (\L+0.5,\L+0.5);
\draw[<-,>=latex] ($(C)!0.95!(B)$) to[bend left] (GN) node[right] {$\Gamma_N^{\text{macro}}$};
\coordinate (GD) at (\L+0.5,-0.5);
\draw[<-,>=latex] ($(A)!0.95!(D)$) to[bend right] (GD) node[right] {$\Gamma_D^{\text{macro}}$};
\node[right=0.1] at ($(B)!0.1!(D)$) {$\partial \Omega^{\text{macro}}$};
\node at ($(A)!0.5!(B)$) {$[S^{\text{meso}}]$};

\begin{scope}[
	xshift=\L cm+1cm,
	]
	\coordinate (I) at (\L/5,\L/5);
	\coordinate (J) at (4*\L/5,4*\L/5);
	\coordinate (K) at (\L/5,4*\L/5);
	\coordinate (L) at (4*\L/5,\L/5);
\end{scope}

\draw[black,dashed,thin] (E) to (I);
\draw[black,dashed,thin] (F) to (J);
\draw[black,dashed,thin] (G) to (K);
\draw[black,dashed,thin] (H) to (L);

\begin{scope}[
	xshift=\L cm+1cm,
	]
	\draw[black,thick,fill=red!30,fill opacity=0.85] (I) rectangle (J);
	\node[right=-0.2] at ($(I)!0.12!(J)$) {$\Omega^{\text{meso}}$};
	\node[right=0.1] at ($(J)!0.1!(L)$) {$\partial \Omega^{\text{meso}}$};
	\node at ($(I)!0.5!(J)$) {$[S^{\text{eff}}]$};
\end{scope}

\end{tikzpicture}
\caption{Linear elastic domains $\Omega^{\text{macro}}$ and $\Omega^{\text{meso}}$ for computing the fine-scale displacement fields by the High Fidelity Computational Mechanical Model ($\Mcb^{\text{HFCMM}}$) and the effective compliance matrix by the Computational Homogenization Model ($\Mcb^{\text{EFF}}$)}
\label{fig:boundary_value_problem}
\end{figure}
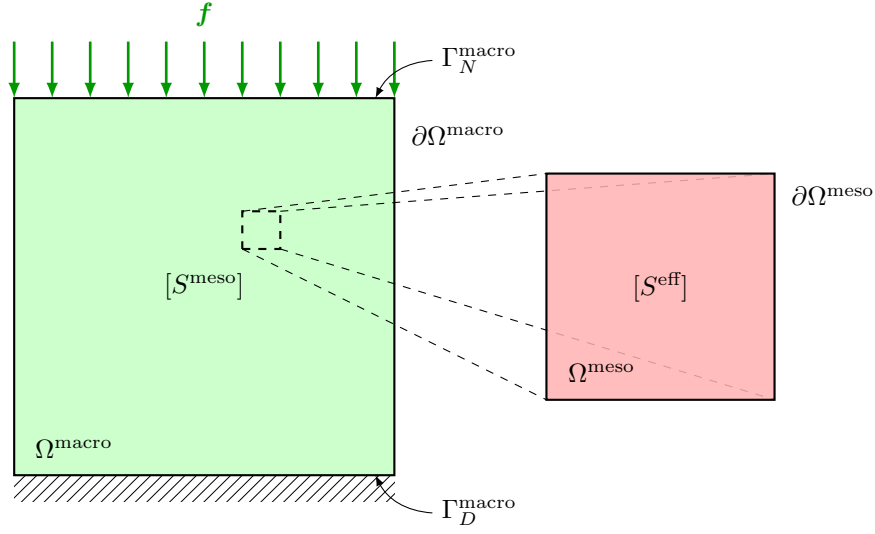

Hereinafter, within the framework of linear elasticity theory, a High Fidelity Computational Mechanical Model (HFCMM) is constructed by using a classical displacement-based Finite Element Method (FEM) \cite{Hug87,Zie05} to compute a fine-scale displacement field of a heterogeneous elastic medium submitted to a given static external loading under the 2D plane stress assumption (see Figure~\ref{fig:boundary_value_problem}). Such an assumption has been introduced only for better representing the experimental configuration for which experimental data are available for the numerical application presented in this paper. Consequently, the fine-scale vector-valued displacement field is only calculated on a 2D open bounded domain $\Omega^{\text{macro}}$ of $\Rbb^2$ that is occupied by a heterogeneous linear elastic medium. In the following, we will consider a 2D square domain $\Omega^{\text{macro}} \subset \Rbb^2$ defined in a fixed Cartesian frame $(O,x_1,x_2)$ of $\Rbb^2$ with macroscopic dimensions $1\times 1$~cm$^2$. A given external line force field $\fb$ is applied on the top part $\Gamma_D^{\text{macro}}$ of the boundary $\partial\Omega^{\text{macro}}$ of $\Omega^{\text{macro}}$, while the right and left parts are both stress-free boundaries and the bottom part $\Gamma_N^{\text{macro}} \subset \partial\Omega^{\text{macro}}$ is assumed to be fixed (see Figure~\ref{fig:boundary_value_problem}). Without loss of generality, we assume no body force field within $\Omega^{\text{macro}}$, the effects of gravity being neglected. Deterministic line force field $\fb$ is uniformly distributed along the (downward vertical) $-x_2$ direction with an intensity of $5$~kN such that $\norm{\fb} = 5~\text{kN/cm} = 5\times 10^{5}$~N/m. The HFCMM is constructed by using the FEM for which the 2D domain $\Omega^{\text{macro}}$ is discretized with a fine structured mesh consisting of $4$-nodes linear quadrangular elements with uniform element size $h = 10~\mu\text{m} = 10^{-5}$~m in each spatial direction. The finite element mesh of domain $\Omega^{\text{macro}}$ then contains $1001 \times 1001 = 1\,002\,001$ nodes and $1\,000 \times 1\,000 = 10^6$ elements, with $2\,000\,000$ unknown degrees of freedom. Within the framework of 2D plane stress linear elasticity theory, the elasticity properties of the heterogeneous linear elastic medium are characterized by a compliance field $[S^{\text{meso}}]$ with values in $\Mbb_3^+(\Rbb)$, where $\Mbb_3^+(\Rbb)$ denotes the set of all the definite-positive symmetric real $(3\times3)$ matrices. For identification purposes, the observed quantities of interest are obtained by postprocessing the kinematics fields that are calculated by the HFCMM on a subdomain $\Omega^{\text{meso}} \subset \Omega^{\text{macro}}$ for which the dimensions $1\times 1$~mm$^2$ do not have to correspond to those of a Representative Volume Element (RVE) of the material since scale separation assumption is not required in all the following. Hence, a first quantity of interest calculated by the HFCMM consists in the spatial dispersion coefficient $\delta^{\epsilonb}$ that quantifies the level of spatial fluctuations of the linearized strain field $\epsilonb$ around its spatial average $\underline{\epsilonb}$ over $\Omega^{\text{meso}}$ and that is defined by
\begin{equation}\label{dispersionstrainfield}
\delta^{\epsilonb} = \dfrac{1}{{\norm{\underline{\epsilonb}}_F}} \(\dfrac{1}{\abs{\Omega^{\text{meso}}}} \int_{\Omega^{\text{meso}}} \norm{\epsilonb(\xb) - \underline{\epsilonb}}_F^2 \, d\xb\)^{1/2} \quad \text{with} \quad \underline{\epsilonb} = \dfrac{1}{\abs{\Omega^{\text{meso}}}} \int_{\Omega^{\text{meso}}} \epsilonb(\xb)\, d\xb,
\end{equation}
where $\abs{\Omega^{\text{meso}}}$ denotes the measure of domain $\Omega^{\text{meso}}$ and $
\Vert \cdot \Vert_F$ denotes the Frobenius norm. The second and third quantities of interest are the two characteristic lengths $\ell^{\epsilonb}_1$ and $\ell^{\epsilonb}_2$ that characterize the spatial fluctuations of $\epsilonb$ around its spatial average $\underline{\epsilonb}$ along the two spatial directions $x_1$ and $x_2$, respectively, and naively computed on domain $\Omega^{\text{meso}}$ in using a usual signal processing method (such as the periodogram method, for instance) although $\ell^{\epsilonb}_1$ and $\ell^{\epsilonb}_2$ should be dependent of spatial position $\xb$ because of the nature of the problem. The interested reader is referred to \cite{Zha19} for the numerical computation of $\ell^{\epsilonb}_1$ and $\ell^{\epsilonb}_2$. Computing the quantities of interest $\delta^{\epsilonb}$, $\ell^{\epsilonb}_1$ and $\ell^{\epsilonb}_2$ by using the HFCMM for any given fine-scale matrix-valued compliance field $[S^\text{meso}]$ allows defining a nonlinear mapping $\Mcb^{\text{HFCMM}}$ defined from $\Mbb_3^+(\Rbb)$ into $\(\Rbb^+\)^3$ such that
\begin{equation}\label{mappingHFCMM}
\(\delta^{\epsilonb},\,\ell^{\epsilonb}_1,\,\ell^{\epsilonb}_2\) = \Mcb^{\text{HFCMM}}([S^{\text{meso}}]).
\end{equation}
It should be noted that when the length scale of the heterogeneities is very small with respect to the dimensions of domain $\Omega^{\text{macro}}$, then the dimension of such a computational model can be very high and the computational cost incurred by such HFCMM can become prohibitive in practical applications. In standard practice, the usual numerical approach then consists in computing the coarse-scale (macroscale) displacement field instead of the fine-scale (mesoscale) displacement field, for instance by calculating the $(3\times3)$ effective compliance matrix $[S^{\text{eff}}]$ in 2D plane stress linear elasticity at a larger scale using an \emph{ad hoc} computational homogenization method. Among the existing computational homogenization methods (see for instance \cite{Nem93,Bor01,Tor02,Zao02,Bou04} and the references therein), the static uniform boundary conditions (SUBC) homogenization approach, in which Neumann boundary conditions (homogeneous stresses) are applied on the whole boundary of $\Omega^{\text{meso}}$, is preferred to the kinematic uniform boundary conditions (KUBC) method, in which Dirichlet boundary conditions (homogeneous strains) are applied on the whole boundary of $\Omega^{\text{meso}}$. Nevertheless, such a coarse-scale displacement field approach avoids resorting to the HFCMM since the coarse-scale displacement field does not bring a sufficient level of granularity in the information for performing the inverse identification of the material properties at the finer scale. Previous research works (see \cite{Ngu15,Ngu16,Zha20}) have been carried out to avoid the use of a HFCMM but the identification methodology requires solving a challenging multi-objective optimization problem involving several disconnected boundary value problems at the fine scale set on domains for which the dimensions are not too large with respect to the characteristic size of the heterogeneities at microscale. A major drawback of this identification method is that a multi-objective optimization problem has to be solved for each experimental data, which severely limits its use in practical applications. Also, the computational cost of this multi-objective optimization problem is non negligible with the current available computer resources and remains high whatever the optimization algorithm considered such as the genetic algorithm used in \cite{Ngu15,Ngu16} or the fixed-point iterative algorithm introduced in \cite{Zha20} for a better computational efficiency. Consequently, such an approach cannot be used for real-time or digital twin applications for instance, and currently requires performing parallel and distributed computations across powerful multicores CPUs to preserve an affordable computational cost. This is the reason why, in the present paper, it is proposed to use Machine Learning (ML) approaches based on Artificial Neural Networks (ANNs) \cite{Hay94,Hag96,Dem14} that avoid solving such a computationally expensive optimization problem and that allow implementing a dedicated software on devices with general-purpose (regular) CPUs. For introducing the last quantity of interest, a Computational Homogenization Model that implements the SUBC homogenization approach is constructed in using a finite element mesh of $\Omega^{\text{meso}}$ made of $101 \times 101 = 10\,201$ nodes and $100 \times 100 = 10^4$ quadrangular elements. This Computational Homogenization Model is then used for computing the vector $\ellb^{\text{eff}} = (\log([L^{\text{eff}}]_{11}),[L^{\text{eff}}]_{12},[L^{\text{eff}}]_{13},\log([L^{\text{eff}}]_{22}),[L^{\text{eff}}]_{23},\log([L^{\text{eff}}]_{33})) \in \Rbb^6$ whose components are defined from the $6$ components of the invertible upper triangular real matrix $[L^{\text{eff}}]$ (with strictly positive diagonal entries $[L^{\text{eff}}]_{11}$, $[L^{\text{eff}}]_{22}$ and $[L^{\text{eff}}]_{33}$) corresponding to the Cholesky factorization of the effective compliance matrix $[S^{\text{eff}}] \in \Mbb_3^+(\Rbb)$, that is $[S^{\text{eff}}] = [L^{\text{eff}}]^T [L^{\text{eff}}]$, where the superscript ${}^T$ denotes the transpose operator. Hence, computing the vector-valued quantity of interest $\ellb^{\text{eff}}$ by using the Computational Homogenization Model for any fine-scale matrix-valued compliance field $[S^{\text{meso}}]$ allows defining a nonlinear mapping $\Mcb^{\text{EFF}}$ defined from $\Mbb_3^+(\Rbb)$ into $\Rbb^6$ such that
\begin{equation}\label{mappingEFF}
\ellb^{\text{eff}} = \Mcb^{\text{EFF}}([S^{\text{meso}}]).
\end{equation}
Additional details can be found in \cite{Ngu15,Ngu16,Zha20} for the explicit construction of both nonlinear mappings $\Mcb^{\text{HFCMM}}$ and $\Mcb^{\text{EFF}}$.

\section{Prior stochastic model of the uncertainties on the matrix-valued compliance field}
\label{sec:stochastic_model}

In the present work, the material is assumed to be heterogeneous and anisotropic with a complex microstructure that cannot be properly described and numerically characterized from the morphological and mechanical properties of its micro-constituents. Matrix-valued compliance field $[S^{\text{meso}}]$ then represents the apparent elasticity properties of a random heterogeneous anisotropic material at a given mesoscale which corresponds to the fine scale that has been introduced in Section~\ref{sec:definition_HFCMM} for the HFCMM. Since the material is random, matrix-valued compliance field $[S^{\text{meso}}]$ is then considered as uncertain and modeled as a matrix-valued random compliance field $[\Sb^{\text{meso}}]$ indexed by $\Rbb^2$ and restricted to bounded domain $\Omega^{\text{macro}}$. A prior stochastic model of 2D matrix-valued random compliance field $[\Sb^{\text{meso}}]$ is constructed as a block matrix decomposition of a 3D matrix-valued random compliance field $[\Sb]$ in the ensemble SFE$^+$ of non-Gaussian second-order stationary almost surely (a.s.) positive-definite symmetric real matrix-valued random fields introduced in \cite{Soi06} (see \cite{Soi17a} for an overview of the existing stochastic models and associated random generators for non-Gaussian random elasticity or compliance matrices or fields). This ensemble is adapted to the representation of elliptic stochastic partial differential operators (so as to ensure the existence and uniqueness of a second-order random solution to the underlying elliptic stochastic boundary value problem) and to the statistical inverse problem related to their experimental identification. Recall that, by construction, random compliance field $[\Sb^{\text{meso}}]$ satisfies a.s. the classical major and minor symmetry properties as well as the usual positive-definiteness properties and therefore takes its values in $\Mbb_3^+(\Rbb)$. As a consequence of such a block decomposition for constructing random compliance field $[\Sb^{\text{meso}}]$, this latter is defined through a deterministic nonlinear mapping $\Gcb$ defined from $\Rbb^6$ into $\Mbb_3^+(\Rbb)$ as
\begin{equation}\label{mapping}
[\Sb^{\text{meso}}] = \Gcb(\Ub;\,\delta,\,\underline{\kappa},\,\underline{\mu}) \quad \text{with} \quad \Ub = \Ucb(\ell),
\end{equation}
where $\delta$ is a positive bounded dispersion parameter such that $0 \leq \delta < \delta_{\text{sup}}$ with $\delta_{\text{sup}} = \sqrt{7/11} \approx 0.7977 <1$ and controlling the level of statistical fluctuations exhibited by random compliance field $[\Sb^{\text{meso}}]$ around its mean function $[\underline{S}^{\text{meso}}]$, which is assumed to be independent of spatial position $\xb$ and completely defined by a mean bulk modulus $\underline{\kappa}$ and a mean shear modulus $\underline{\mu}$ in the particular case of an isotropic mean elastic material, and where $\Ucb(\ell)$ is an explicit random algebraic or spectral representation of a second-order homogeneous normalized Gaussian $\Rbb^{6}$-valued random field $\Ub$ indexed by $\Rbb^2$ whose spatial correlation structure is parameterized by a unique spatial correlation length $\ell$. The prior stochastic model of $[\Sb^{\text{meso}}]$ is finally parameterized by a four-dimensional vector-valued hyperparameter $\hb = (\delta,\ell,\underline{\kappa},\underline{\mu})$ belonging to the admissible set $\Hc = \Hc_1 \times \Hc_2 \times \Hc_3 \times \Hc_4 \subset (\Rbb^+)^4$, with $\Hc_1 = \intervaloo{0}{\delta_{\text{sup}}}$ and $\Hc_2 = \Hc_3 = \Hc_4 = \intervaloo{0}{+\infty}$, and characterizing the complete probabilistic information of random compliance field $[\Sb^{\text{meso}}]$.

Additional details can be found in \cite{Soi06,Soi08a,Soi17a} for the fundamental (algebraic and statistical) properties of random compliance field $[\Sb]$, the explicit construction of the deterministic nonlinear mapping $\Gcb$ and an overview of the numerical methods allowing for the algebraic or spectral representation and numerical simulation (generation of realizations) of homogeneous Gaussian vector- or real-valued random fields. In the present work, we have used the spectral representation method (also called the Shinozuka method) initially introduced in \cite{Shi71,Shi72a,Shi72b} and later revisited and studied in \cite{Poi89,Poi95} from a mathematical standpoint, which is classical numerical simulation method based on the stochastic integral representation of homogeneous Gaussian random fields. The interested reader can refer to \cite{Zha19} for the algebraic representation of $[\Sb^{\text{meso}}]$ and the algorithm for generating independent realizations of $[\Sb^{\text{meso}}]$.

\section{Construction of the initial database}
\label{sec:construction_database}

In order to construct an \emph{ad hoc} database for training an ANN that can be used for the statistical identification of the prior stochastic model of $[\Sb^{\text{meso}}]$, the unknown vector-valued hyperparameter $\hb = (\delta,\ell,\underline{\kappa},\underline{\mu})$ is modeled as a random vector $\Hb = (D,L,\underline{K},\underline{M}) = (H_1,H_2,H_3,H_4)$ with statistically independent random components $H_1=D$, $H_2=L$, $H_3=\underline{K}$ and $H_4=\underline{M}$. Hence, mappings $\Mcb^{\text{HFCMM}}$, $\Mcb^{\text{EFF}}$ and $\Gcb$ respectively defined in \eqref{mappingHFCMM}, \eqref{mappingEFF} and \eqref{mapping} allow for defining the random vector of quantities of interest $\Qb = (Q_1,\dots,Q_n)$ with values in $\Rbb^n$ with $n=9$, given random vector $\Hb = (H_1,\dots,H_m)$ with values in $\Hc = \Hc_1 \times \dots \times \Hc_m \subset \Rbb^m$ with $m=4$, such that
\begin{subequations}\label{randomQoI}
\begin{alignat}{2}
\(Q_1,\,Q_2,\,Q_3\) &= \Mcb^{\text{HFCMM}}\Big(\Gcb(\Ub;\, H_1,\,H_3,\,H_4)\Big) \quad &&\text{with} \quad \Ub = \Ucb(H_2),\label{randomQoIa}\\
\(Q_4,\,\dots,\,Q_9\) &= \Mcb^{\text{EFF}}\Big(\Gcb(\Ub;\, H_1,\,H_3,\,H_4)\Big) \quad &&\text{with} \quad \Ub = \Ucb(H_2).\label{randomQoIb}
\end{alignat}
\end{subequations}

The probabilistic model of random vector $\Hb$ is constructed by using the MaxEnt principle \cite{Jay57a,Jay57b,Sob90,Kap92,Jum00,Jay03,Cov06,Soi17a} with the following algebraically independent constraints to be satisfied: (i) the components $H_1$, $H_2$, $H_3$ and $H_4$ of $\Hb$ are mutually statistically independent random variables, (ii) the support of the probability density function of $\Hb$ is a known bounded hypercube $\Hc_{\text{ad}} \subset \Hc$. Then, the MaxEnt principle leads to a uniform $\Rbb^m$-valued random variable $\Hb$ with compact support $\Hc_{\text{ad}}$ and mutually statistically independent components. Note that in all the following, the reduced admissible set $\Hc_{\text{ad}} = \intervalcc{0.25}{0.65} \times \intervalcc{20}{250} \times \intervalcc{8.5}{17} \times \intervalcc{2.15}{5.00}$ in $[-]\times[\mu\text{m}]\times[\text{GPa}]\times[\text{GPa}]$ has been chosen sufficiently large so that the database can cover a large enough and realistic range of values of the hyperparameters for the application presented in Section~\ref{sec:results_real_data} corresponding to a random heterogeneous microstructure made up of a biological tissue (bovine cortical bone) and by considering the results obtained in \cite{Ngu16}.
Furthermore, in practice, the bounds of admissible set $\Hc_{\text{ad}}$ may be \emph{a posteriori} considered as incorrect if any component of output vector $\hb^{\ast}$ is close to the corresponding bounds of $\Hc_{\text{ad}}$, which is not the case for the numerical examples presented in this paper. 

\begin{figure}[h!]
\centering
\tikzsetnextfilename{flowchart_compute_qoi}
\tikzstyle{line} = [draw, -latex', thick, rounded corners=4pt]
\tikzstyle{block} = [rectangle, draw, thick, fill=cyan!20, text width=10em, text centered, node distance=0.5cm, minimum height=4em]
\tikzstyle{inputcloud} = [rectangle, draw, thick, fill=green!20, text width=10em, text centered, rounded corners, node distance=0.5cm, minimum height=4em]
\tikzstyle{outputcloud} = [rectangle, draw, thick, fill=red!20, text width=10em, text centered, rounded corners, node distance=0.5cm, minimum height=4em]

\begin{tikzpicture}
\centering
\node [inputcloud] (0) {Random vector of hyperparameters\\ $\Hb = (H_1,H_2,H_3,H_4)$};
\node [inputcloud, below= of 0] (1) {Gaussian\\ random field\\ $\Ub = \Ucb(H_2)$};
\node [block, right= of 0] (2) {Prior\\ Stochastic Model\\ $\Gcb$};
\node [inputcloud, right= of 2] (3) {Random\\ compliance field\\ $[\Sb^{\text{meso}}]$};
\node [block, above= of 3] (4) {High Fidelity\\ Computational\\ Mechanical Model\\ $\Mcb^{\text{HFCMM}}$};
\node [block, below= of 3] (5) {Homogenization\\ Computational Model\\ $\Mcb^{\text{EFF}}$};
\node [outputcloud, right= of 4] (6) {Random vector\\ $(D^{\epsilonb},L_1^{\epsilonb},L_2^{\epsilonb}) = (Q_1,Q_2,Q_3)$};
\node [outputcloud, right= of 5] (7) {Random vector\\ $\Lb^{\text{eff}} = (Q_4,\dots,Q_9)$};
\node [outputcloud, right= of 3] (8) {Random vector of quantities of interest\\ $\Qb = (Q_1,\dots,Q_9)$};

\path [line] (0) -- (1);
\path [line] (0) -- (2);
\path [line] (1) -| (2);
\path [line] (2) -- (3);
\path [line] (3) -- (4);
\path [line] (3) -- (5);
\path [line] (4) -- (6);
\path [line] (5) -- (7);
\path [line] (6) -- (8);
\path [line] (7) -- (8);
\end{tikzpicture}
\caption{Flowchart for computing the quantities of interest from the hyperparameters. Blue blocks refer to the models, green blocks refer to the random input parameters and fields of the computational models and red blocks refer to the random output quantities of interest of the computational models.}\label{fig:flowchart_compute_qoi}
\end{figure}
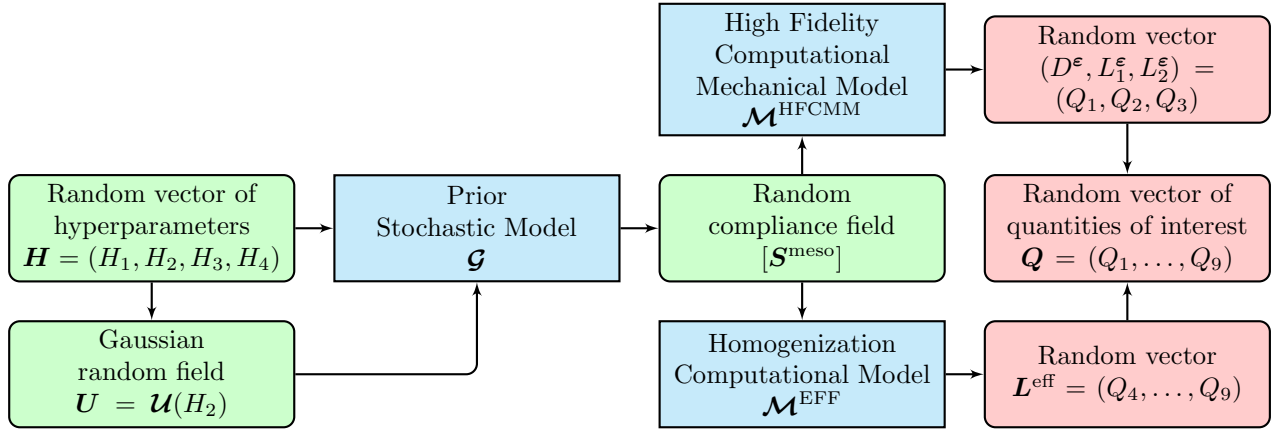

The required numerical database should contain a set of network input and target (desired network output) vectors, where the input vectors define data regarding the random vector $\Qb$ of quantities of interest, and the target vectors define data regarding the random vector $\Hb$ of hyperparameters. Such a database has been numerically simulated and constructed by using the random generator defined by \eqref{randomQoI}. For each realization $\hb^{(i)} = (h^{(i)}_1,\dots,h^{(i)}_m)$ of uniformly distributed random vector $\Hb = (H_1,\dots,H_m)$, a realization of homogeneous normalized Gaussian random field $\Ub$ is generated using mapping $\Ucb$, then the corresponding realization of random compliance field $[\Sb^{\text{meso}}]$ is generated using mapping $\Gcb$, and finally the associated realization $\qb^{(i)} = (q^{(i)}_1,\dots,q^{(i)}_n)$ of random vector $\Qb = (Q_1,\dots,Q_n)$ is numerically simulated using mappings $\Mcb^{\text{HFCMM}}$ and $\Mcb^{\text{EFF}}$. The construction of the database is then straightforward and it consists of $N_d$ independent realizations $\xb^{(1)},\dots,\xb^{(N_d)}$ of random vector $\Xb = (\Qb,\Hb)$. Hence, each element of the database can be written as $\xb^{(i)} = (\qb^{(i)},\hb^{(i)})$ for $i=1,\dots,N_d$. Figure~\ref{fig:flowchart_compute_qoi} provides a schematic representation of the key steps allowing the computation of the quantities of interest from the hyperparameters. Hereinafter, this database will be referred as the \emph{initial database}.

\section{Formulation of the statistical inverse problem}
\label{sec:inverse_problem}

Solving the statistical inverse problem under consideration in this paper can be formulated as solving an optimization problem, as for instance calculating the value $\hb^{\ast}$ of $\Hb$ as the Maximum A Posteriori (MAP) or the Maximum Likelihood Estimation (MLE). Another possible estimation of $\hb^\ast$ can be chosen as the conditional mathematical expectation of $\Hb$ given $\Qb$ is equal to a given observation $\qb^{\text{obs}}$. Such estimations of $\hb^{\ast}$ can be calculated by using usual nonparametric statistical methods \cite{Bow97,Hor12,Giv13,Sco15,Soi17a} with the database constructed in Section~\ref{sec:construction_database}.
Nevertheless, these estimations of $\hb^{\ast}$ require that the database is always available and sufficiently powerful CPUs for performing the computation in a reasonable computing time when digital twin applications are concerned, for instance.
Since the required database may contain a large amount of data to be recorded, such a direct approach can be tricky to carry out in practice.
An alternative approach is proposed in the present work and consists in designing an ANN that can predict another probable value $\hb^{\ast}$ of random vector $\Hb$ given $\Qb = \qb^{\text{obs}}$ with the available database for which the inputs will be the $N_d$ independent realizations $\qb^{(1)},\dots,\qb^{(N_d)}$ of random vector $\Qb$, and the corresponding targets will be the $N_d$ independent realizations $\hb^{(1)},\dots,\hb^{(N_d)}$ of random vector $\Hb$.
Indeed, ANNs are known for being particularly well-suited for addressing and solving function approximation and nonlinear regression problems.
The statistical inverse problem related to the statistical identification of $\hb^{\ast}$ can then be viewed as a function approximation problem and solved by using an ANN trained from the available database.
The solution $\hb^{\ast}$ of the statistical inverse problem can be simply defined as the output vector $\hb^{\text{out}}$ of the trained ANN for the given input vector $\qb^{\text{obs}}$.
Within the framework of ML techniques based on ANNs, the \emph{network input data} of the initial database will refer to the $N_d$ independent realizations $\qb^{(1)},\dots,\qb^{(N_d)}$ of random vector $\Qb$ and the \emph{network target data} of the initial database will refer to the $N_d$ independent realizations $\hb^{(1)},\dots,\hb^{(N_d)}$ of random vector $\Hb$.
Nevertheless, it should be noted that since in \eqref{randomQoIa} and \eqref{randomQoIb}, mapping $\Ucb$ is random for any input argument, then the mapping between $\Qb$ and $\Hb$ is random too.
As a consequence, the supervised training of an ANN with the initial database cannot be efficient since a trained ANN is a deterministic mapping between its inputs and outputs.
This is the reason why $\Qb$ is substituted by another network input vector $\widetilde{\Qb}$ such that the mapping between $\widetilde{\Qb}$ and $\Hb$ is (almost) deterministic.
It would then make it possible to efficiently train an artificial neural network.
In the next section, $\widetilde{Q}_k$ is defined as the conditional mathematical expectation of $Q_k$ given $\Hb$.
In practice, an observation $\tilde{q}^{\text{obs}}_k$ of $\widetilde{Q}_k$ is not available and cannot be deduced from a unique observation $q^{\text{obs}}_k$ of $Q_k$ since it would be equivalent to solve the statistical inverse problem.
Nevertheless, we propose to calculate $\hb^{\ast}$ as the output of the trained ANN with observation $\qb^{\text{obs}} = (q^{\text{obs}}_1,\dots,q^{\text{obs}}_9)$ as input (instead of $\tilde{\qb}^{\text{obs}} = (\tilde{q}^{\text{obs}}_1,\dots,\tilde{q}^{\text{obs}}_9)$).

\section{Construction of the processed database by conditioning the initial database}
\label{sec:conditioning_database}

For a robust computation of the solution $\hb^{\ast}$ of the statistical inverse problem, the network input data consisting of the $N_d$ inputs $q_k^{(1)},\dots,q_k^{(N_d)}$ and $\hb^{(1)},\dots,\hb^{(N_d)}$ for $k=1,\dots,n$ of the initial database are postprocessed and replaced with $N_d$ new inputs $\tilde{q}_k^{(1)},\dots,\tilde{q}_k^{(N_d)}$ defined as the values taken by the conditional mathematical expectation $\Ebb\{Q_k \vert \Hb\}$ of random variable $Q_k$ given random vector $\Hb$ evaluated at $\hb^{(1)},\dots,\hb^{(N_d)}$, respectively. We then have 
\begin{equation}\label{condexpectation}
\tilde{q}_k^{(i)} = \Ebb\{Q_k \vert \Hb=\hb^{(i)}\} = \int_{\Rbb} q \, p_{Q_k\vert \Hb}(q\vert \hb^{(i)}) \, dq,
\end{equation}
where $q\mapsto p_{Q_k\vert \Hb}(q\vert \hb)$ is the conditional pdf of random variable $Q_k$ given event $\Hb=\hb$ for any $\hb\in\Hc$. A nonparametric estimate of the conditional pdf $q\mapsto p_{Q_k\vert \Hb}(q\vert \hb)$ can be constructed by using the multivariate kernel density estimation method with a Gaussian kernel function, that is one of the most efficient and popular kernel smoothing techniques in nonparametric statistics \cite{Bow97,Hor12,Giv13,Sco15,Soi17a}, and the $N_d$ independent realizations $q_k^{(1)},\dots,q_k^{(N_d)}$ and $\hb^{(1)},\dots,\hb^{(N_d)}$ of $Q_k$ and $\Hb$, respectively. We then have

\begin{equation}\label{condpdfestimateexpr}
p_{Q_k\vert \Hb}(q\vert \hb) = \dfrac{p_{Q_k,\Hb}(q,\hb)}{p_{\Hb}(\hb)} \simeq \dfrac{1}{b_{Q_k}} \dfrac{\displaystyle\sum_{i=1}^{N_d} \Kc\(\dfrac{q-q_k^{(i)}}{b_{Q_k}}\) \, \prod_{j=1}^m \Kc\(\dfrac{h_j-h_j^{(i)}}{b_{H_j}}\)}{\displaystyle\sum_{i=1}^{N_d} \prod_{j=1}^m \Kc\(\dfrac{h_j-h_j^{(i)}}{b_{H_j}}\)},
\end{equation}
where $(q,\hb) \mapsto p_{Q_k,\Hb}(q,\hb)$ is the joint pdf of random vector $(Q_k,\Hb)$ and $\hb \mapsto p_{\Hb}(\hb)$ is the joint pdf of random vector $\Hb$, $x \mapsto \Kc(x)$ is a one-dimensional kernel function and the bandwidths $b_{Q_k}$, $b_{H_1}, \dots, b_{H_m}$ are positive real values. 
In the present work, the Gaussian kernel function and the usual multidimensional optimal Silverman smoothing parameters computed using the so-called Silverman's rule of thumb \cite{Sil86} are chosen, and we then have
\begin{equation*}
\Kc(x) = \dfrac{1}{\sqrt{2\pi}} e^{-x^2/2} \quad \text{and} \quad b_S = \hat{\sigma}_S \(\dfrac{4}{N_d(2+m+1)}\)^{1/(4+m+1)},
\end{equation*}
where $\hat{\sigma}_S$ is a robust empirical estimate of the standard deviation of the real-valued random variable $S$, for $S=Q_k,H_1,\dots,H_m$. Finally, the trapezoidal numerical integration method is employed to compute the integral of the one-dimensional function $q \mapsto q \, p_{Q_k\vert \Hb}(q\vert \hb^{(i)})$ in \eqref{condexpectation}. Note that for high-dimensional functions, the numerical integration could have been performed using a Markov Chain Monte Carlo (MCMC) method \cite{Rob04,Kai05,Spa05a}. The conditioning of the initial database allows constructing a second database that consists of $N_d$ elements $\tilde{\xb}^{(1)},\dots,\tilde{\xb}^{(N_d)}$ that are written as $\tilde{\xb}^{(i)} = (\tilde{\qb}^{(i)},\hb^{(i)})$ with $\tilde{\qb}^{(i)} = (\tilde{q}^{(i)}_1,\dots,\tilde{q}^{(i)}_n)$ for $i=1,\dots,N_d$. Hereinafter, vectors $\tilde{\qb}^{(1)},\dots,\tilde{\qb}^{(N_d)}$ are modeled as statistically independent realizations of a vector-valued random variable $\widetilde{\Qb}$ for which the probabilistic model is indirectly constructed by using the nonparametric statistics as presented in this section. In the following, the database containing the $N_d$ elements $\tilde{\xb}^{(1)},\dots,\tilde{\xb}^{(N_d)}$ will be referred to as the \emph{processed database}. Within the framework of ML techniques based on ANNs, the \emph{network input data} of the processed database will refer to the $N_d$ realizations $\tilde{\qb}^{(1)},\dots,\tilde{\qb}^{(N_d)}$ of random vector $\widetilde{\Qb}$ and the \emph{network target data} of the processed database will still refer to the $N_d$ realizations $\hb^{(1)},\dots,\hb^{(N_d)}$ of random vector $\Hb$ as for the initial database. 

In the present work, the complete initial (resp. processed) database consists of $N_d = 200\, 000$ independent realizations of the $9$-element random vector $\Qb$ (resp. $\widetilde{\Qb}$) and of the $4$-element random vector $\Hb$. Such a large dataset (spanning the full range of the admissible output space $\Hc_{\text{ad}}$) is expected to cover the full range of the input space for which the ANN will be used after training. It should be mentioned that ANNs can reliably and accurately predict future outputs for new inputs belonging to the range for which they have been trained, but are generally not able to accurately extrapolate and generalize beyond (outside) this range. The values of the empirical estimate $\hat{\sigma}_S$ of the standard deviation for each input random variable $S=Q_1,\dots,Q_9$ and each output random variable $S=H_1,H_2,H_3,H_4$ are reported in Table~\ref{tab:sig}. In the following, both numerical databases (initial and processed) will be used to train a predictive model by designing an ANN that can reliably and accurately predict the output vector $\hb^{\text{out}}$ for a given observed input vector $\qb^{\text{obs}}$.

\begin{table}[h!]
\caption{Empirical estimates $\hat{\sigma}_S$ of the standard deviation for each input random variable $S=Q_1,\dots,Q_9$ and each output random variable $S=H_1,H_2,H_3,H_4$}
\label{tab:sig}
\centering
\begin{tabular}{|c|c|} \hline
Random variable $S$ & Estimate $\hat{\sigma}_S$ of the standard deviation for $S$ \\ \hline
$Q_1$ & $9.17\times 10^{-2}$ \\ \hline
$Q_2$ & $7.31\times 10^{-2}$ \\ \hline
$Q_3$ & $6.85\times 10^{-5}$ \\ \hline
$Q_4$ & $1.20\times 10^{-1}$ \\ \hline
$Q_5$ & $5.44\times 10^{3}$ \\ \hline
$Q_6$ & $1.84\times 10^{3}$ \\ \hline
$Q_7$ & $1.34\times 10^{-1}$ \\ \hline
$Q_8$ & $1.83\times 10^{3}$ \\ \hline
$Q_9$ & $1.47\times 10^{-1}$ \\ \hline
$H_1$ & $0.148$ \\ \hline
$H_2$ & $85.2$~$\mu$m \\ \hline
$H_3$ & $3.14$~GPa \\ \hline
$H_4$ & $1.05$~GPa \\ \hline
\end{tabular}
\end{table}

\section{Statistical analysis of the initial and processed databases}
\label{sec:sensitivity_analysis_databases}

\begin{figure}[h!]
	\centering
	\begin{subfigure}{.48\linewidth}
		\centering
		\tikzsetnextfilename{matrix_correlation_coefficients_inputs_targets_initial}
%
\definecolor{mycolor1}{rgb}{0.00000,0.00000,0.87500}%
\definecolor{mycolor2}{rgb}{0.00000,0.00000,0.81250}%
\definecolor{mycolor3}{rgb}{0.00000,0.06250,1.00000}%
\definecolor{mycolor4}{rgb}{0.00000,0.12500,1.00000}%
\definecolor{mycolor5}{rgb}{1.00000,1.00000,0.00000}%
\definecolor{mycolor6}{rgb}{0.00000,0.43750,1.00000}%
\definecolor{mycolor7}{rgb}{0.00000,0.50000,1.00000}%
\definecolor{mycolor8}{rgb}{0.62500,0.00000,0.00000}%
\definecolor{mycolor9}{rgb}{0.00000,0.00000,0.56250}%
\begin{tikzpicture}

\begin{axis}[%
width=0.444\figureheight,
height=\figureheight,
at={(0\figureheight,0\figureheight)},
scale only axis,
point meta min=-0.0821222108268344,
point meta max=0.96072864785625,
axis on top,
xmin=0.5,
xmax=4.5,
xtick={1,2,3,4},
xticklabels={{$H_1$},{$H_2$},{$H_3$},{$H_4$}},
xlabel={Targets},
y dir=reverse,
ymin=0.5,
ymax=9.5,
ytick={1,2,3,4,5,6,7,8,9},
yticklabels={{$Q_1$},{$Q_2$},{$Q_3$},{$Q_4$},{$Q_5$},{$Q_6$},{$Q_7$},{$Q_8$},{$Q_9$}},
ylabel={Inputs},
colormap/jet,
colorbar
]
	\fill [black!10!mycolor8] (axis cs:0.5,0.5) rectangle (axis cs:1.5,1.5);
	\fill [mycolor2] (axis cs:0.5,1.5) rectangle (axis cs:1.5,2.5);
	\fill [mycolor2] (axis cs:0.5,2.5) rectangle (axis cs:1.5,3.5);
	\fill [mycolor3] (axis cs:0.5,3.5) rectangle (axis cs:1.5,4.5);
	\fill [blue] (axis cs:0.5,4.5) rectangle (axis cs:1.5,5.5);
	\fill [mycolor1] (axis cs:0.5,5.5) rectangle (axis cs:1.5,6.5);
	\fill [mycolor3] (axis cs:0.5,6.5) rectangle (axis cs:1.5,7.5);
	\fill [mycolor1] (axis cs:0.5,7.5) rectangle (axis cs:1.5,8.5);
	\fill [blue] (axis cs:0.5,8.5) rectangle (axis cs:1.5,9.5);
	\fill [mycolor9] (axis cs:1.5,0.5) rectangle (axis cs:2.5,1.5);
	\fill [black!25!red] (axis cs:1.5,1.5) rectangle (axis cs:2.5,2.5);
	\fill [mycolor8] (axis cs:1.5,2.5) rectangle (axis cs:2.5,3.5);
	\fill [black!25!blue] (axis cs:1.5,3.5) rectangle (axis cs:2.5,4.5);
	\fill [mycolor2] (axis cs:1.5,4.5) rectangle (axis cs:2.5,5.5);
	\fill [mycolor1] (axis cs:1.5,5.5) rectangle (axis cs:2.5,6.5);
	\fill [black!25!blue] (axis cs:1.5,6.5) rectangle (axis cs:2.5,7.5);
	\fill [mycolor1] (axis cs:1.5,7.5) rectangle (axis cs:2.5,8.5);
	\fill [black!25!blue] (axis cs:1.5,8.5) rectangle (axis cs:2.5,9.5);
	\fill [mycolor2] (axis cs:2.5,0.5) rectangle (axis cs:3.5,1.5);
	\fill [mycolor2] (axis cs:2.5,1.5) rectangle (axis cs:3.5,2.5);
	\fill [mycolor1] (axis cs:2.5,2.5) rectangle (axis cs:3.5,3.5);
	\fill [mycolor7] (axis cs:2.5,3.5) rectangle (axis cs:3.5,4.5);
	\fill [mycolor5] (axis cs:2.5,4.5) rectangle (axis cs:3.5,5.5);
	\fill [mycolor1] (axis cs:2.5,5.5) rectangle (axis cs:3.5,6.5);
	\fill [mycolor4] (axis cs:2.5,6.5) rectangle (axis cs:3.5,7.5);
	\fill [mycolor2] (axis cs:2.5,7.5) rectangle (axis cs:3.5,8.5);
	\fill [mycolor1] (axis cs:2.5,8.5) rectangle (axis cs:3.5,9.5);
	\fill [blue!50!mycolor1] (axis cs:3.5,0.5) rectangle (axis cs:4.5,1.5);
	\fill [mycolor1] (axis cs:3.5,1.5) rectangle (axis cs:4.5,2.5);
	\fill [mycolor2] (axis cs:3.5,2.5) rectangle (axis cs:4.5,3.5);
	\fill [mycolor8] (axis cs:3.5,3.5) rectangle (axis cs:4.5,4.5);
	\fill [mycolor6] (axis cs:3.5,4.5) rectangle (axis cs:4.5,5.5);
	\fill [mycolor1] (axis cs:3.5,5.5) rectangle (axis cs:4.5,6.5);
	\fill [black!50!red] (axis cs:3.5,6.5) rectangle (axis cs:4.5,7.5);
	\fill [mycolor1] (axis cs:3.5,7.5) rectangle (axis cs:4.5,8.5);
	\fill [black!50!red] (axis cs:3.5,8.5) rectangle (axis cs:4.5,9.5);
\end{axis}
\end{tikzpicture}%
		\caption{Initial database with input random vector $\Qb~=~(Q_1,\dots,Q_9)$}
		\label{fig:matrix_correlation_coefficients_inputs_targets_initial}
	\end{subfigure}\hfill
	\begin{subfigure}{.48\linewidth}
		\centering
		\tikzsetnextfilename{matrix_correlation_coefficients_inputs_targets_processed}
%
\definecolor{mycolor1}{rgb}{0.00000,0.06250,1.00000}%
\definecolor{mycolor2}{rgb}{0.00000,0.00000,0.87500}%
\definecolor{mycolor3}{rgb}{0.00000,0.00000,0.68750}%
\definecolor{mycolor4}{rgb}{0.00000,0.12500,1.00000}%
\definecolor{mycolor5}{rgb}{0.00000,0.18750,1.00000}%
\definecolor{mycolor6}{rgb}{0.00000,0.81250,1.00000}%
\definecolor{mycolor7}{rgb}{0.56250,0.00000,0.00000}%
\definecolor{mycolor8}{rgb}{0.00000,0.00000,0.56250}%
\begin{tikzpicture}

\begin{axis}[%
width=0.444\figureheight,
height=\figureheight,
at={(0\figureheight,0\figureheight)},
scale only axis,
point meta min=-0.0888329660868738,
point meta max=0.995394146864279,
axis on top,
xmin=0.5,
xmax=4.5,
xtick={1,2,3,4},
xticklabels={{$H_1$},{$H_2$},{$H_3$},{$H_4$}},
xlabel={Targets},
y dir=reverse,
ymin=0.5,
ymax=9.5,
ytick={1,2,3,4,5,6,7,8,9},
yticklabels={{$\widetilde{Q}_1$},{$\widetilde{Q}_2$},{$\widetilde{Q}_3$},{$\widetilde{Q}_4$},{$\widetilde{Q}_5$},{$\widetilde{Q}_6$},{$\widetilde{Q}_7$},{$\widetilde{Q}_8$},{$\widetilde{Q}_9$}},
ylabel={Inputs},
colormap/jet,
colorbar
]
	\fill [black!50!red] (axis cs:0.5,0.5) rectangle (axis cs:1.5,1.5);
	\fill [blue!40!mycolor3] (axis cs:0.5,1.5) rectangle (axis cs:1.5,2.5);
	\fill [blue!40!mycolor3] (axis cs:0.5,2.5) rectangle (axis cs:1.5,3.5);
	\fill [mycolor4] (axis cs:0.5,3.5) rectangle (axis cs:1.5,4.5);
	\fill [mycolor4] (axis cs:0.5,4.5) rectangle (axis cs:1.5,5.5);
	\fill [mycolor5] (axis cs:0.5,5.5) rectangle (axis cs:1.5,6.5);
	\fill [mycolor1] (axis cs:0.5,6.5) rectangle (axis cs:1.5,7.5);
	\fill [mycolor3] (axis cs:0.5,7.5) rectangle (axis cs:1.5,8.5);
	\fill [mycolor1] (axis cs:0.5,8.5) rectangle (axis cs:1.5,9.5);
	\fill [mycolor8] (axis cs:1.5,0.5) rectangle (axis cs:2.5,1.5);
	\fill [black!50!red] (axis cs:1.5,1.5) rectangle (axis cs:2.5,2.5);
	\fill [black!50!red] (axis cs:1.5,2.5) rectangle (axis cs:2.5,3.5);
	\fill [black!25!blue] (axis cs:1.5,3.5) rectangle (axis cs:2.5,4.5);
	\fill [blue!40!mycolor3] (axis cs:1.5,4.5) rectangle (axis cs:2.5,5.5);
	\fill [mycolor5] (axis cs:1.5,5.5) rectangle (axis cs:2.5,6.5);
	\fill [black!25!blue] (axis cs:1.5,6.5) rectangle (axis cs:2.5,7.5);
	\fill [black!25!blue] (axis cs:1.5,7.5) rectangle (axis cs:2.5,8.5);
	\fill [black!25!blue] (axis cs:1.5,8.5) rectangle (axis cs:2.5,9.5);
	\fill [blue!40!mycolor3] (axis cs:2.5,0.5) rectangle (axis cs:3.5,1.5);
	\fill [blue!40!mycolor3] (axis cs:2.5,1.5) rectangle (axis cs:3.5,2.5);
	\fill [mycolor2] (axis cs:2.5,2.5) rectangle (axis cs:3.5,3.5);
	\fill [mycolor5!40!mycolor6] (axis cs:2.5,3.5) rectangle (axis cs:3.5,4.5);
	\fill [black!25!red] (axis cs:2.5,4.5) rectangle (axis cs:3.5,5.5);
	\fill [blue!50!mycolor2] (axis cs:2.5,5.5) rectangle (axis cs:3.5,6.5);
	\fill [mycolor4] (axis cs:2.5,6.5) rectangle (axis cs:3.5,7.5);
	\fill [black!25!blue] (axis cs:2.5,7.5) rectangle (axis cs:3.5,8.5);
	\fill [mycolor2] (axis cs:2.5,8.5) rectangle (axis cs:3.5,9.5);
	\fill [blue!50!mycolor2] (axis cs:3.5,0.5) rectangle (axis cs:4.5,1.5);
	\fill [mycolor2] (axis cs:3.5,1.5) rectangle (axis cs:4.5,2.5);
	\fill [mycolor2] (axis cs:3.5,2.5) rectangle (axis cs:4.5,3.5);
	\fill [mycolor7] (axis cs:3.5,3.5) rectangle (axis cs:4.5,4.5);
	\fill [mycolor6] (axis cs:3.5,4.5) rectangle (axis cs:4.5,5.5);
	\fill [mycolor4] (axis cs:3.5,5.5) rectangle (axis cs:4.5,6.5);
	\fill [black!50!red] (axis cs:3.5,6.5) rectangle (axis cs:4.5,7.5);
	\fill [mycolor2] (axis cs:3.5,7.5) rectangle (axis cs:4.5,8.5);
	\fill [black!50!red] (axis cs:3.5,8.5) rectangle (axis cs:4.5,9.5);
\end{axis}
\end{tikzpicture}%
		\caption{Processed database with input random vector $\widetilde{\Qb}~=~(\widetilde{Q}_1,\dots,\widetilde{Q}_9)$}
		\label{fig:matrix_correlation_coefficients_inputs_targets_processed}
	\end{subfigure}
	\caption{Matrix of correlation coefficients between each of the components of random vector $\Qb$ (resp. $\widetilde{\Qb}$) and of random vector $\Hb$ estimated from the $N_d$ network input vectors $\qb^{(1)},\dots,\qb^{(N_d)}$ (resp. $\tilde{\qb}^{(1)},\dots,\tilde{\qb}^{(N_d)}$) and corresponding target vectors $\hb^{(1)},\dots,\hb^{(N_d)}$ for the initial (resp. processed) database}
	\label{fig:matrix_correlation_coefficients_inputs_targets}
\end{figure}
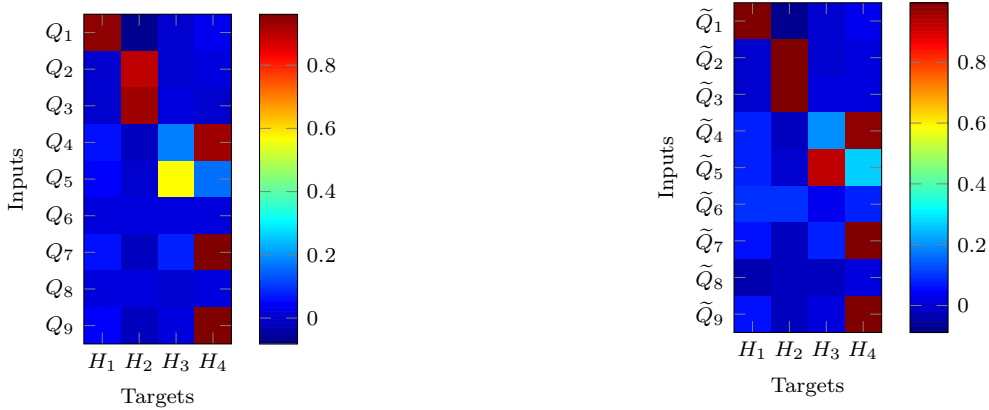

A sensitivity analysis of the network target data with respect to the network input data has been performed for both the initial and processed databases. Figure~\ref{fig:matrix_correlation_coefficients_inputs_targets} shows a classical estimate of the matrix of correlation coefficients between each of the components $Q_1,\dots,Q_9$ (resp. $\widetilde{Q}_1,\dots,\widetilde{Q}_9$) of random vector $\Qb$ (resp. $\widetilde{\Qb}$) and each of the components $H_1,\dots,H_4$ of random vector $\Hb$ computed from the $N_d$ network input vectors $\qb^{(1)},\dots,\qb^{(N_d)}$ (resp. $\tilde{\qb}^{(1)},\dots,\tilde{\qb}^{(N_d)}$) and corresponding target vectors $\hb^{(1)},\dots,\hb^{(N_d)}$ for the initial (resp. processed) database. First, we observe that random variable $H_1$ is highly correlated to random variable $Q_1$ (resp. $\widetilde{Q}_1$) and is almost not correlated with the other components of $\Qb$ (resp. $\widetilde{\Qb}$) for the initial (resp. processed) database. Secondly, random variable $H_2$ is strongly correlated to input random variables $Q_2$ and $Q_3$ (resp. $\widetilde{Q}_2$ and $\widetilde{Q}_3$) and have very small correlation with the other random components of $\Qb$ (resp. $\widetilde{\Qb}$) for the initial (resp. processed) database. Lastly, random variable $H_3$ is mostly correlated with random variable $Q_5$ (resp. $\widetilde{Q}_5$) and to a lesser extent with $Q_4$ (resp. $\widetilde{Q}_4$), while random variable $H_4$ is highly correlated with random variables $Q_4$, $Q_7$ and $Q_9$ (resp. $\widetilde{Q}_4$, $\widetilde{Q}_7$ and $\widetilde{Q}_9$) and to a lesser extent with $Q_5$ (resp. $\widetilde{Q}_5$) for the initial (resp. processed) database. It should be noted that the values of the highest correlation coefficients are higher for the processed database (containing the $N_d$ independent realizations of random vector $\widetilde{\Qb}$ as input data) than for the initial database (containing the $N_d$ independent realizations of random vector $\Qb$ as input data). Also, note that random variables $Q_6$ and $Q_8$ (resp. $\widetilde{Q}_6$ and $\widetilde{Q}_8$) are almost not correlated with any component of $\Hb$ for the initial (resp. processed) database, so that the corresponding realizations $q_6^{(1)},\dots,q_6^{(N_d)}$ and $q_8^{(1)},\dots,q_8^{(N_d)}$ (resp. $\tilde{q}_6^{(1)},\dots,\tilde{q}_6^{(N_d)}$ and $\tilde{q}_8^{(1)},\dots,\tilde{q}_8^{(N_d)}$) could have been removed from the initial (resp. processed) database.

\section{Design of the Artificial Neural Network}
\label{sec:design_network}

In the present work, we focus on multilayer feedforward static neural networks (often referred to as series neural networks) that have only feedforward connections from the input layer (initial layer) to the first hidden layer, then from the first hidden layer to the second hidden layer and so on until the last hidden layer, and finally from the last hidden layer (penultimate layer) to the output layer (last layer). Recall that, while simple two-layer feedforward neural networks (with only one hidden layer and one output layer) have the ability to learn any multidimensional input-output relationship arbitrarily well given consistent data and enough hidden neurons in the hidden layer (see \eg{} \cite{Cyb89,Hor89,Hor91,Les93,Bar93}), multilayer feedforward networks (with more than one hidden layer) are likely to learn complex input-output relationships more quickly and typically show better performance in some practical applications (see \eg{} \cite{LeCun15,Goo16} and the references therein).

\subsection{Definition of the neural network architecture, data division and training algorithm} \label{sec:architecture}

The architecture of the multilayer neural network involves a single input vector with $9$ components and a single output vector with $4$ components. The considered multilayer feedforward neural network is then composed of an input layer with $9$ neurons, an output layer with $4$ neurons, and one (or two) hidden layer(s) of neurons in between. Sigmoid hidden neurons, defined by a hyperbolic tangent sigmoid transfer function \cite{Vogl88}, are used in the hidden layer(s), while linear output neurons, defined by a linear transfer function, are used in the output layer. 
Various configurations have been tested for the two-layer (resp. three-layer) neural network with one (resp. two) hidden layer(s) and one output layer. For the two-layer neural network, the number of hidden neurons in the hidden layer is successively set to $4$, $6$, $8$, $10$, $15$, $20$, $25$, $50$, $75$, $100$, $150$, $200$, $250$, $300$, $350$, $400$, $450$ and $500$, for a total of $18$ different configurations. For the three-layer neural network, the number of hidden neurons in each of the two hidden layers is successively set to $4$, $6$, $8$, $10$, $15$, $20$, $25$, $50$ and $75$, for a total of $81$ different configurations.

The input vectors and target vectors have been randomly divided into three distinct sets for training, validation and testing, with $70\%$ of the complete dataset assigned to the training set, $15\%$ to the validation set and $15\%$ to the test set. Recall that the training and validation sets are used to build the model, while the test set is used to assess the performance of the trained model against test data that was set aside and not used during the training and validation processes in order to evaluate its ability to perform well on unseen data. More precisely, the training dataset is used to train the neural network with the backpropagation training algorithm \cite{Hag96} and adjust the network parameters, namely the weights and biases, according to the training performance function, that is the mean squared error of the training dataset. The validation set is used to measure network generalization and to prematurely interrupt training when generalization stops improving which is indicated by an increase in the validation performance function, that is the mean squared error of the validation dataset. Finally, the test set is used to provide an independent measure of network performance after training and validation. In the present data-fitting problem, the different training, validation and test sets have been simply defined by holding out an \emph{ad hoc} percentage of the entire dataset. The training set consists of $70\%$ of the complete dataset and therefore includes $140\,000$ samples of $13$ elements (with $9$ inputs and $4$ outputs), while the validation and test sets are each set to $15\%$ of the complete dataset and therefore include $30\,000$ samples of $13$ elements (with $9$ inputs and $4$ outputs). The values of both the input and target vectors are preprocessed and mapped into the normalized range $\intervalcc{-1}{1}$ before presenting to the neural network for training. After training, the network output vectors are then transformed back to the original scale (units) of the network target vectors. Such preprocessed and postprocessed transformations allow for the relative accuracy of the $4$ components of output vectors to be optimized equally well although these $4$ output elements have differing target value ranges.

The learning model has been constructed and developed from scratch without using transfer learning and a pretrained model and directly trained on the available training and validation datasets to fit the input vectors and target vectors. The neural network has been set up with initial weight and bias values generated using the Nguyen-Widrow method \cite{Ngu90} for each hidden layer and for the output layer. The neural network has been trained in batch mode, so that the weights and biases are adjusted and updated only once in each iteration corresponding to a full pass over the training dataset after all the input and target vectors in the training dataset are presented and applied to the network. Also, it has been retrained five times starting from several different initial conditions to ensure good network generalization and robust network performance and only the neural network with the best performance on the test dataset is considered for each configuration of the two-layer (resp. three-layer) neural network. As the network training may require considerable resources in terms of computational cost due to the large dataset size ($N_d = 200\,000$), we have used parallel and distributed GPU-based computing to speed up neural network training and simulation and manage large data by taking advantage of the massively parallelized architecture of GPUs. The neural network has been trained and simulated by 
using a high-performance GPU on a single computer with three GPUs and a hundred CPUs.
The scaled conjugate gradient (SCG) algorithm has been chosen as training algorithm, since the conjugate gradient algorithms (in particular, the SCG algorithm) are among the most efficient algorithms for training large networks with thousands or millions of weights and biases on a broad class of problems, including function approximation and pattern recognition problems, with relatively small memory storage requirements compared to Jacobian-based training algorithms, such as the classical Levenberg-Marquardt (LM) and Bayesian Regularization (BR) algorithms. Also, note that the classical LM and BR algorithms have not been considered here, since these training algorithms are based on Jacobian computations that are not supported on GPU hardware (only gradient computations are supported on GPU devices). 
All the computations have been performed using the MATLAB Neural Network Toolbox\texttrademark{} \cite{Bea92} (now part of the Deep Learning Toolbox\texttrademark{}) in conjunction with the Parallel Computing Toolbox\texttrademark{}, the Statistics and Machine Learning Toolbox\texttrademark{} and the Optimization Toolbox\texttrademark{}.

\subsection{Analysis of the neural network performance after training}\label{sec:test_network_performance}

Once the neural network has fit the training dataset and generalized the input-output relationship using the validation dataset, it can be used to generate outputs for inputs it was not trained on and calculate its performance using the test dataset.

\subsubsection{Measures of the neural network performance}\label{sec:measures_performance}

The performances of the trained neural networks have been evaluated by (i) computing the normalized mean squared error between the network outputs and corresponding targets, (ii) performing a linear regression analysis, and (iii) displaying and comparing the marginal pdfs of each component of random vector $\Hb = (H_1,H_2,H_3,H_4)$ of hyperparameters estimated by using the univariate Gaussian kernel density estimation method \cite{Bow97} with the $N_d$ network output data on the one hand and with the $N_d$ network target data on the other hand. 

The normalized mean squared error (mse) measures the neural network performance according to the mean of squared errors (corresponding to the average squared difference between outputs and targets) weighted by the squared distance between the maximum and minimum values for each target element, that is
\begin{equation}
\text{normalized mse} = \frac{1}{4} \sum_{j=1}^4 \frac{1}{N_d} \sum_{i=1}^{N_d} \(\frac{h_j^{\text{out},(i)} - h_j^{\text{target},(i)}}{h_j^{\text{target},\text{max}} - h_j^{\text{target},\text{min}}}\)^2,
\end{equation}
where $\hb^{\text{out},(i)} = (h_1^{\text{out},(i)},h_2^{\text{out},(i)},h_3^{\text{out},(i)},h_4^{\text{out},(i)})$ is the $i$-th network output vector, $\hb^{\text{target},(i)} = (h_1^{\text{target},(i)},h_2^{\text{target},(i)},h_3^{\text{target},(i)},h_4^{\text{target},(i)})$ is the corresponding target vector, $h_j^{\text{target},\text{max}} = \max_{1\leq i \leq N_d} h_j^{\text{target},(i)}$ and $h_j^{\text{target},\text{min}} = \min_{1\leq i \leq N_d} h_j^{\text{target},(i)}$ denote respectively the maximum and minimum values of the $j$-th target element. 

The regression value ($R$-value) is defined and computed as the usual statistical estimate of the correlation coefficient between each output and the corresponding target, such that $R=1$ (resp. $R$ close to $1$) indicates an exact (resp. almost) linear output-target relationship corresponding to a perfect (resp. very good) fit or correlation between output and target values, while $R=0$ (resp. $R$ close to $0$) indicates a random (resp. almost random) output-target relationship corresponding to no (resp. a very poor) fit or correlation. Since the neural network has multiple outputs with different ranges of values, the errors between outputs and corresponding targets have been normalized between $-1$ and $1$ instead of their differing ranges so that the relative accuracy of each output element is optimized equally well (instead of optimizing and favoring the relative accuracy of the output elements with the largest range of values to the detriment of the output elements with the smallest range of values).

\subsubsection{First measure: normalized mean squared error}\label{sec:normalized_mse}

\begin{figure}[h!]
	\centering
	\begin{subfigure}{.5\linewidth}
		\centering
		\tikzsetnextfilename{performance_numWeightElements_2Layers_initial}
%
\begin{tikzpicture}

\begin{axis}[%
width=1.253\figureheight,
height=\figureheight,
at={(0\figureheight,0\figureheight)},
scale only axis,
xmin=0,
xmax=8000,
xlabel={Number of weights and biases},
ymin=1.4e-2,
ymax=1.8e-2,
yminorticks=true,
ylabel={Normalized mean squared error},
xmajorgrids,
ymajorgrids,
yminorgrids,
legend style={legend cell align=left, align=left
},
legend pos=north east
]
\addplot [only marks, mark=*, mark options={}, mark size=1.5000pt, draw=black, fill=black]
  table[row sep=crcr]{%
60	0.0167472077166521\\
88	0.0158348760303678\\
116	0.0154261899473413\\
144	0.0152848480933371\\
214	0.0150104440737352\\
284	0.0149058390733507\\
354	0.0148170464408493\\
704	0.0146151346821522\\
1054	0.0146115815809029\\
1404	0.0146298033562074\\
2104	0.0147397945012077\\
2804	0.0147055168999871\\
3504	0.0147026928799991\\
4204	0.0147541586689776\\
4904	0.0148097509189925\\
5604	0.0148153153368682\\
6304	0.0147347045970654\\
7004	0.0147366881829294\\
};
\addlegendentry{All}

\addplot [only marks, mark=*, mark options={}, mark size=1.5000pt, draw=black, fill=blue]
  table[row sep=crcr]{%
60	0.0167421930039869\\
88	0.0158343723427311\\
116	0.0154275970435646\\
144	0.0152903168979721\\
214	0.0150110628882002\\
284	0.0149052654786158\\
354	0.014813453774766\\
704	0.0146073596831875\\
1054	0.0145982593710328\\
1404	0.0146186965236047\\
2104	0.0147269860469859\\
2804	0.0146947866927321\\
3504	0.0146891195500306\\
4204	0.0147387635568858\\
4904	0.014790629157364\\
5604	0.0147967531297283\\
6304	0.0147169653212457\\
7004	0.0147140175540215\\
};
\addlegendentry{Training}

\addplot [only marks, mark=*, mark options={}, mark size=1.5000pt, draw=black, fill=black!20!green]
  table[row sep=crcr]{%
60	0.0169734200175136\\
88	0.0160333723056536\\
116	0.0156070776489911\\
144	0.0154384340884986\\
214	0.015181438574496\\
284	0.0150611864698431\\
354	0.0149810965562217\\
704	0.0147779974189379\\
1054	0.0147736802755232\\
1404	0.0147841291244843\\
2104	0.0149253940941262\\
2804	0.0148677485105021\\
3504	0.014873366389349\\
4204	0.0149361928383914\\
4904	0.0149983577807176\\
5604	0.0150117712079546\\
6304	0.0149202422739652\\
7004	0.014944806691489\\
};
\addlegendentry{Validation}

\addplot [only marks, mark=*, mark options={}, mark size=1.5000pt, draw=black, fill=red]
  table[row sep=crcr]{%
60	0.0165443974082281\\
88	0.0156387302973868\\
116	0.0152387357966495\\
144	0.0151057410098788\\
214	0.014836561772138\\
284	0.0147531684522879\\
354	0.0146697621005319\\
704	0.0144885552738686\\
1054	0.0145116531990097\\
1404	0.0145273094734102\\
2104	0.0146139676946578\\
2804	0.0145933595899951\\
3504	0.0145953615771687\\
4204	0.0146439683559916\\
4904	0.0147103789448673\\
5604	0.0147054830991014\\
6304	0.0146319502073239\\
7004	0.0146343659426063\\
};
\addlegendentry{Test}

\end{axis}
\end{tikzpicture}%
		\caption{Initial database}
		\label{fig:performance_numWeightElements_2Layers_initial}
	\end{subfigure}\hfill
	\begin{subfigure}{.5\linewidth}
		\centering
		\tikzsetnextfilename{performance_numWeightElements_2Layers_processed}
%
\begin{tikzpicture}

\begin{axis}[%
width=1.269\figureheight,
height=\figureheight,
at={(0\figureheight,0\figureheight)},
scale only axis,
xmin=0,
xmax=8000,
xlabel={Number of weights and biases},
ymin=1e-05,
ymax=5e-04,
yminorticks=true,
ylabel={Normalized mean squared error},
xmajorgrids,
ymajorgrids,
yminorgrids,
legend style={legend cell align=left, align=left
},
legend pos=north east
]
\addplot [only marks, mark=*, mark options={}, mark size=1.5000pt, draw=black, fill=black]
  table[row sep=crcr]{%
60	0.000455853684391478\\
88	0.000290827137932206\\
116	0.000182982449566634\\
144	0.000179047212309489\\
214	0.000153629684498386\\
284	0.000136027768590493\\
354	0.000120655882788103\\
704	9.3899590183101e-05\\
1054	8.52609419595093e-05\\
1404	8.18962051322619e-05\\
2104	6.942045517393e-05\\
2804	6.45931313219266e-05\\
3504	5.49601068282061e-05\\
4204	5.07477586621237e-05\\
4904	4.74120722709879e-05\\
5604	4.56607673031412e-05\\
6304	5.44050339410672e-05\\
7004	4.72334416905565e-05\\
};
\addlegendentry{All}

\addplot [only marks, mark=*, mark options={}, mark size=1.5000pt, draw=black, fill=blue]
  table[row sep=crcr]{%
60	0.000456366038735666\\
88	0.000290846165645933\\
116	0.000183096407645513\\
144	0.000179147777491556\\
214	0.000153820205247581\\
284	0.000136088006073631\\
354	0.000120826926350407\\
704	9.39427627937991e-05\\
1054	8.5147898037346e-05\\
1404	8.18018008769168e-05\\
2104	6.92641953783751e-05\\
2804	6.43855833543638e-05\\
3504	5.47968481993232e-05\\
4204	5.05768128567464e-05\\
4904	4.71752089950691e-05\\
5604	4.55022015678324e-05\\
6304	5.42015720382983e-05\\
7004	4.69687093262776e-05\\
};
\addlegendentry{Training}

\addplot [only marks, mark=*, mark options={}, mark size=1.5000pt, draw=black, fill=black!20!green]
  table[row sep=crcr]{%
60	0.000457571620915234\\
88	0.000292826239593681\\
116	0.000184417835294765\\
144	0.000180543037550158\\
214	0.000155178677567199\\
284	0.000137671883054618\\
354	0.000121942493450964\\
704	9.53275569009672e-05\\
1054	8.71172807041866e-05\\
1404	8.36822054874076e-05\\
2104	7.04731021333921e-05\\
2804	6.62186564892774e-05\\
3504	5.58923312243492e-05\\
4204	5.16037724324459e-05\\
4904	4.8527124884354e-05\\
5604	4.66118122574062e-05\\
6304	5.62334140247541e-05\\
7004	4.86248024965336e-05\\
};
\addlegendentry{Validation}

\addplot [only marks, mark=*, mark options={}, mark size=1.5000pt, draw=black, fill=red]
  table[row sep=crcr]{%
60	0.000451744760928172\\
88	0.000288739240273338\\
116	0.000181015259470401\\
144	0.000177082082885843\\
214	0.000151191594599991\\
284	0.000134102545871723\\
354	0.000118571068834494\\
704	9.22701512819767e-05\\
1054	8.39321415182599e-05\\
1404	8.05507579687277e-05\\
2104	6.90970205937238e-05\\
2804	6.39361633365351e-05\\
3504	5.47897560335165e-05\\
4204	5.06894919835621e-05\\
4904	4.74023816119101e-05\\
5604	4.54496957803168e-05\\
6304	5.35261427369688e-05\\
7004	4.70774985845483e-05\\
};
\addlegendentry{Test}

\end{axis}
\end{tikzpicture}%
		\caption{Processed database}
		\label{fig:performance_numWeightElements_2Layers_processed}
	\end{subfigure}
	\caption{Evolution of the performance function (normalized mean squared error) with respect to the number of weights and biases of the two-layer neural network on the complete dataset (black symbols), training dataset (blue symbols), validation dataset (green symbols) and test dataset (red symbols) for (\subref{fig:performance_numWeightElements_2Layers_initial}) the initial database and (\subref{fig:performance_numWeightElements_2Layers_processed}) the processed database}
	\label{fig:performance_numWeightElements_2Layers}
\end{figure}
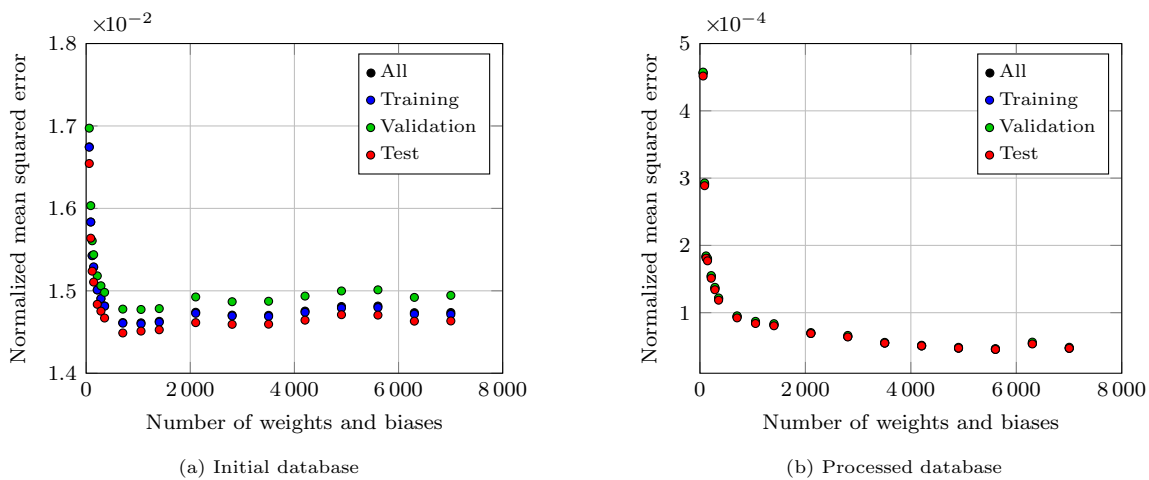

\begin{figure}[h!]
	\centering
	\begin{subfigure}{\linewidth}
		\centering
		\includegraphics[height=0.125\textheight]{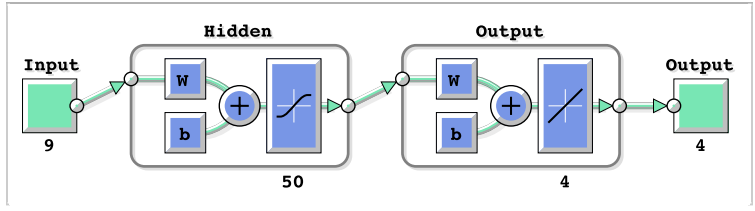}
		\caption{Two-layer neural network with one hidden layer containing $50$ tan-sigmoid hidden neurons and one output layer containing $4$ linear output neurons}
		\label{fig:net_2Layers_initial}
	\end{subfigure}\\
	\begin{subfigure}{\linewidth}
		\centering
		\includegraphics[height=0.125\textheight]{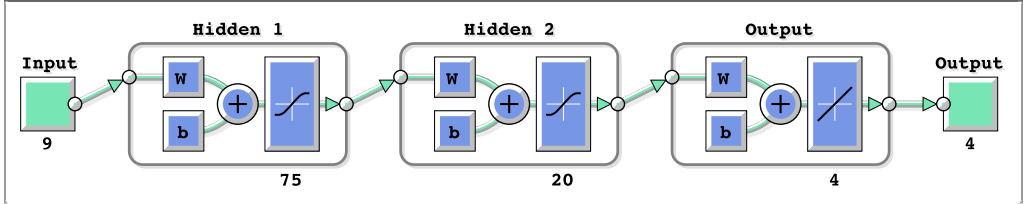}
		\caption{Three-layer neural network with two hidden layers containing $75$ and $20$ tan-sigmoid hidden neurons, respectively, and one output layer containing $4$ linear output neurons}
		\label{fig:net_3Layers_initial}
	\end{subfigure}
	\caption{Initial database: graphical diagrams of (\subref{fig:net_2Layers_initial}) the best two-layer feedforward neural network and (\subref{fig:net_3Layers_initial}) the best three-layer feedforward neural network}\label{fig:net_initial}
\end{figure}

\begin{figure}[h!]
	\centering
	\begin{subfigure}{\linewidth}
		\centering
		\includegraphics[height=0.125\textheight]{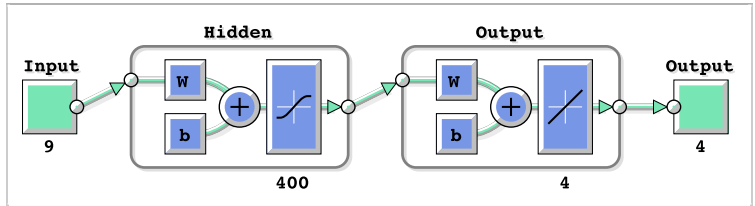}
		\caption{Two-layer neural network with one hidden layer containing $400$ tan-sigmoid hidden neurons and one output layer containing $4$ linear output neurons}
		\label{fig:net_2Layers_processed}
	\end{subfigure}\\
	\begin{subfigure}{\linewidth}
		\centering
		\includegraphics[height=0.125\textheight]{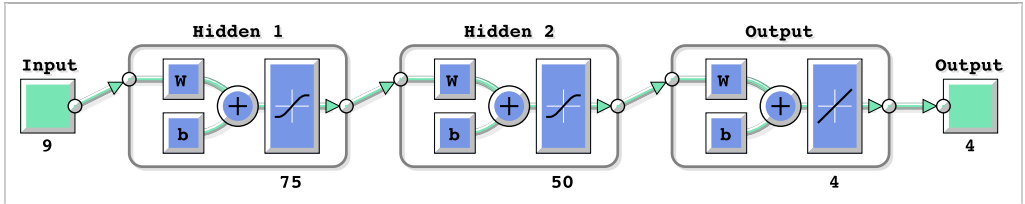}
		\caption{Three-layer neural network with two hidden layers containing $75$ and $50$ tan-sigmoid hidden neurons, respectively, and one output layer containing $4$ linear output neurons}
		\label{fig:net_3Layers_processed}
	\end{subfigure}
	\caption{Processed database: graphical diagrams of (\subref{fig:net_2Layers_processed}) the best two-layer feedforward neural network and (\subref{fig:net_3Layers_processed}) the best three-layer feedforward neural network}\label{fig:net_processed}
\end{figure}

As a first evaluation of the network performance, the normalized mse of the trained neural network is measured for the complete initial (resp. processed) dataset and for each of the training, validation and test subsets. The best trained neural network obtained using the CSG algorithm has been selected as the one with the best performance on the test set, \ie{} the one that generalized best to the test set. Figure~\ref{fig:performance_numWeightElements_2Layers} shows the evolution of the normalized mse (plotted in a linear scale) with respect to the number of network parameters (weights and biases) for the two-layer neural network and for each of the initial and processed databases. For the initial database, the normalized mse slightly decreases and then reaches a plateau from a few hundreds of parameters with a relatively high value of about $1.5\times 10^{-2}$ for the complete, training, validation and test datasets, while for the processed database, the normalized mse sharply decreases with the number of parameters and then converges toward a very low value of $4\times 10^{-5}$ with several thousands of parameters. For the initial database, the best trained two-layer neural network contains $50$ hidden neurons in the hidden layer with $704$ parameters, while the best trained three-layer neural network contains $75$ and $20$ hidden neurons in the first and second hidden layers, respectively, with a total of $2\,354$ parameters. For the processed database, the best trained two-layer neural network contains $400$ hidden neurons in the hidden layer with $5\,604$ parameters, while the best trained three-layer neural network contains $75$ and $50$ hidden neurons in the first and second hidden layers, respectively, with a total of $4\,754$ parameters. Figures~\ref{fig:net_initial} and \ref{fig:net_processed} show graphical diagrams of the best trained two- and three-layer neural networks obtained for each of the initial and processed databases.

\begin{figure}[h!]
	\centering
	\begin{subfigure}{.5\linewidth}
		\centering
		\includegraphics[height=0.25\textheight]{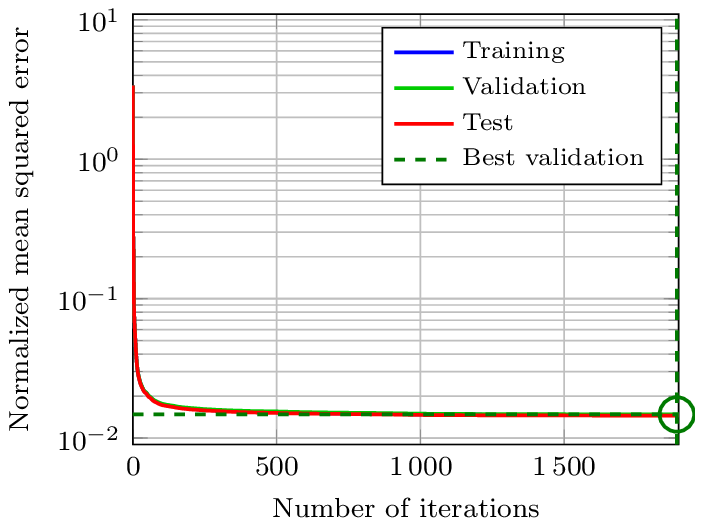}
		\caption{Two-layer neural network}
		\label{fig:perform_2Layers_initial}
	\end{subfigure}\hfill
	\begin{subfigure}{.5\linewidth}
		\centering
		\includegraphics[height=0.25\textheight]{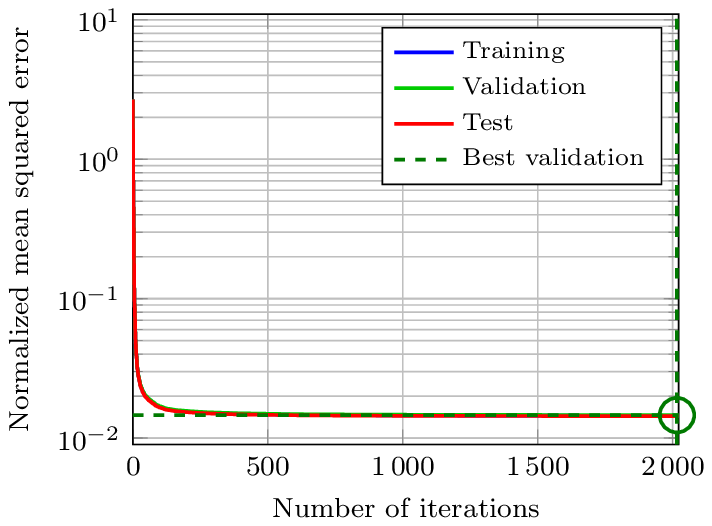}
		\caption{Three-layer neural network}
		\label{fig:perform_3Layers_initial}
	\end{subfigure}
	\caption{Initial database: evolution of the performance function (normalized mean squared error) with respect to the number of training iterations for the training (blue curve), validation (green curve) and test (red curve) datasets, with the best validation performance indicated by green dashed lines, for (\subref{fig:net_2Layers_initial}) the best two-layer neural network and (\subref{fig:net_3Layers_initial}) the best three-layer neural network}\label{fig:perform_initial}
\end{figure}

\begin{figure}[h!]
	\centering
	\begin{subfigure}{.5\linewidth}
		\centering
		\includegraphics[height=0.25\textheight]{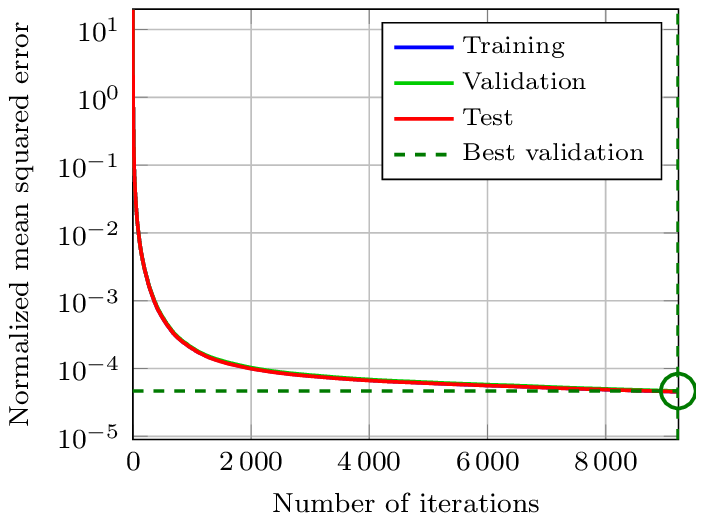}
		\caption{Two-layer neural network}
		\label{fig:perform_2Layers_processed}
	\end{subfigure}\hfill
	\begin{subfigure}{.5\linewidth}
		\centering
		\includegraphics[height=0.25\textheight]{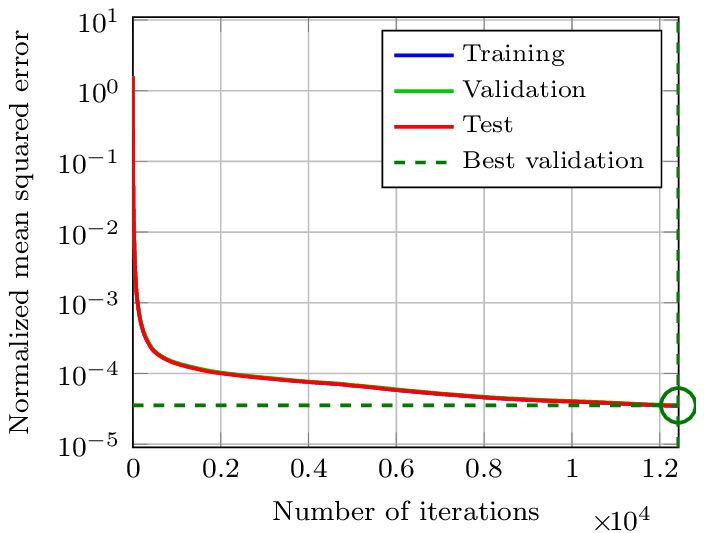}
		\caption{Three-layer neural network}
		\label{fig:perform_3Layers_processed}
	\end{subfigure}
	\caption{Processed database: evolution of the performance function (normalized mean squared error) with respect to the number of training iterations for the training (blue curve), validation (green curve) and test (red curve) datasets, with the best validation performance indicated by green dashed lines, for (\subref{fig:net_2Layers_processed}) the best two-layer neural network and (\subref{fig:net_3Layers_processed}) the best three-layer neural network}\label{fig:perform_processed}
\end{figure}

Figures~\ref{fig:perform_initial} and \ref{fig:perform_processed} show the graphs of the performance function (normalized mse) with respect to the number of training iterations for evaluating the training, validation and test performances of the best two- and three-layer trained neural networks (on the training, validation and test datasets, respectively) for each of the initial and processed databases. The normalized mse is plotted in a logarithmic scale. The network training continued until the validation performance function (normalized mean squared error on the validation dataset) failed to decrease after $6$ iterations (validation checks). For the initial (resp. processed) database, the validation performance function reached a minimum at iterations $1\,892$ and $2\,022$ (resp. $9\,233$ and $12\,423$) for the best two-layer and three-layer neural networks and the training continued for $6$ more iterations before it stopped. 
The normalized mean squared errors rapidly decrease during the first iterations and then slowly converge until validation stops for each of the training, validation and test datasets. The performance curves (normalized mse versus number of iterations) are similar for both validation and test datasets, indicating no significant overfitting occurred. The normalized mean squared errors obtained at final iteration, where the best validation performance occurs, are given in Table~\ref{tab:mse} for the training, validation, test and complete datasets and for the initial and processed databases. For each of the two numerical databases and each of the two multilayer neural networks, the network performances are similar for each of the training, validation, test and complete datasets. However, the network performances obtained with the processed database, which are around $10^{-5}$, are significantly better than that obtained with the initial database, which are around $10^{-2}$. Also, for each database, the best three-layer neural network shows slightly higher performances than the best two-layer neural network. Finally, the best overall network performance (normalized mse computed on the complete dataset) is obtained with the processed database and the three-layer neural network and is equal to $3.48\times 10^{-5}$, and the training, validation and test performances are equal to $3.47\times 10^{-5}$, $3.53\times 10^{-5}$ and $3.48\times 10^{-5}$, respectively.

\begin{table}[h!]
\caption{Normalized mean squared errors obtained for the best two-layer and three-layer neural networks trained from the initial and processed databases}
\label{tab:mse}
\centering
\begin{tabular}{|c|c|c|c|c|} \hline
 & \multicolumn{2}{c|}{Initial database} & \multicolumn{2}{c|}{Processed database} \\ \hline
\multirow{2}{*}{Dataset} & Two-layer & Three-layer & Two-layer & Three-layer \\
& neural network & neural network & neural network & neural network \\ \hline
Training & $1.46 \times 10^{-2}$ & $1.44 \times 10^{-2}$ & $4.55 \times 10^{-5}$ & $3.47 \times 10^{-5}$ \\ \hline
Validation & $1.48 \times 10^{-2}$ & $1.46 \times 10^{-2}$ & $4.66 \times 10^{-5}$ & $3.53 \times 10^{-5}$ \\ \hline
Test & $1.45 \times 10^{-2}$ & $1.44 \times 10^{-2}$ & $4.55 \times 10^{-5}$ & $3.48 \times 10^{-5}$ \\ \hline
All & $1.46 \times 10^{-2}$ & $1.45 \times 10^{-2}$ & $4.57 \times 10^{-5}$ & $3.48 \times 10^{-5}$ \\ \hline
\end{tabular}
\end{table}

\subsubsection{Second measure: linear regression between network outputs and targets}\label{sec:linear_regression}

\begin{figure}[h!]
	\centering
	\begin{subfigure}{.5\linewidth}
		\centering
		\includegraphics[width=1.0\linewidth]{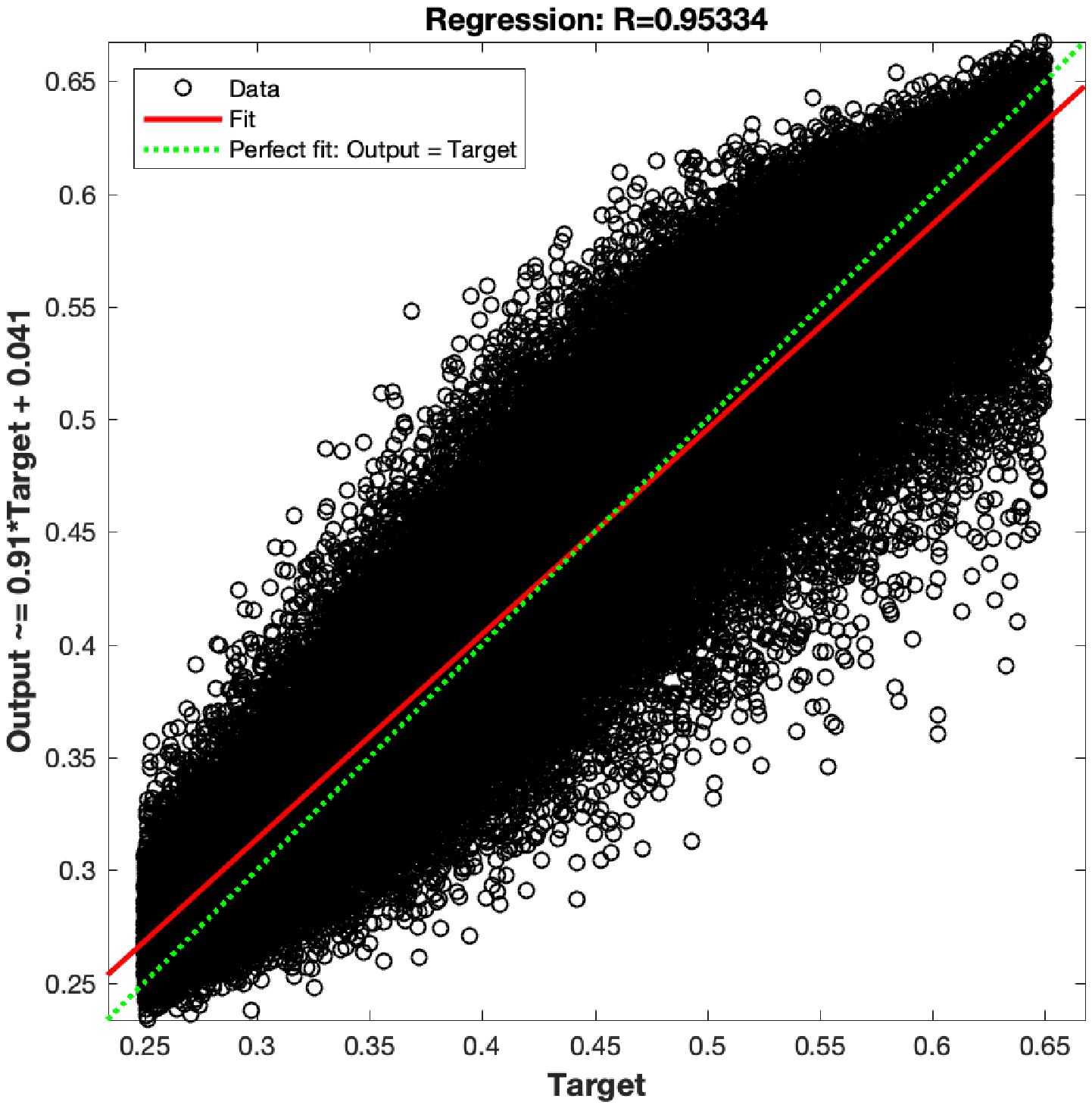}
		\caption{dispersion parameter $H_1$}
		\label{fig:regression_delta_3Layers_initial}
	\end{subfigure}\hfill
	\begin{subfigure}{.5\linewidth}
		\centering
		\includegraphics[width=1.0\linewidth]{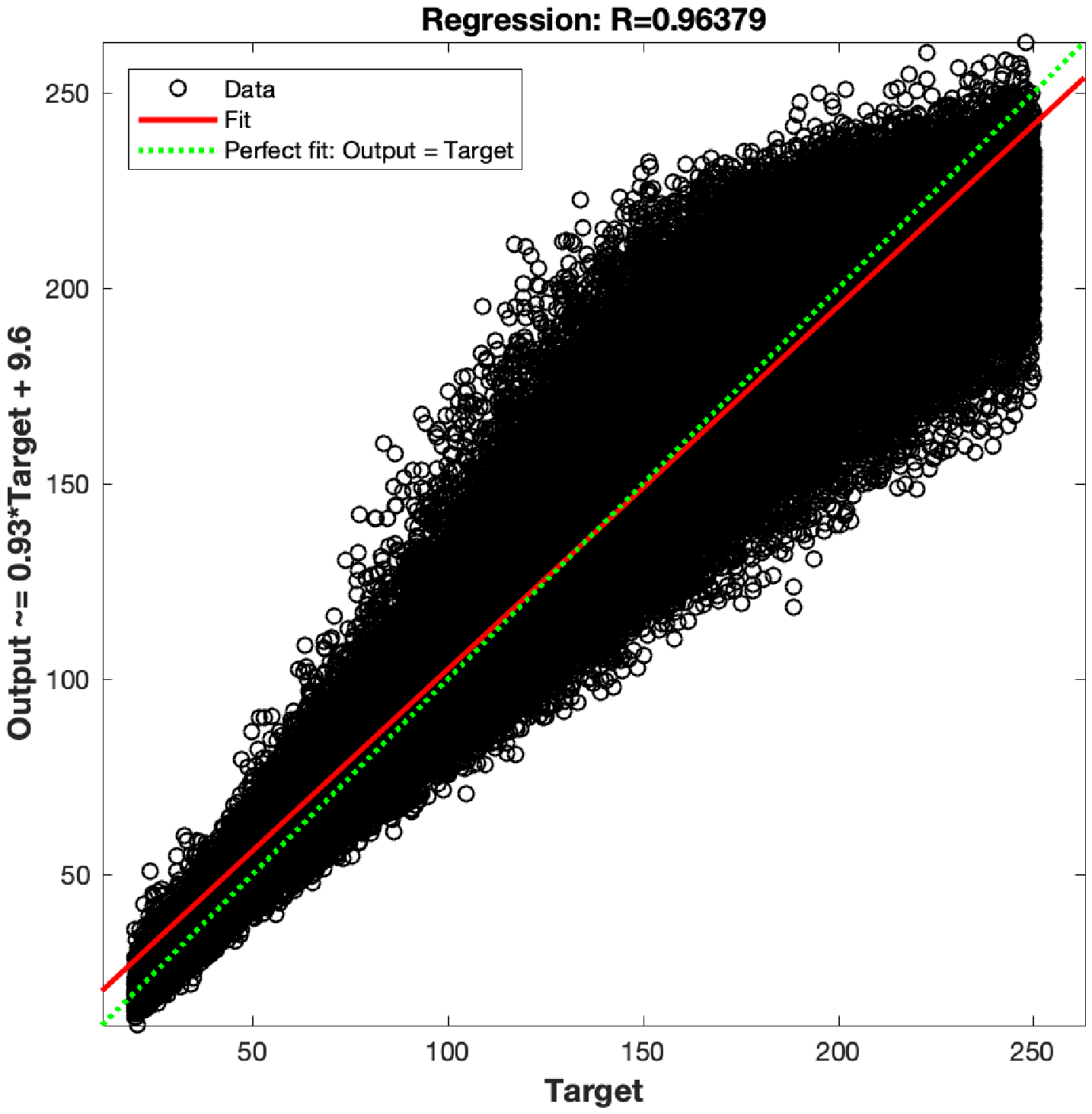}
		\caption{spatial correlation length $H_2$ [$\mu$m]}
		\label{fig:regression_lcorr_3Layers_initial}
	\end{subfigure}\\
	\begin{subfigure}{.5\linewidth}
		\centering
		\includegraphics[width=1.0\linewidth]{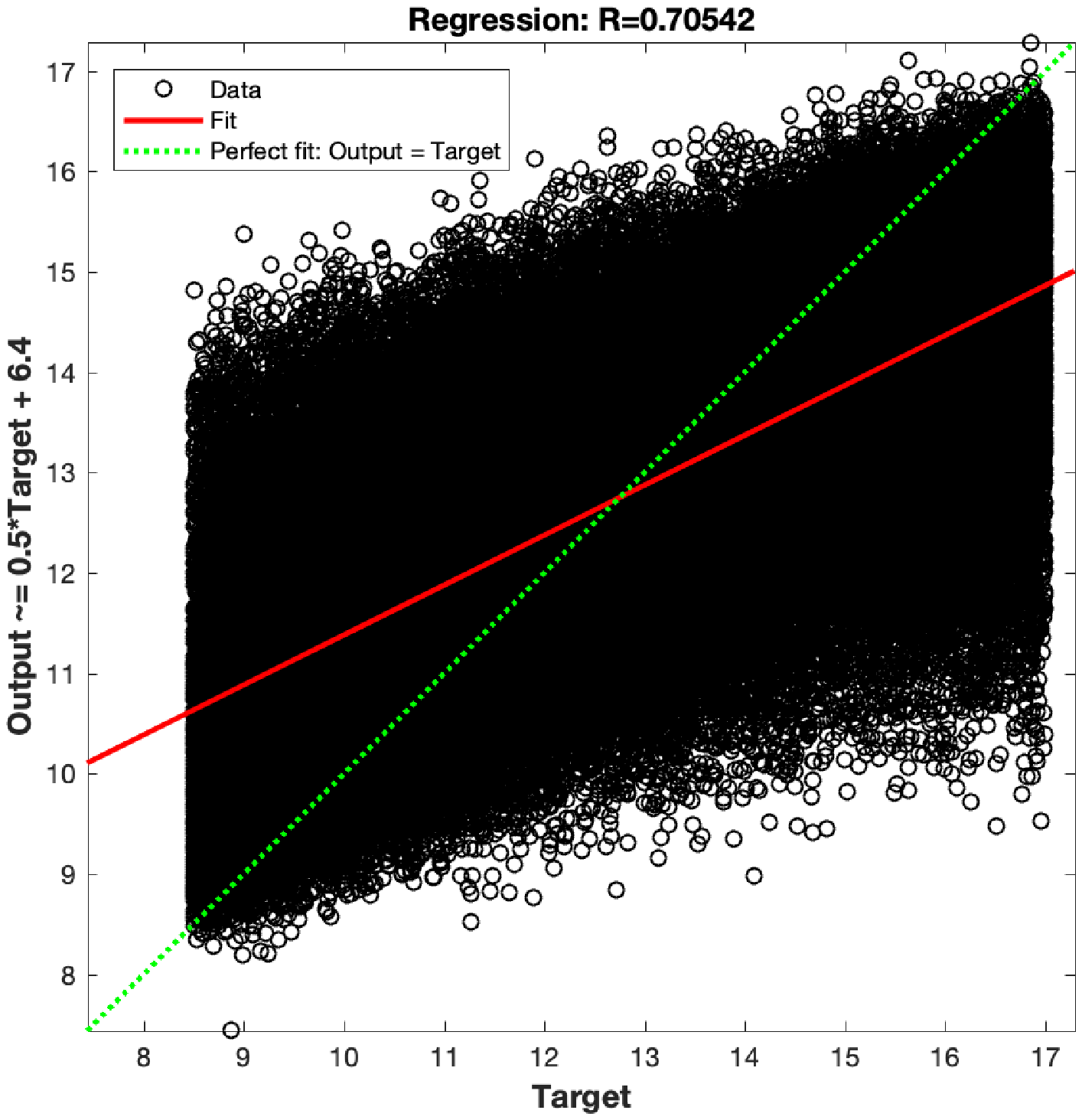}
		\caption{mean bulk modulus $H_3$ [GPa]}
		\label{fig:regression_kappa_3Layers_initial}
	\end{subfigure}\hfill
	\begin{subfigure}{.5\linewidth}
		\centering
		\includegraphics[width=1.0\linewidth]{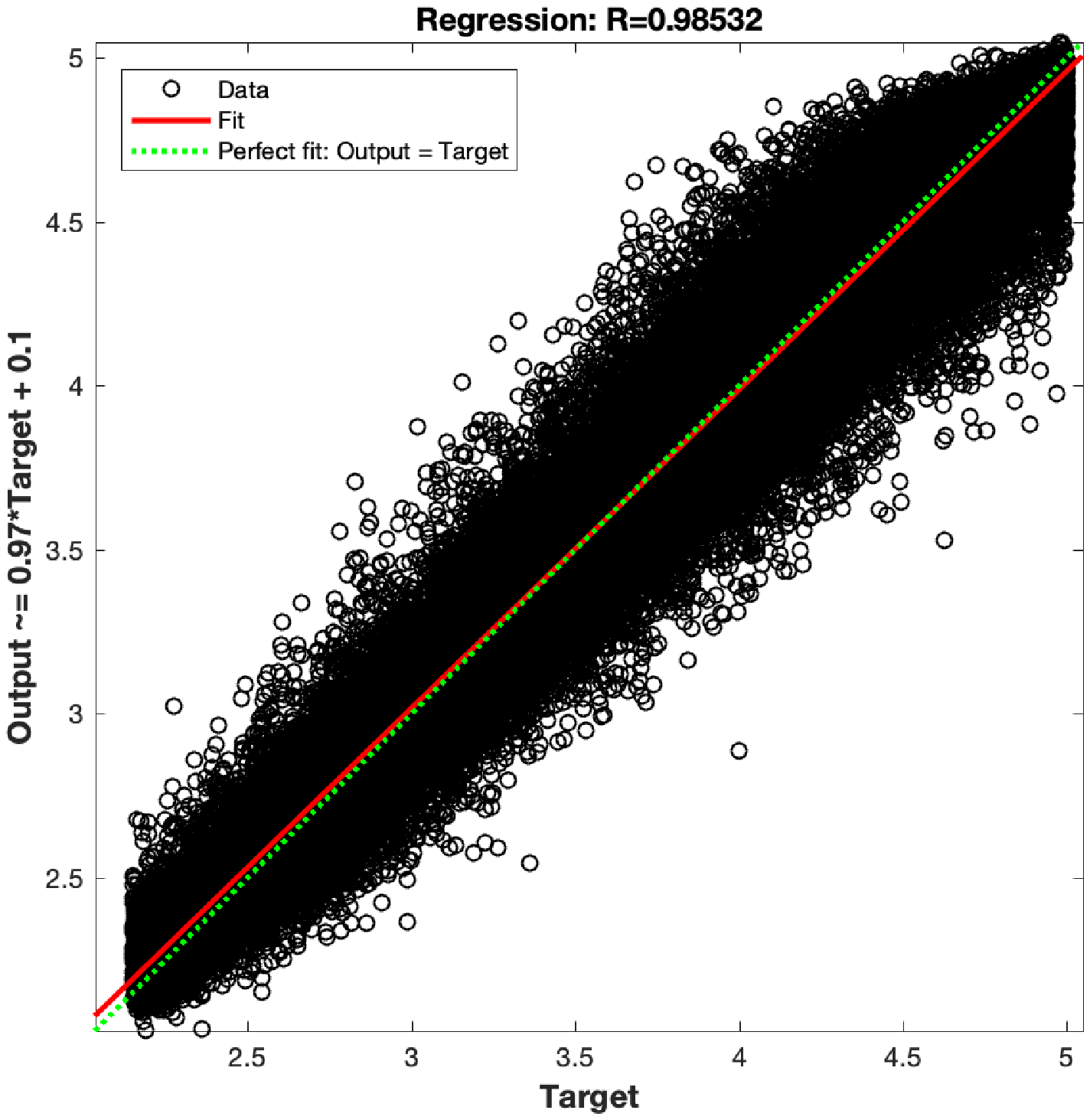}
		\caption{mean shear modulus $H_4$ [GPa]}
		\label{fig:regression_mu_3Layers_initial}
	\end{subfigure}
	\caption{Initial database and three-layer neural network: linear regression between network outputs and corresponding targets for each random hyperparameter $H_1$, $H_2$, $H_3$ and $H_4$, for the 
	complete 
	dataset. In each plot, the network outputs and targets are represented by open black circles, the perfect fit (outputs exactly equal to targets) is represented by a dashed green line, and the best linear fit (linear regression between outputs and targets) is represented by a solid red line for the 
	complete 
	dataset. The regression value ($R$-value) is given at the top of each regression plot}\label{fig:regression_3Layers_initial}
\end{figure}

\begin{figure}[h!]
	\centering
	\begin{subfigure}{.5\linewidth}
		\centering
		\includegraphics[width=1.0\linewidth]{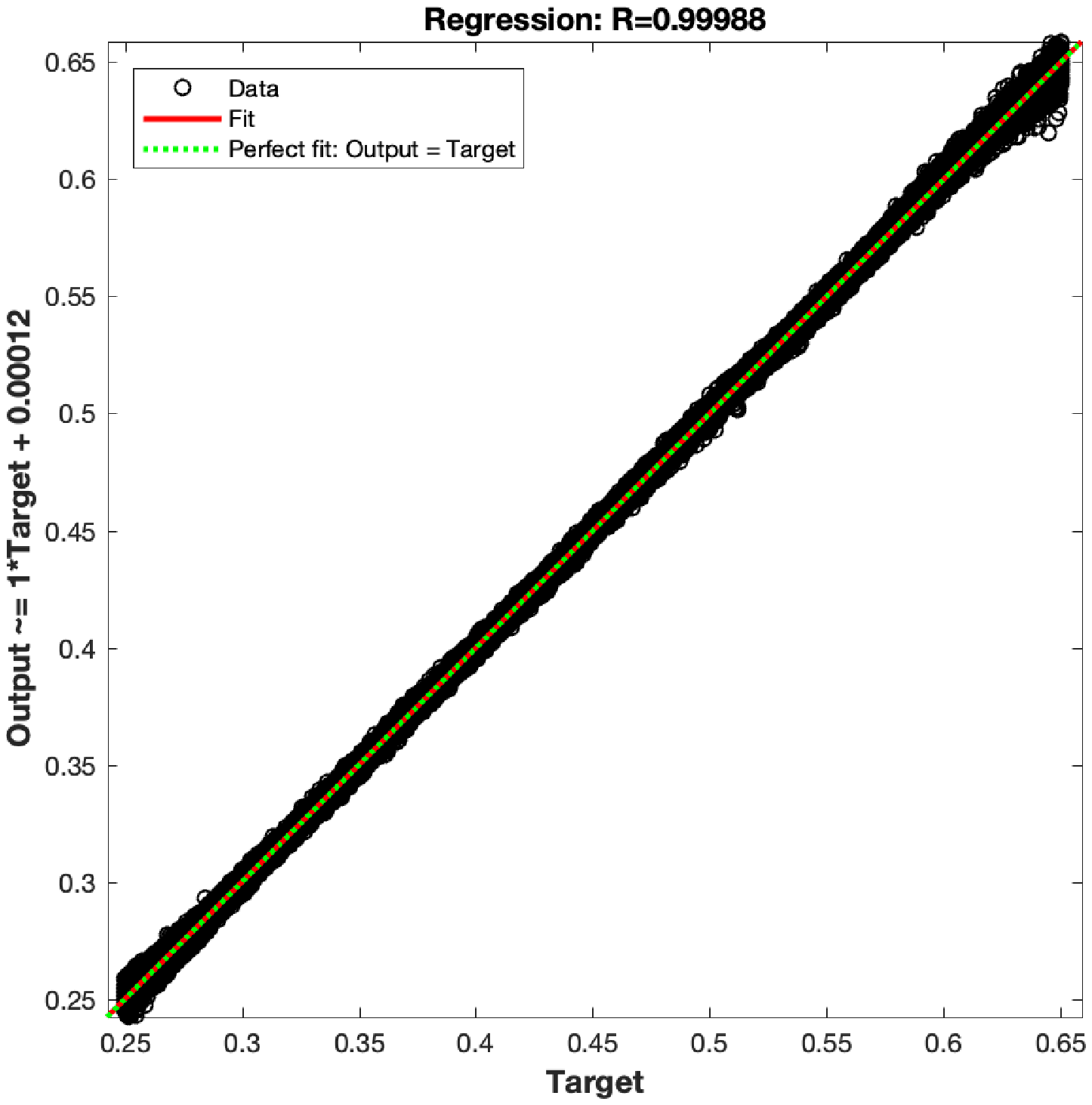}
		\caption{dispersion parameter $H_1$}
		\label{fig:regression_delta_3Layers_processed}
	\end{subfigure}\hfill
	\begin{subfigure}{.5\linewidth}
		\centering
		\includegraphics[width=1.0\linewidth]{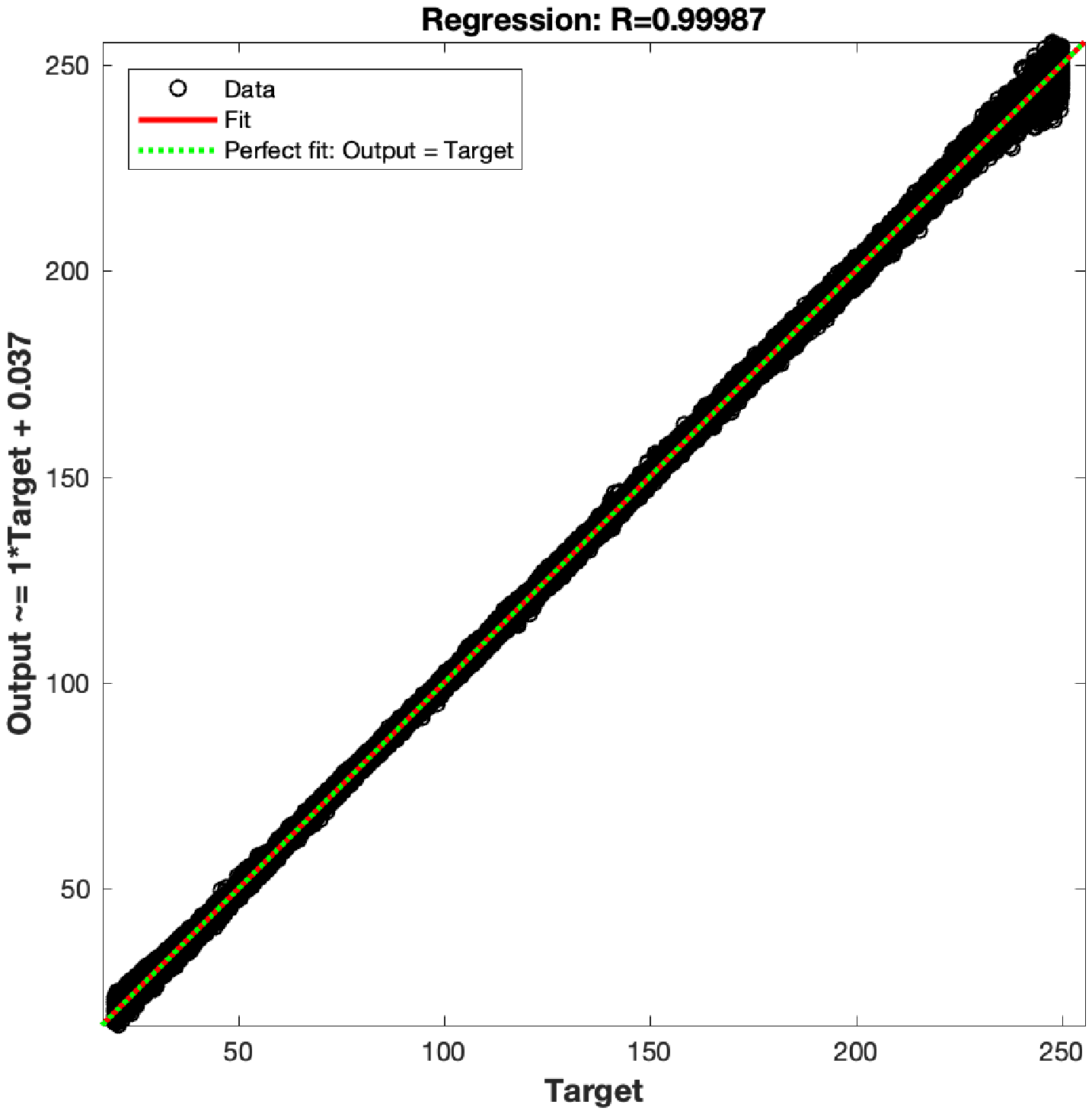}
		\caption{spatial correlation length $H_2$ [$\mu$m]}
		\label{fig:regression_lcorr_3Layers_processed}
	\end{subfigure}\\
	\begin{subfigure}{.5\linewidth}
		\centering
		\includegraphics[width=1.0\linewidth]{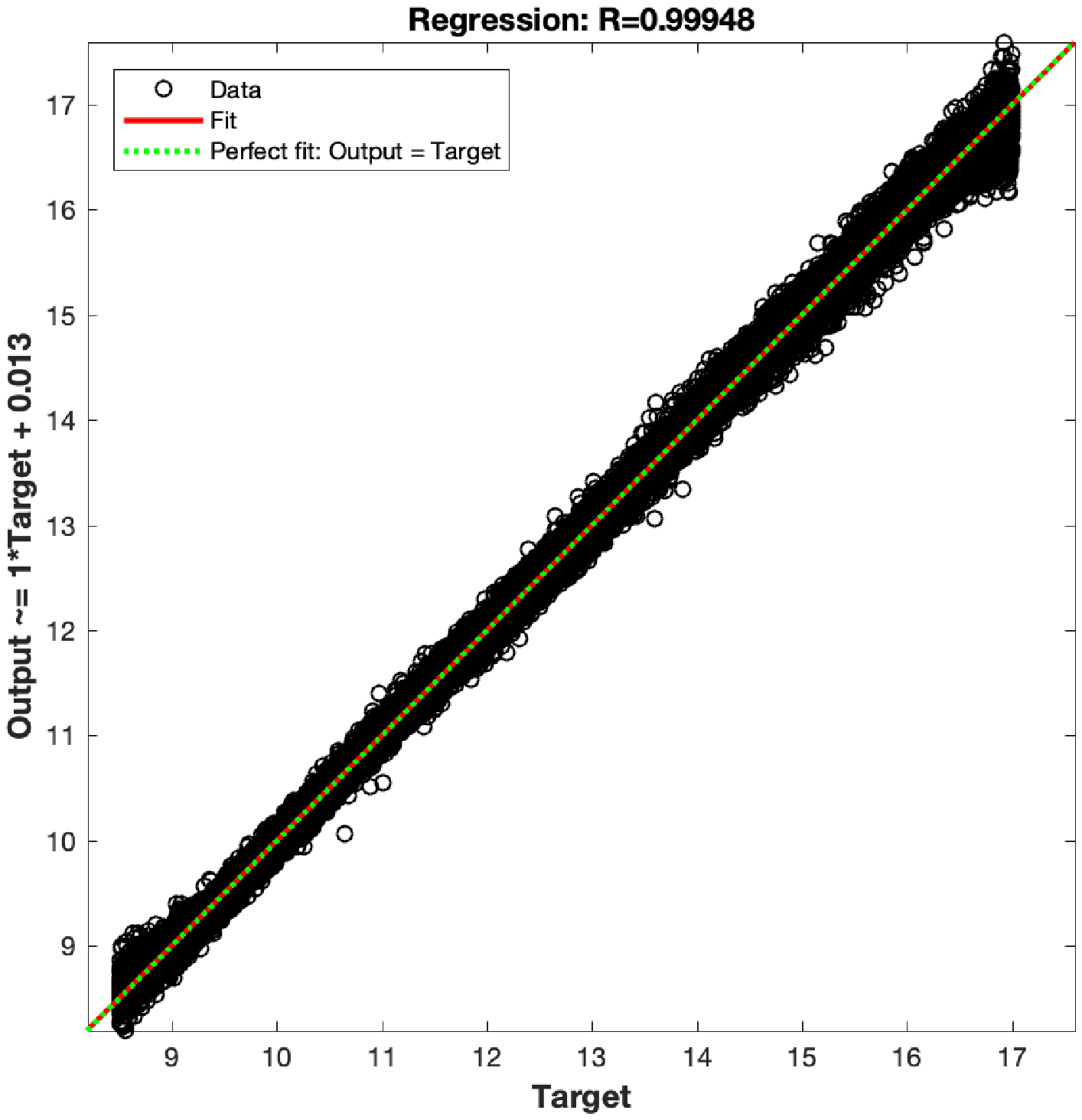}
		\caption{mean bulk modulus $H_3$ [GPa]}
		\label{fig:regression_kappa_3Layers_processed}
	\end{subfigure}\hfill
	\begin{subfigure}{.5\linewidth}
		\centering
		\includegraphics[width=1.0\linewidth]{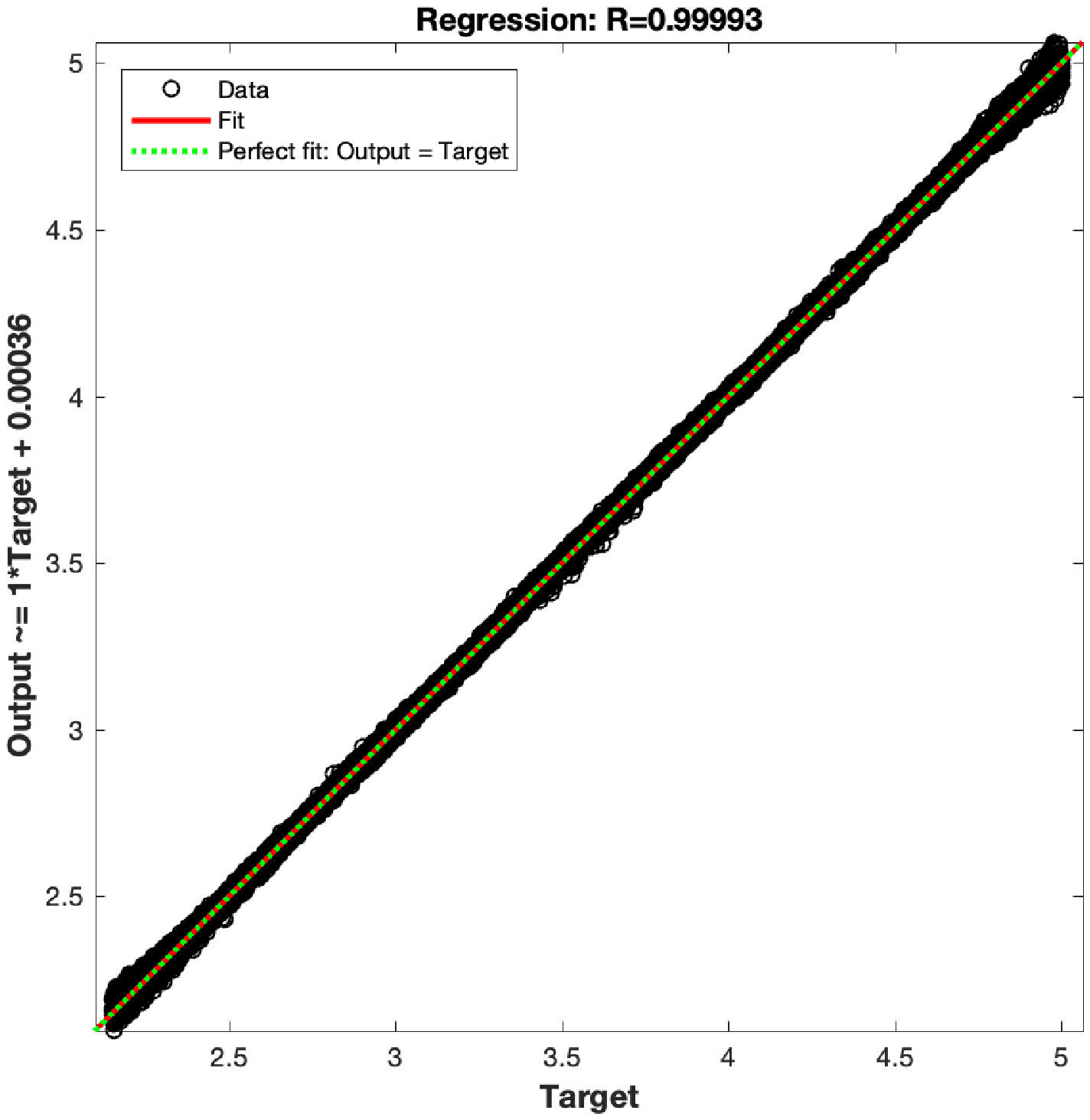}
		\caption{mean shear modulus $H_4$ [GPa]}
		\label{fig:regression_mu_3Layers_processed}
	\end{subfigure}
	\caption{Processed database and three-layer neural network: linear regression between network outputs and corresponding targets for each random hyperparameter $H_1$, $H_2$, $H_3$ and $H_4$, for the 
	complete 
	dataset. In each plot, the network outputs and targets are represented by open black circles, the perfect fit (outputs exactly equal to targets) is represented by a dashed green line, and the best linear fit (linear regression between outputs and targets) is represented by a solid red line for the 
	complete 
	dataset. The regression value ($R$-value) is given at the top of each regression plot}\label{fig:regression_3Layers_processed}
\end{figure}

As a second evaluation of the network performance, the linear regression of the network outputs relative to targets is plotted across 
the complete 
dataset in Figures~\ref{fig:regression_3Layers_initial} and \ref{fig:regression_3Layers_processed} for the initial and processed databases, respectively, and for the best three-layer neural network. Note that very similar regression plots have been obtained for the best two-layer neural network and are not reported here for the sake of brevity. 
For both numerical databases (initial and processed), the trained network output vectors have been computed for all the input vectors in the complete dataset, then the output and target vectors belonging to each of the training, validation and test subsets have been extracted, and finally the network outputs have been plotted with respect to targets for each of the training, validation and test subsets as well as for the complete dataset. Very similar trends have been observed for the complete dataset and for each of the training, validation, test data subsets separately, so that only the results for the complete dataset are displayed in Figures~\ref{fig:regression_3Layers_initial} and \ref{fig:regression_3Layers_processed} for the sake of simplicity and conciseness. On the one hand, for the initial database, the best linear fit between network outputs and corresponding targets, although not perfect, is fairly good for the complete dataset (and for each data subset) with regression values ($R$-values) over $0.95$ for dispersion parameter $H_1$, $0.96$ for spatial correlation length $H_2$, $0.70$ for mean bulk modulus $H_3$, and $0.98$ for mean shear modulus $H_4$. Nevertheless, the scatter plots of the network outputs and corresponding targets are highly dispersed and show that some data points in the 
dataset have poor fits, especially for $H_4$. Such a large dispersion is due to the stochastic nature of the non-linear mapping defined in \eqref{randomQoI} between random vector $\Qb$ of quantities of interest and random vector $\Hb$ of hyperparameters. On the other hand, for the processed database, the best linear fit between network outputs and corresponding targets is almost perfect for the complete dataset (and for each data subset) with regression values ($R$-values) very close to $1$ for all components $H_1$, $H_2$, $H_3$ and $H_4$ of $\Hb$ (and for all data subsets). The network outputs track the targets with very small dispersion for each network output $H_1$, $H_2$, $H_3$ and $H_4$, showing a significantly better fit for the processed database than for the initial database.

\subsubsection{Third measure: marginal probability density functions of the components of the output random vector}\label{sec:marginal_pdfs}

\begin{figure}[h!]
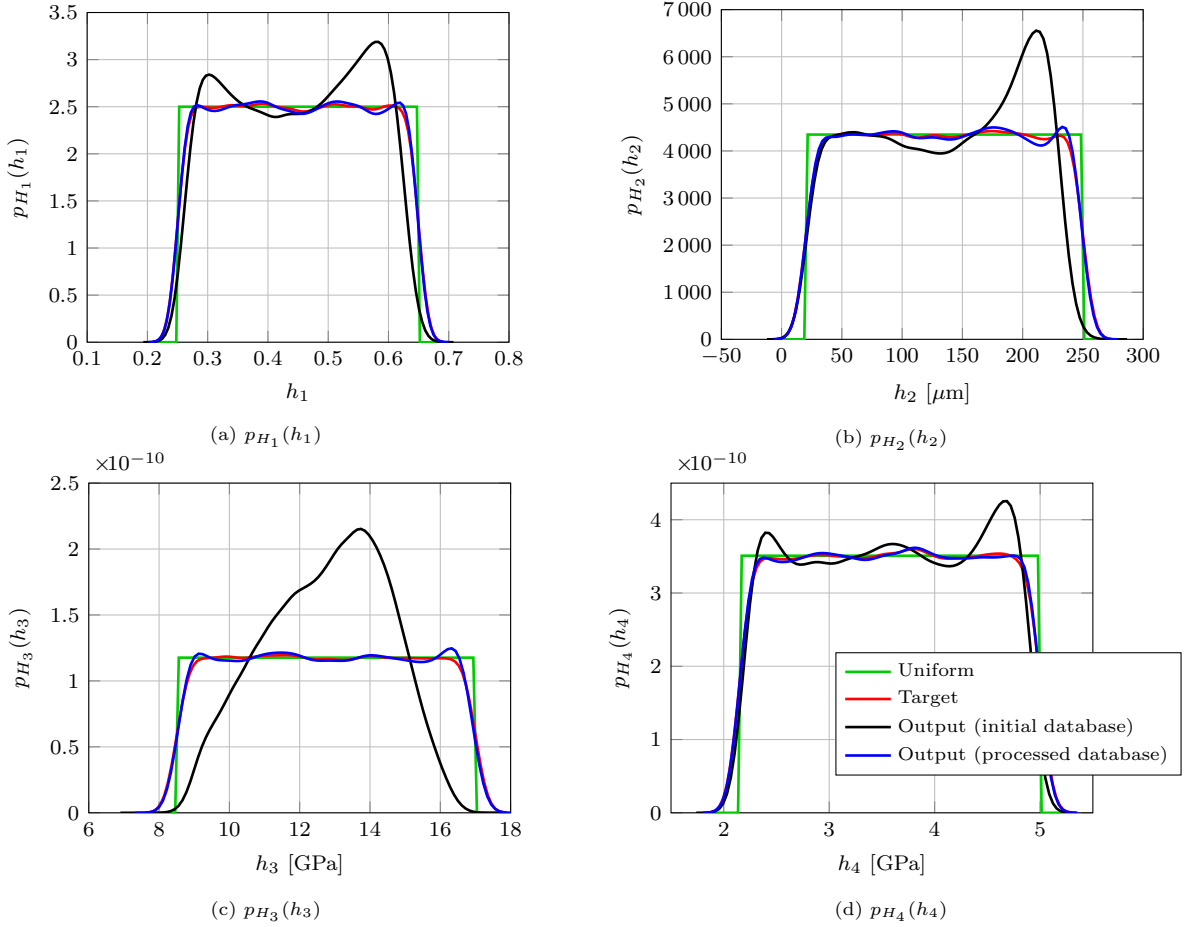

	\centering
	\begin{subfigure}{.5\linewidth}
		\centering
		\tikzsetnextfilename{pdf_delta}
		\input{pdf_delta}
		\caption{$p_{H_1}(h_1)$}
		\label{fig:pdf_delta}
	\end{subfigure}\hfill
	\begin{subfigure}{.5\linewidth}
		\centering
		\tikzsetnextfilename{pdf_lcorr}
		\input{pdf_lcorr}
		\caption{$p_{H_2}(h_2)$}
		\label{fig:pdf_lcorr}
	\end{subfigure}\\
	\begin{subfigure}{.5\linewidth}
		\centering
		\tikzsetnextfilename{pdf_kappa}
		\input{pdf_kappa}
		\caption{$p_{H_3}(h_3)$}
		\label{fig:pdf_kappa}
	\end{subfigure}\hfill
	\begin{subfigure}{.5\linewidth}
		\centering
		\tikzsetnextfilename{pdf_mu}
		\input{pdf_mu}
		\caption{$p_{H_4}(h_4)$}
		\label{fig:pdf_mu}
	\end{subfigure}
	\caption{Probability density functions (pdfs) $p_{H_1}$, $p_{H_2}$, $p_{H_3}$ and $p_{H_4}$ of random variables $H_1$, $H_2$, $H_3$ and $H_4$, respectively, with the uniform target pdfs (green), the estimated target pdfs (red), and the estimated output pdfs computed by using the best neural network trained with the initial database (black) and with the processed database (blue)}
	\label{fig:pdf}
\end{figure}

As a third evaluation of the network performance, the marginal pdfs $p_{H_1}$, $p_{H_2}$, $p_{H_3}$ and $p_{H_4}$ of each component of random vector $\Hb = (H_1,H_2,H_3,H_4)$ of hyperparameters, which are assumed to be uniform random variables, are estimated by using the univariate Gaussian kernel density estimation method \cite{Bow97} with the $N_d$ network output data obtained from the initial (resp. processed) database with the best three-layer neural network, and compared to the uniform target pdfs and to the target pdfs estimated by using the univariate Gaussian kernel density estimation method with the $N_d$ associated target data in Figure~\ref{fig:pdf}. The output pdfs constructed from the output vectors of the best neural network trained with the processed database perfectly match the associated target pdfs, while the output pdfs constructed from the output vectors of the best neural network trained with the initial database have a worse fit, especially for $H_3$.

\subsubsection{Discussion}\label{sec:discussion}

According to the aforementioned results concerning the network performances, the neural network trained with the processed database, that is to say obtained by conditioning the network input vectors contained in the initial database with respect to the network target vectors, can directly be used for identifying the value $\hb^{\text{out}}$ of random hyperparameters $\Hb$ corresponding to a given observed vector $\qb^{\text{obs}}$ of quantities of interest. The conditioning of the initial database then appears to be a determining key factor in obtaining an efficient trained neural network. For a given input vector $\qb^{\text{obs}}$, the output vector $\hb^{\text{out}}$ computed by the best neural network trained with the processed database corresponds to the solution $\hb^{\ast}$ of the statistical inverse problem formulated in Section~\ref{sec:inverse_problem}. Finally, computing network output vector $\hb^{\text{out}}$ for any network input vector $\qb^{\text{obs}}$ allows for defining a deterministic nonlinear mapping $\Ncb$ defined from $\Rbb^n$ into $\Rbb^m$ as
\begin{equation}\label{ANN_mapping}
\hb^{\text{out}} = \Ncb(\qb^{\text{obs}}).
\end{equation}

\section{Robust solution of the statistical inverse problem}
\label{sec:robustness}

In order to assess the robustness of the proposed identification method by taking into account experimental errors (measurement errors and epistemic uncertainties) on the input vector $\qb^{\text{obs}} = (\delta^{\epsilonb}_{\text{obs}},\ell^{\epsilonb}_{\text{obs},1},\ell^{\epsilonb}_{\text{obs},2},\ellb^{\text{eff}}_{\text{obs}})$ of quantities of interest, this latter can be considered and modeled as a random vector $\Qb^{\text{obs}} = (D^{\epsilonb}_{\text{obs}},L^{\epsilonb}_{\text{obs},1},L^{\epsilonb}_{\text{obs},2},\Lb^{\text{eff}}_{\text{obs}})$ where $\Lb^{\text{eff}}_{\text{obs}} = (\log([\Lb^{\text{eff}}_{\text{obs}}]_{11}),[\Lb^{\text{eff}}_{\text{obs}}]_{12},[\Lb^{\text{eff}}_{\text{obs}}]_{13},\log([\Lb^{\text{eff}}_{\text{obs}}]_{22}),[\Lb^{\text{eff}}_{\text{obs}}]_{23},\log([\Lb^{\text{eff}}_{\text{obs}}]_{33}))$ is the random vector with values in $\Rbb^6$ whose components are deduced from the $6$ components of the invertible upper triangular random matrix $[\Lb^{\text{eff}}_{\text{obs}}]$ with values in $\Mbb_3^+(\Rbb)$ (with positive-valued diagonal entries $[\Lb^{\text{eff}}_{\text{obs}}]_{11}$, $[\Lb^{\text{eff}}_{\text{obs}}]_{22}$ and $[\Lb^{\text{eff}}_{\text{obs}}]_{33}$) resulting from the Cholesky factorization of the random compliance matrix $[\Sb^{\text{eff}}_{\text{obs}}]$ with values in $\Mbb_3^+(\Rbb)$, that is $[\Sb^{\text{eff}}_{\text{obs}}] = [\Lb^{\text{eff}}_{\text{obs}}]^T [\Lb^{\text{eff}}_{\text{obs}}]$. 
The prior probabilistic model of $\Qb^{\text{obs}}$ is constructed by having recourse to the MaxEnt principle \cite{Jay57a,Jay57b,Sob90,Kap92,Jum00,Jay03,Cov06,Soi17a} based on the following algebraically independent constraints to be satisfied: (i) $D^{\epsilonb}_{\text{obs}}$, $L^{\epsilonb}_{\text{obs},1}$, $L^{\epsilonb}_{\text{obs},2}$ and $\Lb^{\text{eff}}_{\text{obs}}$ are mutually statistically independent random variables, (ii) $D^{\epsilonb}_{\text{obs}}$, $L^{\epsilonb}_{\text{obs},1}$ and $L^{\epsilonb}_{\text{obs},2}$ are a.s. $\Rbb^+$-valued random variables for which the values are unlikely close to zero and consequently $\Eb\{\log(D^{\epsilonb}_{\text{obs}})\}$, $\Eb\{\log(L^{\epsilonb}_{\text{obs}, 1})\}$ and $\Eb\{\log(L^{\epsilonb}_{\text{obs}, 2})\}$ are finite, (iii) the mean values $\Ebb\{D^{\epsilonb}_{\text{obs}}\}$, $\Ebb\{L^{\epsilonb}_{\text{obs},1}\}$ and $\Ebb\{L^{\epsilonb}_{\text{obs},2}\}$ of $D^{\epsilonb}_{\text{obs}}$, $L^{\epsilonb}_{\text{obs},1}$ and $L^{\epsilonb}_{\text{obs},2}$ are known and given by $\delta^{\epsilonb}_{\text{obs}}$, $\ell^{\epsilonb}_{\text{obs},1}$ and $\ell^{\epsilonb}_{\text{obs},2}$, respectively, that are $\Ebb\{D^{\epsilonb}_{\text{obs}}\} = \delta^{\epsilonb}_{\text{obs}}$, $\Ebb\{L^{\epsilonb}_{\text{obs},1}\} = \ell^{\epsilonb}_{\text{obs},1}$ and $\Ebb\{L^{\epsilonb}_{\text{obs},2}\} = \ell^{\epsilonb}_{\text{obs},2}$, (iv) $[\Sb^{\text{eff}}_{\text{obs}}]$ is a second-order a.s. positive-definite symmetric real-valued random matrix whose mean value $\Ebb\{[\Sb^{\text{eff}}_{\text{obs}}]\}$ is known and given by the positive-definite symmetric real matrix $[\underline{S}^{\text{eff}}_{\text{obs}}]$, that is $\Ebb\{[\Sb^{\text{eff}}_{\text{obs}}]\} = [\underline{S}^{\text{eff}}_{\text{obs}}]$, and whose inverse matrix $[\Sb^{\text{eff}}_{\text{obs}}]^{-1}$ is a second-order random matrix. Then, the MaxEnt principle leads to statistically independent gamma (positive-valued) random variables $D^{\epsilonb}_{\text{obs}}$, $L^{\epsilonb}_{\text{obs},1}$ and $L^{\epsilonb}_{\text{obs},2}$ whose mean values are known and given by $\delta^{\epsilonb}_{\text{obs}}$, $\ell^{\epsilonb}_{\text{obs},1}$ and $\ell^{\epsilonb}_{\text{obs},2}$, respectively, and whose levels of statistical fluctuations are given by the unknown positive dispersion parameters $s_0$, $s_1$ and $s_2$, respectively, corresponding to the coefficients of variation of each of the random variables $D^{\epsilonb}_{\text{obs}}$, $L^{\epsilonb}_{\text{obs},1}$ and $L^{\epsilonb}_{\text{obs},2}$
.
As a consequence, dispersion parameters $s_0$, $s_1$ and $s_2$ directly control the level of uncertainties of the $3$ first random quantities of interest $Q^{\text{obs}}_1 = D^{\epsilonb}_{\text{obs}}$, $Q^{\text{obs}}_2 = L^{\epsilonb}_{\text{obs},1}$ and $Q^{\text{obs}}_3 = L^{\epsilonb}_{\text{obs},2}$
, respectively. 
The MaxEnt principle also leads to a random matrix $[\Sb^{\text{eff}}_{\text{obs}}]$ in the ensemble SE$_0^+$ of non-Gaussian second-order a.s. positive-definite symmetric real random matrices \cite{Soi05a,Soi16a,Soi17a} with a known mean value given by $[\underline{S}^{\text{eff}}_{\text{obs}}]$ and parameterized by an unknown positive dispersion parameter $s^{\text{eff}}$ that is proportional to the coefficient of variation of $[\Sb^{\text{eff}}_{\text{obs}}]$ (see \cite{Soi06,Soi08a,Soi17a}), thus characterizing the level of statistical fluctuations of $[\Sb^{\text{eff}}_{\text{obs}}]$ around its mean value $[\underline{S}^{\text{eff}}_{\text{obs}}]$. As a consequence, dispersion parameter $s^{\text{eff}}$ indirectly controls the level of uncertainties of the $6$ last random quantities of interest $Q^{\text{obs}}_4 = \log([\Lb^{\text{eff}}_{\text{obs}}]_{11})$, $Q^{\text{obs}}_5 = [\Lb^{\text{eff}}_{\text{obs}}]_{12}$, $Q^{\text{obs}}_6 = [\Lb^{\text{eff}}_{\text{obs}}]_{13}$, $Q^{\text{obs}}_7 = \log([\Lb^{\text{eff}}_{\text{obs}}]_{22})$, $Q^{\text{obs}}_8 = [\Lb^{\text{eff}}_{\text{obs}}]_{23}$ and $Q^{\text{obs}}_9 = \log([\Lb^{\text{eff}}_{\text{obs}}]_{33})$.
Finally, the prior probabilistic model of $\Qb^{\text{obs}}$ depends on the four-dimensional vector-valued hyperparameter $\ssb = (s_0,s_1,s_2,s^{\text{eff}}) \in \intervaloo{0}{+\infty}^{4}$ allowing the level of statistical fluctuations of random vector $\Qb^{\text{obs}}$ to be controlled. If experimental data are available on input random vector $\Qb^{\text{obs}}$, then an estimation of $\ssb$ can be carried out, \eg{} by using the least-squares method \cite{Law95,Soi17a} or the maximum likelihood estimation method \cite{Ser80,Pap02,Spa05a,Soi17a}. If only one input vector $\qb^{\text{obs}}$ of observed quantities of interest is available, then $\ssb$ can be used to perform a sensitivity analysis of the network output value $\hb^{\text{out}}$ of hyperparameters. It is worth pointing out that the fundamental problem related to the identification of $\ssb$ is a challenging task that falls out of the scope of the present paper and is therefore left for future works. In the numerical examples presented in Sections~\ref{sec:results_synthetic_data} and \ref{sec:results_real_data}, the robustness analysis of the network output with respect to $\ssb$ has been performed by considering the same value $s$ for each component of $\ssb$, that is $s_0 = s_1 = s_2 = s^{\text{eff}} = s$, and for different values of $s$ arbitrarily chosen between $1\%$ and $5\%$ in order to provide a simple illustration of the proposed methodology. Such a probabilistic model of the random vector $\Qb^{\text{obs}}$ of observed quantities of interest then allows the robustness of the output value $\hb^{\text{out}}$ of hyperparameters to be analyzed with respect to the level of statistical fluctuations of $\Qb^{\text{obs}}$ controlled by $\ssb$. The network output $\hb^{\text{out}}$ is then modeled as a random variable $\Hb^{\text{out}} = (H_1^{\text{out}},H_2^{\text{out}},H_3^{\text{out}},H_4^{\text{out}})$ defined in using \eqref{ANN_mapping} and such that 
\begin{equation}\label{Hout_random}
\Hb^{\text{out}} = \Ncb(\Qb^{\text{obs}}).
\end{equation}
Let $\qb^{\text{obs},(1)},\dots,\qb^{\text{obs},(N_s)}$ be $N_s$ independent realizations of $\Qb^{\text{obs}}$, then $N_s$ independent realizations $\hb^{\text{out},(1)},\dots,\hb^{\text{out},(N_s)}$ of $\Hb^{\text{out}}$ are constructed in using \eqref{Hout_random} and we then have $\hb^{\text{out},(i)} = \Ncb(\qb^{\text{obs},(i)})$ for $i=1,\dots,N_s$. For identification purposes in the presence of experimental errors, the solution $\hb^{\ast}$ of the underlying statistical inverse problem can be defined as the output mean value $\underline{\hb}^{\text{out}} = \Ebb\{\Hb^{\text{out}}\}$ estimated in using the $N_s$ independent realizations $\hb^{\text{out},(1)},\dots,\hb^{\text{out},(N_s)}$ of $\Hb^{\text{out}}$ with the mathematical statistics \cite{Ser80}. In addition, the network outputs can be used for constructing the marginal probability density functions $p_{H_1^{\text{out}}}$, $p_{H_2^{\text{out}}}$, $p_{H_3^{\text{out}}}$ and $p_{H_4^{\text{out}}}$ of the components of output random vector $\Hb^{\text{out}} = (H_1^{\text{out}},H_2^{\text{out}},H_3^{\text{out}},H_4^{\text{out}})$ by using the univariate Gaussian kernel density estimation method \cite{Bow97} in order to quantify the robustness of the output vectors of hyperparameters generated by the trained neural network with respect to some experimental errors on the input vector of observed quantities of interest.

\section{Numerical example on synthetic data}\label{sec:results_synthetic_data}

Based on the numerical results obtained in Section~\ref{sec:test_network_performance}, we consider the best three-layer neural network trained with the processed database for identification purposes. Hereinafter, the neural network-based identification method is first applied to synthetic data coming from numerical simulations and then carried out on real experimental data coming from experimental measurements on a bovine cortical bone specimen in Section~\ref{sec:results_real_data}.

We first consider a given input vector $\qb^{\text{obs}}$ of quantities of interest contained in the test dataset for validating the proposed neural network-based identification method. The network output vector $\hb^{\text{out}}$ is directly computed by using the trained neural network with $\qb^{\text{obs}}$ as input vector and compared to the corresponding target vector $\hb^{\text{target}}$. For analyzing the robustness of the output vector $\hb^{\text{out}}$ of hyperparameters with respect to the uncertainties on the input vector $\qb^{\text{obs}}$ of observed quantities of interest, $N_s = 10^6$ independent realizations $\qb^{\text{obs},(1)},\dots,\qb^{\text{obs},(N_s)}$ of input random vector $\Qb^{\text{obs}}$ are generated according to its probabilistic model presented in Section~\ref{sec:robustness} and parameterized by the vector-valued parameter $\ssb=(s_0,s_1,s_2,s^{\text{eff}})$ controlling the level of statistical fluctuations of $\Qb^{\text{obs}}$ around its mean value $\qb^{\text{obs}}$. The best trained neural network is then used for simulating the corresponding $N_s$ independent realizations $\hb^{\text{out},(1)},\dots,\hb^{\text{out},(N_s)}$ of output random vector $\Hb^{\text{out}}$, from which the mean value $\underline{\hb}^{\text{out}} = \Ebb\{\Hb^{\text{out}}\}$ and the confidence interval $I^{\text{out}}$ with a probability level $95\%$ of $\Hb^{\text{out}}$ are estimated by using the mathematical statistics. In order to quantify the robustness of the best trained neural network with respect to the uncertainties on input vector $\qb^{\text{obs}}$, we consider the same input uncertainty level $s$ for each of the components of $\ssb$, that is $s_0 = s_1 = s_2 = s^{\text{eff}} = s$, and we perform a robustness analysis of the network output mean vector $\underline{\hb}^{\text{out}}$ with respect to the input uncertainty level $s$ by considering increasing values for $s \in \set{0.01,0.02,0.03,0.04,0.05}$. Recall that $\underline{\hb}^{\text{out}}$ \emph{a priori} coincides with $\hb^{\text{out}}$ only when $s=0$ (\ie{} in the absence of uncertainties on input vector $\qb^{\text{obs}}$). According to Section~\ref{sec:robustness}, the value of $s$ corresponds to the coefficient of variation of each of the $3$ first components $Q^{\text{obs}}_1 = D^{\epsilonb}_{\text{obs}}$, $Q^{\text{obs}}_2 = L^{\epsilonb}_{\text{obs},1}$ and $Q^{\text{obs}}_3 = L^{\epsilonb}_{\text{obs},2}$ of random vector $\Qb^{\text{obs}}$ and therefore allows the level of statistical fluctuations of these input random variables (around their respective mean values) to be directly controlled. Also, the value of $s$ is proportional to the coefficient of variation of $[\Sb^{\text{eff}}_{\text{obs}}]$ (see \cite{Soi06,Soi08a,Soi17a}) and therefore allows the level of statistical fluctuations of the $6$ last components $Q^{\text{obs}}_4$, $Q^{\text{obs}}_5$, $Q^{\text{obs}}_6$, $Q^{\text{obs}}_7$, $Q^{\text{obs}}_8$ and $Q^{\text{obs}}_9$ of random vector $\Qb^{\text{obs}}$ (around their respective mean values) to be indirectly controlled. Figure~\ref{fig:network_input_uncertainty_level} shows the evolutions of the coefficient of variation $\delta_{Q^{\text{obs}}_i}
$ of each component $Q^{\text{obs}}_i$ of random vector $\Qb^{\text{obs}}$, for $i=4,\dots,9$, with respect to $s$ in order to quantify the impact of dispersion parameter $s$ on the network input random variables $Q^{\text{obs}}_4,\dots,Q^{\text{obs}}_9$. For each input random variable $Q^{\text{obs}}_i$, the coefficient of variation $\delta_{Q^{\text{obs}}_i}$ increases linearly with input uncertainty level $s$.

\setlength\figureheight{0.12\textheight}
\begin{figure}[h!]
	\centering
	\begin{subfigure}{.33\linewidth}
		\centering
		\tikzsetnextfilename{network_input_uncertainty_level_4}
%
\begin{tikzpicture}

\begin{axis}[%
width=1.269\figureheight,
height=\figureheight,
at={(0\figureheight,0\figureheight)},
scale only axis,
xmin=0,
xmax=0.05,
xlabel={$s$},
ymin=0,
ymax=0.0012,
ylabel={$\delta_{Q_4^{\text{obs}}}$},
xmajorgrids,
ymajorgrids
]
\addplot [color=blue, line width=1.0pt, forget plot]
  table[row sep=crcr]{%
0	0\\
0.01	0.000228712950465329\\
0.02	0.000457436454831738\\
0.03	0.000686181156303872\\
0.04	0.000914957700979713\\
0.05	0.0011437767392257\\
};
\end{axis}
\end{tikzpicture}%
		\caption{input random variable $Q_4^{\text{obs}}$}
		\label{fig:network_input_uncertainty_level_4}
	\end{subfigure}\hfill
	\begin{subfigure}{.33\linewidth}
		\centering
		\tikzsetnextfilename{network_input_uncertainty_level_5}
%
\begin{tikzpicture}

\begin{axis}[%
width=1.269\figureheight,
height=\figureheight,
at={(0\figureheight,0\figureheight)},
scale only axis,
xmin=0,
xmax=0.05,
xlabel={$s$},
ymin=0,
ymax=0.06,
ylabel={$\delta_{Q_5^{\text{obs}}}$},
xmajorgrids,
ymajorgrids
]
\addplot [color=blue, line width=1.0pt, forget plot]
  table[row sep=crcr]{%
0	0\\
0.01	0.011003813796265\\
0.02	0.0220079836844272\\
0.03	0.0330127383907845\\
0.04	0.0440183066668803\\
0.05	0.0550249172963806\\
};
\end{axis}
\end{tikzpicture}%
		\caption{input random variable $Q_5^{\text{obs}}$}
		\label{fig:network_input_uncertainty_level_5}
	\end{subfigure}\hfill
	\begin{subfigure}{.33\linewidth}
		\centering
		\tikzsetnextfilename{network_input_uncertainty_level_6}
%
\begin{tikzpicture}

\begin{axis}[%
width=1.269\figureheight,
height=\figureheight,
at={(0\figureheight,0\figureheight)},
scale only axis,
xmin=0,
xmax=0.05,
xlabel={$s$},
ymin=0,
ymax=25,
ylabel={$\delta_{Q_6^{\text{obs}}}$},
xmajorgrids,
ymajorgrids
]
\addplot [color=blue, line width=1.0pt, forget plot]
  table[row sep=crcr]{%
0	0\\
0.01	4.06883315738818\\
0.02	8.20396205837978\\
0.03	12.4071110608195\\
0.04	16.6800646758774\\
0.05	21.0246702111607\\
};
\end{axis}
\end{tikzpicture}%
		\caption{input random variable $Q_6^{\text{obs}}$}
		\label{fig:network_input_uncertainty_level_6}
	\end{subfigure}\\
	\begin{subfigure}{.33\linewidth}
		\centering
		\tikzsetnextfilename{network_input_uncertainty_level_7}
%
\begin{tikzpicture}

\begin{axis}[%
width=1.269\figureheight,
height=\figureheight,
at={(0\figureheight,0\figureheight)},
scale only axis,
xmin=0,
xmax=0.05,
xlabel={$s$},
ymin=0,
ymax=0.0012,
ylabel={$\delta_{Q_7^{\text{obs}}}$},
xmajorgrids,
ymajorgrids
]
\addplot [color=blue, line width=1.0pt, forget plot]
  table[row sep=crcr]{%
0	0\\
0.01	0.000229640610753931\\
0.02	0.000459303651555143\\
0.03	0.000689010501404616\\
0.04	0.000918782556652421\\
0.05	0.00114864123948201\\
};
\end{axis}
\end{tikzpicture}%
		\caption{input random variable $Q_7^{\text{obs}}$}
		\label{fig:network_input_uncertainty_level_7}
	\end{subfigure}\hfill
	\begin{subfigure}{.33\linewidth}
		\centering
		\tikzsetnextfilename{network_input_uncertainty_level_8}
%
\begin{tikzpicture}

\begin{axis}[%
width=1.269\figureheight,
height=\figureheight,
at={(0\figureheight,0\figureheight)},
scale only axis,
xmin=0,
xmax=0.05,
xlabel={$s$},
ymin=0,
ymax=6,
ylabel={$\delta_{Q_8^{\text{obs}}}$},
xmajorgrids,
ymajorgrids
]
\addplot [color=blue, line width=1.0pt, forget plot]
  table[row sep=crcr]{%
0	0\\
0.01	1.02067493763966\\
0.02	2.04053491427994\\
0.03	3.05964640541745\\
0.04	4.07807569813594\\
0.05	5.09588891107882\\
};
\end{axis}
\end{tikzpicture}%
		\caption{input random variable $Q_8^{\text{obs}}$}
		\label{fig:network_input_uncertainty_level_8}
	\end{subfigure}\hfill
	\begin{subfigure}{.33\linewidth}
		\centering
		\tikzsetnextfilename{network_input_uncertainty_level_9}
%
\begin{tikzpicture}

\begin{axis}[%
width=1.269\figureheight,
height=\figureheight,
at={(0\figureheight,0\figureheight)},
scale only axis,
xmin=0,
xmax=0.05,
xlabel={$s$},
ymin=0,
ymax=0.0014,
ylabel={$\delta_{Q_9^{\text{obs}}}$},
xmajorgrids,
ymajorgrids
]
\addplot [color=blue, line width=1.0pt, forget plot]
  table[row sep=crcr]{%
0	0\\
0.01	0.0002398476131846\\
0.02	0.000479728157523655\\
0.03	0.000719675266771567\\
0.04	0.00095972261705875\\
0.05	0.00119990394827158\\
};
\end{axis}
\end{tikzpicture}%
		\caption{input random variable $Q_9^{\text{obs}}$}
		\label{fig:network_input_uncertainty_level_9}
	\end{subfigure}
	\caption{Synthetic data: evolutions of the coefficient of variation $\delta_{Q_i^{\text{obs}}}$ of each network input random variable $Q_i^{\text{obs}}$ estimated from the $N_s=10^6$ independent realizations $q_i^{\text{obs},(1)},\dots,q_i^{\text{obs},(N_s)}$, for $i=4,\dots,9$, with respect to input uncertainty level $s$}
	\label{fig:network_input_uncertainty_level}
\end{figure}
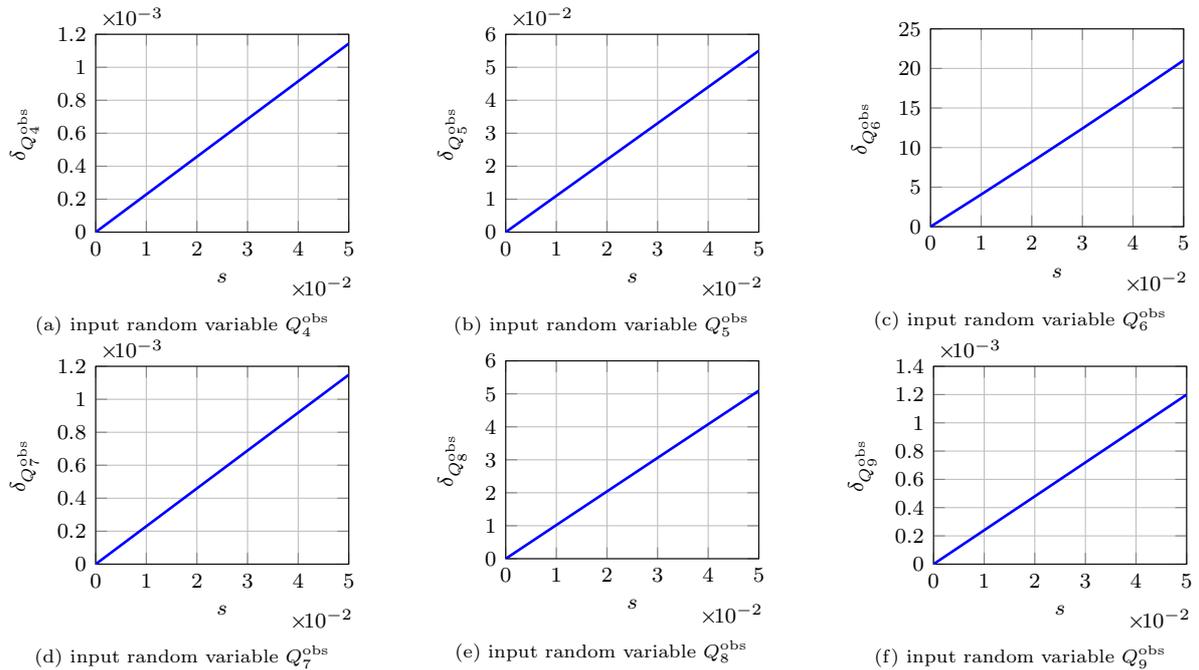

The identification results obtained with the trained neural network for the different input uncertainty levels are summarized in Table~\ref{tab:identification}. In addition, Figure~\ref{fig:robustness} displays the graphs of the mean output values $\underline{h}_1^{\text{out}}$, $\underline{h}_2^{\text{out}}$, $\underline{h}_3^{\text{out}}$ and $\underline{h}_4^{\text{out}}$ and the corresponding confidence regions (with a probability level $95\%$) of $H_1^{\text{out}}$, $H_2^{\text{out}}$, $H_3^{\text{out}}$ and $H_4^{\text{out}}$, respectively, with respect to the input uncertainty level $s$. First, in the absence of uncertainty on input vector $\qb^{\text{obs}}$ (\ie{} for an input uncertainty level $s=0$), the values of output vector $\hb^{\text{out}}$ computed by using the trained neural network 
are very close to the corresponding values of target vector $\hb^{\text{target}}$ with relative errors less than $1\%$ for each of the random hyperparameters $H_1$, $H_2$, $H_3$ and $H_4$. Secondly, in the presence of uncertainties on input vector $\qb^{\text{obs}}$, the values of output mean vector $\underline{\hb}^{\text{out}}$ remain close to the corresponding values of target vector $\hb^{\text{target}}$ with maximum relative errors less than $0.3\%$, $2\%$, $3\%$ and $3\%$ for the mean output values $\underline{h}_1^{\text{out}}$, $\underline{h}_2^{\text{out}}$, $\underline{h}_3^{\text{out}}$ and $\underline{h}_4^{\text{out}}$, respectively, for the highest input uncertainty level $s=0.05=5\%$ considered here. Thus, even though the output $95\%$ confidence intervals become wider as the input uncertainty level $s$ increases, the values of output mean vector $\underline{\hb}^{\text{out}}$ present small variations with respect to the input uncertainty level $s$.

\begin{table}[h!]
\caption{Synthetic data: comparison of output vector $\hb^{\text{out}}$ and output mean vector $\underline{\hb}^{\text{out}}$ with associated target vector $\hb^{\text{target}}$, and output $95\%$ confidence interval $I^{\text{out}}$ obtained for a given input vector $\qb^{\text{obs}}$ contained in the test dataset and for different values of input uncertainty level $s$}
\label{tab:identification}
\centering
\begin{tabular}{|cc|c|c|c|c|} \hline
 & & $h_1$ & $h_2$ [$\mu$m] & $h_3$ [GPa] & $h_4$ [GPa] \\ \hline
Target vector $\hb^{\text{target}}$ & & $0.5514$ & $172.36$ & $12.398$ & $4.672$ \\ \hline
Output vector $\hb^{\text{out}}$ & $s = 0\%$ & $0.5501$ & $172.61$ & $12.322$ & $4.693$ \\ \cline{2-6}
\multirow{5}{*}{Output mean vector $\underline{\hb}^{\text{out}}$} & $s=1\%$ & $0.5503$ & $172.71$ & $12.297$ & $4.695$ \\
 & $s=2\%$ & $0.5507$ & $173.03$ & $12.250$ & $4.705$ \\
 & $s=3\%$ & $0.5508$ & $173.47$ & $12.211$ & $4.725$ \\
 & $s=4\%$ & $0.5513$ & $174.11$ & $12.167$ & $4.758$ \\
 & $s=5\%$ & $0.5527$ & $175.25$ & $12.106$ & $4.802$ \\ \hline
 \multirow{6}{*}{Relative error [\%]} & $s=0\%$ & $0.24$ & $0.14$ & $0.61$ & $0.46$ \\ \cline{2-6}
 & $s=1\%$ & $0.20$ & $0.20$ & $0.81$ & $0.51$ \\
 & $s=2\%$ & $0.14$ & $0.39$ & $1.19$ & $0.71$ \\
 & $s=3\%$ & $0.11$ & $0.64$ & $1.50$ & $1.14$ \\
 & $s=4\%$ & $0.02$ & $1.01$ & $1.86$ & $1.86$ \\
 & $s=5\%$ & $0.22$ & $1.67$ & $2.35$ & $2.80$ \\ \hline
\multirow{4}{*}{Output $95\%$} & $s=1\%$ & $\intervalcc{0.5388}{0.5623}$ & $\intervalcc{168.53}{177.07}$ & $\intervalcc{11.856}{12.741}$ & $\intervalcc{4.644}{4.750}$ \\ 
\multirow{4}{*}{confidence interval $I^{\text{out}}$} & $s=2\%$ & $\intervalcc{0.5259}{0.5762}$ & $\intervalcc{164.65}{181.86}$ & $\intervalcc{11.296}{13.184}$ & $\intervalcc{4.595}{4.843}$ \\
 & $s=3\%$ & $\intervalcc{0.5080}{0.5928}$ & $\intervalcc{160.10}{187.76}$ & $\intervalcc{10.494}{13.858}$ & $\intervalcc{4.542}{5.008}$ \\
 & $s=4\%$ & $\intervalcc{0.4823}{0.6160}$ & $\intervalcc{153.91}{197.04}$ & $\intervalcc{9.105}{15.072}$ & $\intervalcc{4.484}{5.218}$ \\
 & $s=5\%$ & $\intervalcc{0.4519}{0.6498}$ & $\intervalcc{145.62}{211.39}$ & $\intervalcc{6.700}{16.852}$ & $\intervalcc{4.424}{5.432}$ \\ \hline
\end{tabular}
\end{table}

\setlength\figureheight{0.2\textheight}
\begin{figure}[h!]
	\centering
	\begin{subfigure}{.5\linewidth}
		\centering
		\tikzsetnextfilename{robustness_delta}
%
\begin{tikzpicture}

\begin{axis}[%
width=1.269\figureheight,
height=\figureheight,
at={(0\figureheight,0\figureheight)},
scale only axis,
xmin=0,
xmax=0.05,
xlabel={$s$},
ymin=0.45,
ymax=0.65,
ylabel={$h_1^{\text{out}}$},
xmajorgrids,
ymajorgrids,
legend style={legend cell align=left, align=left
}
]

\addplot[area legend, draw=blue, fill=blue, fill opacity=0.2]
table[row sep=crcr] {%
x	y\\
0	0.550092530084554\\
0.01	0.562300630132845\\
0.02	0.576187030628472\\
0.03	0.59277867889269\\
0.04	0.615998372756246\\
0.05	0.649827562655241\\
0.05	0.451865102899687\\
0.04	0.482330291147727\\
0.03	0.507994318216257\\
0.02	0.525903314372757\\
0.01	0.538779056303422\\
0	0.550092530084554\\
}--cycle;

\addplot [color=blue, line width=1.0pt]
  table[row sep=crcr]{%
0	0.550092530084554\\
0.01	0.550306833380102\\
0.02	0.550652325485249\\
0.03	0.550823740565809\\
0.04	0.551292084996919\\
0.05	0.552657321531926\\
};

\addplot [color=red, line width=1.0pt]
  table[row sep=crcr]{%
0	0.551428970840346\\
0.01	0.551428970840346\\
0.02	0.551428970840346\\
0.03	0.551428970840346\\
0.04	0.551428970840346\\
0.05	0.551428970840346\\
};

\end{axis}
\end{tikzpicture}%
		\caption{dispersion parameter $H_1^{\text{out}}$}
		\label{fig:robustness_delta}
	\end{subfigure}\hfill
	\begin{subfigure}{.5\linewidth}
		\centering
		\tikzsetnextfilename{robustness_lcorr}
%
\begin{tikzpicture}

\begin{axis}[%
width=1.269\figureheight,
height=\figureheight,
at={(0\figureheight,0\figureheight)},
scale only axis,
xmin=0,
xmax=0.05,
xlabel={$s$},
ymin=140,
ymax=220,
ylabel={$h_2^{\text{out}}$ [$\mu$m]},
xmajorgrids,
ymajorgrids,
legend style={legend cell align=left, align=left
}
]

\addplot[area legend, draw=blue, fill=blue, fill opacity=0.2]
table[row sep=crcr] {%
x	y\\
0	172.610701467813\\
0.01	177.071308740281\\
0.02	181.854991355934\\
0.03	187.761099513655\\
0.04	197.043353005823\\
0.05	211.388082193543\\
0.05	145.615476173924\\
0.04	153.911279922276\\
0.03	160.098423760963\\
0.02	164.646840224866\\
0.01	168.524580167525\\
0	172.610701467813\\
}--cycle;

\addplot [color=blue, line width=1.0pt]
  table[row sep=crcr]{%
0	172.610701467813\\
0.01	172.71344903383\\
0.02	173.027992207482\\
0.03	173.46852955496\\
0.04	174.112274745411\\
0.05	175.246426817768\\
};

\addplot [color=red, line width=1.0pt]
  table[row sep=crcr]{%
0	172.364297271163\\
0.01	172.364297271163\\
0.02	172.364297271163\\
0.03	172.364297271163\\
0.04	172.364297271163\\
0.05	172.364297271163\\
};

\end{axis}
\end{tikzpicture}%
		\caption{spatial correlation length $H_2^{\text{out}}$}
		\label{fig:robustness_lcorr}
	\end{subfigure}\\
	\begin{subfigure}{.5\linewidth}
		\centering
		\tikzsetnextfilename{robustness_kappa}
%
\begin{tikzpicture}

\begin{axis}[%
width=1.269\figureheight,
height=\figureheight,
at={(0\figureheight,0\figureheight)},
scale only axis,
xmin=0,
xmax=0.05,
xlabel={$s$},
ymin=6,
ymax=18,
ylabel={$h_3^{\text{out}}$ [GPa]},
xmajorgrids,
ymajorgrids,
legend style={legend cell align=left, align=left
}
]

\addplot[area legend, draw=blue, fill=blue, fill opacity=0.2]
table[row sep=crcr] {%
x	y\\
0	12.3217601212031\\
0.01	12.7409755013647\\
0.02	13.1837422664821\\
0.03	13.8575039051211\\
0.04	15.0715878193776\\
0.05	16.8521033313744\\
0.05	6.69987810803637\\
0.04	9.104823361641\\
0.03	10.493835621537\\
0.02	11.2956616615241\\
0.01	11.855945251095\\
0	12.3217601212031\\
}--cycle;

\addplot [color=blue, line width=1.0pt]
  table[row sep=crcr]{%
0	12.3217601212031\\
0.01	12.2971280059276\\
0.02	12.2502133155349\\
0.03	12.2110273908663\\
0.04	12.1670645217679\\
0.05	12.1060777966416\\
};

\addplot [color=red, line width=1.0pt]
  table[row sep=crcr]{%
0	12.3975482376603\\
0.01	12.3975482376603\\
0.02	12.3975482376603\\
0.03	12.3975482376603\\
0.04	12.3975482376603\\
0.05	12.3975482376603\\
};

\end{axis}
\end{tikzpicture}%
		\caption{mean bulk modulus $H_3^{\text{out}}$}
		\label{fig:robustness_kappa}
	\end{subfigure}\hfill
	\begin{subfigure}{.5\linewidth}
		\centering
		\tikzsetnextfilename{robustness_mu}
%
\begin{tikzpicture}

\begin{axis}[%
width=1.269\figureheight,
height=\figureheight,
at={(0\figureheight,0\figureheight)},
scale only axis,
xmin=0,
xmax=0.05,
xlabel={$s$},
ymin=4.4,
ymax=5.6,
ylabel={$h_4^{\text{out}}$ [GPa]},
xmajorgrids,
ymajorgrids,
legend style={legend cell align=left, align=left
},
legend pos=north east
]

\addplot[area legend, draw=blue, fill=blue, fill opacity=0.2]
table[row sep=crcr] {%
x	y\\
0	4.69290490675317\\
0.01	4.74991570647908\\
0.02	4.84272884880651\\
0.03	5.00804435332183\\
0.04	5.21847716057856\\
0.05	5.43161930545581\\
0.05	4.42406152136539\\
0.04	4.48374456791555\\
0.03	4.54170236469999\\
0.02	4.59515398819448\\
0.01	4.64402432550943\\
0	4.69290490675317\\
}--cycle;
\addlegendentry{Output $95\%$ confidence interval}

\addplot [color=blue, line width=1.0pt]
  table[row sep=crcr]{%
0	4.69290490675317\\
0.01	4.69531851866888\\
0.02	4.7045888490602\\
0.03	4.72488238492318\\
0.04	4.75833457105887\\
0.05	4.80246747521449\\
};
\addlegendentry{Output mean value}

\addplot [color=red, line width=1.0pt]
  table[row sep=crcr]{%
0	4.67151710393869\\
0.01	4.67151710393869\\
0.02	4.67151710393869\\
0.03	4.67151710393869\\
0.04	4.67151710393869\\
0.05	4.67151710393869\\
};
\addlegendentry{Target value}

\end{axis}
\end{tikzpicture}%
		\caption{mean shear modulus $H_4^{\text{out}}$}
		\label{fig:robustness_mu}
	\end{subfigure}
	\caption{Synthetic data: evolutions of the output mean values $\underline{\hb}^{\text{out}}$ (blue curve) and the $95\%$ confidence intervals $I^{\text{out}}$ (blue areas) of random variables $H_1^{\text{out}}$, $H_2^{\text{out}}$, $H_3^{\text{out}}$ and $H_4^{\text{out}}$, respectively, with respect to input uncertainty level $s$, obtained for a given input vector $\qb^{\text{obs}}$, with the corresponding target values $\hb^{\text{target}}$ (red lines)}
	\label{fig:robustness}
\end{figure}
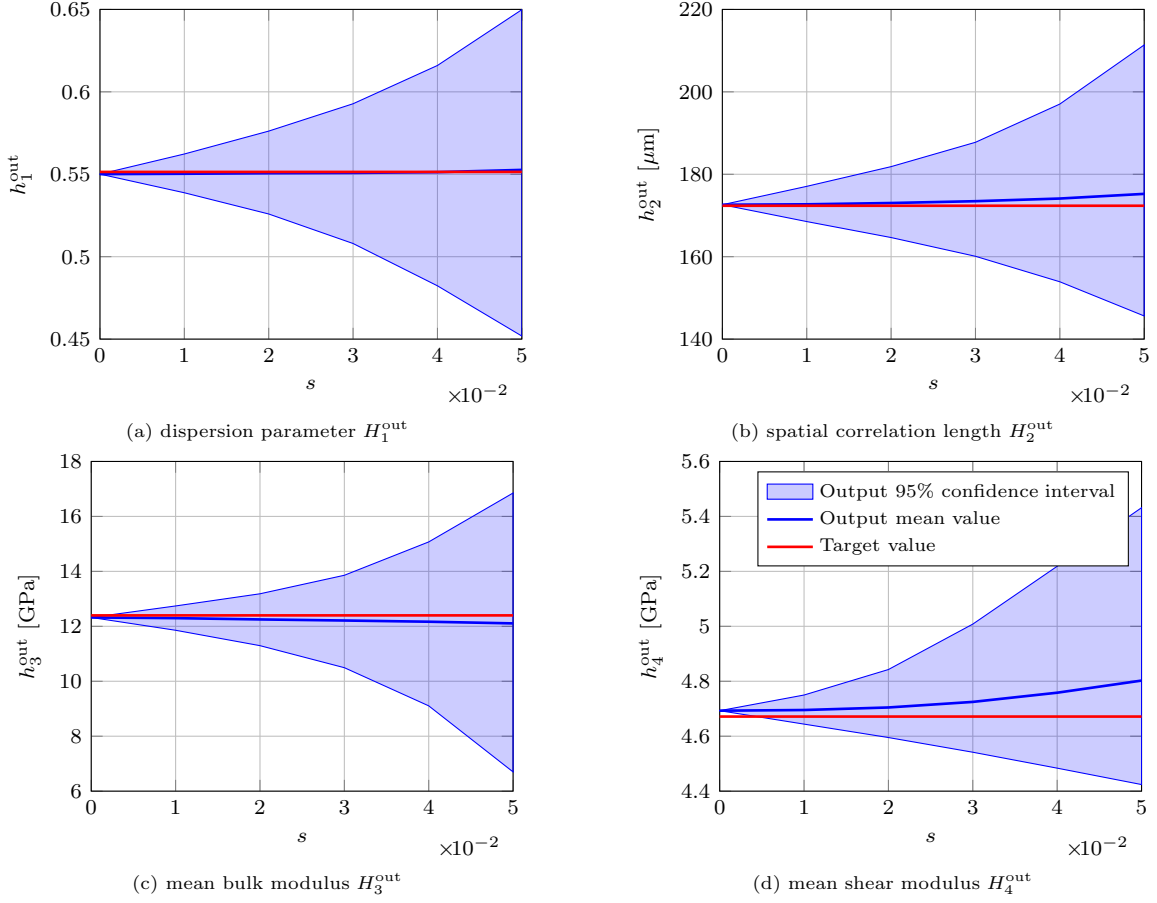

Figure~\ref{fig:pdf_posterior} shows the marginal probability density functions $p_{H_1^{\text{out}}}$, $p_{H_2^{\text{out}}}$, $p_{H_3^{\text{out}}}$ and $p_{H_4^{\text{out}}}$ of random variables $H_1^{\text{out}}$, $H_2^{\text{out}}$, $H_3^{\text{out}}$ and $H_4^{\text{out}}$, respectively, estimated by using the univariate Gaussian kernel density estimation method with the $N_s = 10^6$ independent realizations $\hb^{\text{out},(1)},\dots,\hb^{\text{out},(N_s)}$ of $\Hb^{\text{out}} = (H_1^{\text{out}},H_2^{\text{out}},H_3^{\text{out}},H_4^{\text{out}})$ for a given input uncertainty level $s = 0.01 = 1\%$. For such a small dispersion on the input values, the output values generated by the trained neural network present small fluctuations and remain concentrated around the output mean value, and the associated target value lies within the output $95\%$ confidence interval, for each of the random variables $H_1^{\text{out}}$, $H_2^{\text{out}}$, $H_3^{\text{out}}$ and $H_4^{\text{out}}$. Figure~\ref{fig:pdfs_posterior} represents the marginal probability density functions $p_{H_1^{\text{out}}}$, $p_{H_2^{\text{out}}}$, $p_{H_3^{\text{out}}}$ and $p_{H_4^{\text{out}}}$ for the different values of input uncertainty level $s$ ranging from $0.01=1\%$ to $0.05=5\%$. For each component $H_j^{\text{out}}$ of output random vector $\Hb^{\text{out}}$, the higher the uncertainty level $s$, the more flattened the probability density function $p_{H_j^{\text{out}}}$ is, but the mean value $\underline{h}_j^{\text{out}}$ of output random variable $H_j^{\text{out}}$ still remains a good approximation of the target value $h_j^{\text{target}}$ even with an input level of uncertainties $s = 0.05 = 5\%$ on observed input vector $\qb^{\text{obs}}$. The proposed neural network-based identification method then remains accurate even in the presence of uncertainties on the given input vector $\qb^{\text{obs}}$ of quantities of interest. It can therefore be applied to experimentally measured quantities of interest.

\begin{figure}[h!]
	\centering
	\begin{subfigure}{.5\linewidth}
		\centering
		\tikzsetnextfilename{pdf_posterior_delta_1}
%
\begin{tikzpicture}

\begin{axis}[%
width=1.269\figureheight,
height=\figureheight,
at={(0\figureheight,0\figureheight)},
scale only axis,
xmin=0.52,
xmax=0.58,
xlabel={$h_1$},
ymin=0,
ymax=70,
ylabel={$p_{H_1^{\text{out}}}(h_1)$},
xmajorgrids,
ymajorgrids,
legend style={legend cell align=left, align=left
}
]
\addplot [color=blue, line width=1.0pt]
  table[row sep=crcr]{%
0.520805241529915	1.19897692907121e-05\\
0.521445023998611	0.000473326153186515\\
0.522084806467306	0.00185315560456405\\
0.522724588936001	0.00140949888703729\\
0.523364371404696	0.00218380383767098\\
0.524004153873391	0.00229463635962569\\
0.524643936342087	0.00514195936554649\\
0.525283718810782	0.0129962693624405\\
0.525923501279477	0.0139833789255549\\
0.526563283748172	0.0161017316332382\\
0.527203066216868	0.0264162854105695\\
0.527842848685563	0.046290899214076\\
0.528482631154258	0.0717090804742207\\
0.529122413622953	0.105118707805265\\
0.529762196091648	0.143345136626078\\
0.530401978560344	0.227769825005359\\
0.531041761029039	0.306740471517492\\
0.531681543497734	0.428523191797681\\
0.532321325966429	0.606675387144824\\
0.532961108435125	0.827159259079314\\
0.53360089090382	1.1598303622124\\
0.534240673372515	1.60006473696541\\
0.53488045584121	2.16997305390487\\
0.535520238309905	2.88682531700168\\
0.536160020778601	3.85682035410062\\
0.536799803247296	5.03220984007511\\
0.537439585715991	6.33905006444798\\
0.538079368184686	7.94286833226386\\
0.538719150653381	10.0836930850267\\
0.539358933122077	12.6190656235527\\
0.539998715590772	15.2140329079558\\
0.540638498059467	18.2955036507452\\
0.541278280528162	21.7974526035376\\
0.541918062996858	25.4170027531662\\
0.542557845465553	29.3637864139398\\
0.543197627934248	33.9875262200364\\
0.543837410402943	38.4150401217976\\
0.544477192871639	42.6919054806958\\
0.545116975340334	46.8576050054497\\
0.545756757809029	51.040001397248\\
0.546396540277724	54.7594563196178\\
0.547036322746419	58.2521696815366\\
0.547676105215115	61.3093708980967\\
0.54831588768381	63.3010807451172\\
0.548955670152505	64.9362481349036\\
0.5495954526212	66.1290287894202\\
0.550235235089895	66.0949419497083\\
0.550875017558591	65.946521326947\\
0.551514800027286	64.9825198933188\\
0.552154582495981	62.9969396370157\\
0.552794364964676	60.1804446579964\\
0.553434147433372	56.9666998899654\\
0.554073929902067	53.5821889780573\\
0.554713712370762	49.62448819925\\
0.555353494839457	45.4220119423784\\
0.555993277308153	41.1200458164079\\
0.556633059776848	37.0491005395431\\
0.557272842245543	32.9717903837366\\
0.557912624714238	29.1634885794557\\
0.558552407182933	25.3438880419044\\
0.559192189651629	21.93233081204\\
0.559831972120324	18.6088618555297\\
0.560471754589019	15.6687326919582\\
0.561111537057714	13.21267333623\\
0.561751319526409	10.9275703580833\\
0.562391101995105	8.94832974816096\\
0.5630308844638	7.28697426293594\\
0.563670666932495	5.93995127868072\\
0.56431044940119	4.7386382379091\\
0.564950231869886	3.73308510228981\\
0.565590014338581	2.97552157806238\\
0.566229796807276	2.25866253205764\\
0.566869579275971	1.72154944491289\\
0.567509361744666	1.32951661141319\\
0.568149144213362	1.02618877374001\\
0.568788926682057	0.792662428202254\\
0.569428709150752	0.569035268547024\\
0.570068491619447	0.414098383303734\\
0.570708274088143	0.327087039806759\\
0.571348056556838	0.230148029028733\\
0.571987839025533	0.148795967361975\\
0.572627621494228	0.109884873946712\\
0.573267403962923	0.0884865474907405\\
0.573907186431619	0.0559857540480742\\
0.574546968900314	0.037131965092148\\
0.575186751369009	0.0281444971737804\\
0.575826533837704	0.0232967732036237\\
0.5764663163064	0.0154523128153435\\
0.577106098775095	0.011930893736846\\
0.57774588124379	0.00816554462126264\\
0.578385663712485	0.00302709352443156\\
0.57902544618118	0.00353488924674007\\
0.579665228649876	0.00390112674529164\\
0.580305011118571	0.00151114661407068\\
0.580944793587266	0.000575241814484976\\
0.581584576055961	0.000953782514641164\\
0.582224358524656	0.000361770985349474\\
0.582864140993352	0.000979436088847434\\
0.583503923462047	0.000372271470630825\\
0.584143705930742	1.10634848281775e-05\\
};

\addplot[area legend, draw=blue, fill=blue, fill opacity=0.2] table[row sep=crcr]{%
0.539358933122077	12.6190656235527\\
0.539998715590772	15.2140329079558\\
0.540638498059467	18.2955036507452\\
0.541278280528162	21.7974526035376\\
0.541918062996858	25.4170027531662\\
0.542557845465553	29.3637864139398\\
0.543197627934248	33.9875262200364\\
0.543837410402943	38.4150401217976\\
0.544477192871639	42.6919054806958\\
0.545116975340334	46.8576050054497\\
0.545756757809029	51.040001397248\\
0.546396540277724	54.7594563196178\\
0.547036322746419	58.2521696815366\\
0.547676105215115	61.3093708980967\\
0.54831588768381	63.3010807451172\\
0.548955670152505	64.9362481349036\\
0.5495954526212	66.1290287894202\\
0.550235235089895	66.0949419497083\\
0.550875017558591	65.946521326947\\
0.551514800027286	64.9825198933188\\
0.552154582495981	62.9969396370157\\
0.552794364964676	60.1804446579964\\
0.553434147433372	56.9666998899654\\
0.554073929902067	53.5821889780573\\
0.554713712370762	49.62448819925\\
0.555353494839457	45.4220119423784\\
0.555993277308153	41.1200458164079\\
0.556633059776848	37.0491005395431\\
0.557272842245543	32.9717903837366\\
0.557912624714238	29.1634885794557\\
0.558552407182933	25.3438880419044\\
0.559192189651629	21.93233081204\\
0.559831972120324	18.6088618555297\\
0.560471754589019	15.6687326919582\\
0.561111537057714	13.21267333623\\
0.561751319526409	10.9275703580833\\
}
\closedcycle;

\addplot[only marks, mark=diamond*, mark options={}, mark size=2.0000pt, draw=black, fill=blue] table[row sep=crcr]{%
x	y\\
0.550306833380102	0\\
};

\addplot[only marks, mark=*, mark options={}, mark size=1.5000pt, draw=black, fill=black!20!green] table[row sep=crcr]{%
x	y\\
0.550092530084554	0\\
};

\addplot[only marks, mark=*, mark options={}, mark size=1.5000pt, draw=black, fill=red] table[row sep=crcr]{%
x	y\\
0.551428970840346	0\\
};

\end{axis}
\end{tikzpicture}%
		\caption{$p_{H_1^{\text{out}}}(h_1)$}
		\label{fig:pdf_posterior_delta_1}
	\end{subfigure}\hfill
	\begin{subfigure}{.5\linewidth}
		\centering
		\tikzsetnextfilename{pdf_posterior_lcorr_1}
%
\begin{tikzpicture}

\begin{axis}[%
width=1.269\figureheight,
height=\figureheight,
at={(0\figureheight,0\figureheight)},
scale only axis,
xmin=163,
xmax=182,
xlabel={$h_2$ [$\mu$m]},
ymin=0,
ymax=200000,
ylabel={$p_{H_2^{\text{out}}}(h_2)$},
xmajorgrids,
ymajorgrids,
legend style={legend cell align=left, align=left
}
]
\addplot [color=blue, line width=1.0pt]
  table[row sep=crcr]{%
162.530535213145	0.0305642723563655\\
162.749425082186	0.917882960713346\\
162.968314951228	3.51244313944401\\
163.187204820269	5.66701080231558\\
163.40609468931	7.64017623696702\\
163.624984558351	19.5905717665996\\
163.843874427392	27.7435409569908\\
164.062764296433	32.9393987250772\\
164.281654165475	50.2159188453446\\
164.500544034516	90.3539771174401\\
164.719433903557	134.754857613148\\
164.938323772598	203.318956808243\\
165.157213641639	314.261647471136\\
165.376103510681	465.314156999129\\
165.594993379722	679.931498974198\\
165.813883248763	979.52887757646\\
166.032773117804	1293.92162223665\\
166.251662986845	1803.18000892258\\
166.470552855887	2573.56634876868\\
166.689442724928	3484.62817649725\\
166.908332593969	4503.14892760322\\
167.12722246301	6086.53024782437\\
167.346112332051	7912.38532779409\\
167.565002201093	10424.3488270044\\
167.783892070134	13459.9595726605\\
168.002781939175	16718.855705871\\
168.221671808216	20941.390089416\\
168.440561677257	26001.5596652127\\
168.659451546299	31696.1208210731\\
168.87834141534	38726.9842666556\\
169.097231284381	46331.2665888824\\
169.316121153422	54332.7507832958\\
169.535011022463	64123.5691684295\\
169.753900891504	74099.6964172717\\
169.972790760546	84933.7109271227\\
170.191680629587	96371.1094658859\\
170.410570498628	108091.778007881\\
170.629460367669	118655.992811881\\
170.84835023671	129611.394883587\\
171.067240105752	140935.541471454\\
171.286129974793	151900.083630799\\
171.505019843834	160567.890182158\\
171.723909712875	167870.103242785\\
171.942799581916	174390.450640216\\
172.161689450958	179187.422665949\\
172.380579319999	182444.623827585\\
172.59946918904	184137.276984727\\
172.818359058081	183117.34512933\\
173.037248927122	180031.634869218\\
173.256138796164	175690.081372154\\
173.475028665205	170048.181628508\\
173.693918534246	162121.482358347\\
173.912808403287	153184.383534088\\
174.131698272328	144106.781300333\\
174.350588141369	134211.319621801\\
174.569478010411	123630.118637237\\
174.788367879452	113357.717367744\\
175.007257748493	102552.437457968\\
175.226147617534	91688.458505514\\
175.445037486575	81339.2602777669\\
175.663927355617	71517.6684724401\\
175.882817224658	62062.3536260967\\
176.101707093699	53750.0272338299\\
176.32059696274	46133.834324927\\
176.539486831781	39237.4653832198\\
176.758376700823	33028.7426114051\\
176.977266569864	27322.1460983418\\
177.196156438905	22252.3481949572\\
177.415046307946	18533.1871531209\\
177.633936176987	15439.2248530339\\
177.852826046029	12575.3834669336\\
178.07171591507	10010.1727711789\\
178.290605784111	8021.01526228151\\
178.509495653152	6250.56246274983\\
178.728385522193	4931.48403127319\\
178.947275391235	3832.94143299927\\
179.166165260276	2890.22781655975\\
179.385055129317	2163.00279879739\\
179.603944998358	1641.30466253278\\
179.822834867399	1312.16613273012\\
180.041724736441	916.609215859051\\
180.260614605482	688.330466626489\\
180.479504474523	528.564840435673\\
180.698394343564	394.495936948837\\
180.917284212605	317.362290311719\\
181.136174081646	234.866118400027\\
181.355063950688	160.709146928588\\
181.573953819729	120.084210928331\\
181.79284368877	86.7204221464283\\
182.011733557811	49.9240075925627\\
182.230623426852	36.9513973027047\\
182.449513295894	25.5115148996628\\
182.668403164935	20.4957477593703\\
182.887293033976	11.7219365248838\\
183.106182903017	3.69597647194864\\
183.325072772058	0.240782235906644\\
183.5439626411	1.82487918633982\\
183.762852510141	5.49191554498144\\
183.981742379182	1.70563251235337\\
184.200632248223	0.0546655572577398\\
};

\addplot[area legend, draw=blue, fill=blue, fill opacity=0.2] table[row sep=crcr]{%
168.659451546299	31696.1208210731\\
168.87834141534	38726.9842666556\\
169.097231284381	46331.2665888824\\
169.316121153422	54332.7507832958\\
169.535011022463	64123.5691684295\\
169.753900891504	74099.6964172717\\
169.972790760546	84933.7109271227\\
170.191680629587	96371.1094658859\\
170.410570498628	108091.778007881\\
170.629460367669	118655.992811881\\
170.84835023671	129611.394883587\\
171.067240105752	140935.541471454\\
171.286129974793	151900.083630799\\
171.505019843834	160567.890182158\\
171.723909712875	167870.103242785\\
171.942799581916	174390.450640216\\
172.161689450958	179187.422665949\\
172.380579319999	182444.623827585\\
172.59946918904	184137.276984727\\
172.818359058081	183117.34512933\\
173.037248927122	180031.634869218\\
173.256138796164	175690.081372154\\
173.475028665205	170048.181628508\\
173.693918534246	162121.482358347\\
173.912808403287	153184.383534088\\
174.131698272328	144106.781300333\\
174.350588141369	134211.319621801\\
174.569478010411	123630.118637237\\
174.788367879452	113357.717367744\\
175.007257748493	102552.437457968\\
175.226147617534	91688.458505514\\
175.445037486575	81339.2602777669\\
175.663927355617	71517.6684724401\\
175.882817224658	62062.3536260967\\
176.101707093699	53750.0272338299\\
176.32059696274	46133.834324927\\
176.539486831781	39237.4653832198\\
176.758376700823	33028.7426114051\\
176.977266569864	27322.1460983418\\
}
\closedcycle;

\addplot[only marks, mark=diamond*, mark options={}, mark size=2.0000pt, draw=black, fill=blue] table[row sep=crcr]{%
x	y\\
172.71344903383	0\\
};

\addplot[only marks, mark=*, mark options={}, mark size=1.5000pt, draw=black, fill=black!20!green] table[row sep=crcr]{%
x	y\\
172.610701467813	0\\
};

\addplot[only marks, mark=*, mark options={}, mark size=1.5000pt, draw=black, fill=red] table[row sep=crcr]{%
x	y\\
172.364297271163	0\\
};

\end{axis}
\end{tikzpicture}%
		\caption{$p_{H_2^{\text{out}}}(h_2)$}
		\label{fig:pdf_posterior_lcorr_1}
	\end{subfigure}\\
	\begin{subfigure}{.5\linewidth}
		\centering
		\tikzsetnextfilename{pdf_posterior_kappa_1}
%
\begin{tikzpicture}

\begin{axis}[%
width=1.269\figureheight,
height=\figureheight,
at={(0\figureheight,0\figureheight)},
scale only axis,
xmin=11.3,
xmax=13.2,
xlabel={$h_3$ [GPa]},
ymin=0,
ymax=1.8e-09,
ylabel={$p_{H_3^{\text{out}}}(h_3)$},
xmajorgrids,
ymajorgrids,
legend style={legend cell align=left, align=left
}
]
\addplot [color=blue, line width=1.0pt]
  table[row sep=crcr]{%
11.0244540681982	2.93366223849398e-16\\
11.0485675888252	9.85847661951639e-15\\
11.0726811094522	2.59237541644822e-14\\
11.0967946300791	5.33740322857353e-15\\
11.1209081507061	1.15679931300414e-15\\
11.1450216713331	2.5742925374479e-14\\
11.16913519196	7.21038383743811e-14\\
11.193248712587	9.61060402787737e-14\\
11.217362233214	9.06457476255811e-14\\
11.2414757538409	7.92277004646251e-14\\
11.2655892744679	7.17788556616114e-14\\
11.2897027950949	1.37805956666467e-13\\
11.3138163157218	3.29924935936987e-13\\
11.3379298363488	3.17960941503723e-13\\
11.3620433569758	4.01741273697392e-13\\
11.3861568776027	6.09560975405055e-13\\
11.4102703982297	1.14982618030773e-12\\
11.4343839188567	1.79493901213686e-12\\
11.4584974394836	2.32946420373155e-12\\
11.4826109601106	3.41761014550427e-12\\
11.5067244807376	4.63630634762013e-12\\
11.5308380013645	5.66840605809734e-12\\
11.5549515219915	8.20890878063701e-12\\
11.5790650426185	1.2055602996995e-11\\
11.6031785632454	1.531098439537e-11\\
11.6272920838724	2.0710196338486e-11\\
11.6514056044994	2.96564715728354e-11\\
11.6755191251263	4.06862390595068e-11\\
11.6996326457533	5.30634865585208e-11\\
11.7237461663803	6.9941179065153e-11\\
11.7478596870073	9.13225672307416e-11\\
11.7719732076342	1.17926410517019e-10\\
11.7960867282612	1.49606695826965e-10\\
11.8202002488882	1.86554484366559e-10\\
11.8443137695151	2.35347841008449e-10\\
11.8684272901421	2.93341895515324e-10\\
11.8925408107691	3.56859458012431e-10\\
11.916654331396	4.26021273497159e-10\\
11.940767852023	5.07478881875972e-10\\
11.96488137265	6.01316642538475e-10\\
11.9889948932769	6.99068271695332e-10\\
12.0131084139039	8.00316832283738e-10\\
12.0372219345309	9.14451421329096e-10\\
12.0613354551578	1.0369366868574e-09\\
12.0854489757848	1.14545352166014e-09\\
12.1095624964118	1.25517649379657e-09\\
12.1336760170387	1.36075315744212e-09\\
12.1577895376657	1.46263766094917e-09\\
12.1819030582927	1.55351345313364e-09\\
12.2060165789196	1.62411275470031e-09\\
12.2301300995466	1.6859051048701e-09\\
12.2542436201736	1.73381721146983e-09\\
12.2783571408005	1.76686486070538e-09\\
12.3024706614275	1.7606459222173e-09\\
12.3265841820545	1.73964514801353e-09\\
12.3506977026814	1.70180792104804e-09\\
12.3748112233084	1.6518258507432e-09\\
12.3989247439354	1.58318501438896e-09\\
12.4230382645623	1.5081496827427e-09\\
12.4471517851893	1.4249393314104e-09\\
12.4712653058163	1.31456725300514e-09\\
12.4953788264432	1.19681170978958e-09\\
12.5194923470702	1.08647735101574e-09\\
12.5436058676972	9.74463817961567e-10\\
12.5677193883241	8.59508046001038e-10\\
12.5918329089511	7.54922115122545e-10\\
12.6159464295781	6.55623038941555e-10\\
12.640059950205	5.58927420846268e-10\\
12.664173470832	4.67427927503018e-10\\
12.688286991459	3.91583426011316e-10\\
12.7124005120859	3.2701245884327e-10\\
12.7365140327129	2.69542138417796e-10\\
12.7606275533399	2.20280478228064e-10\\
12.7847410739669	1.78259325190053e-10\\
12.8088545945938	1.40739274886871e-10\\
12.8329681152208	1.070374765841e-10\\
12.8570816358478	8.10187635023058e-11\\
12.8811951564747	6.3835242713648e-11\\
12.9053086771017	4.85401626031672e-11\\
12.9294221977287	3.49867748012158e-11\\
12.9535357183556	2.54561854049289e-11\\
12.9776492389826	1.8125875451227e-11\\
13.0017627596096	1.30962693702343e-11\\
13.0258762802365	1.00861113162658e-11\\
13.0499898008635	7.25523083351449e-12\\
13.0741033214905	5.41500064119668e-12\\
13.0982168421174	3.48153511845106e-12\\
13.1223303627444	2.50456904471888e-12\\
13.1464438833714	1.99036174555595e-12\\
13.1705574039983	1.14439037469141e-12\\
13.1946709246253	6.09350069460614e-13\\
13.2187844452523	5.23149356132775e-13\\
13.2428979658792	2.92438365668921e-13\\
13.2670114865062	1.91777276898395e-13\\
13.2911250071332	1.03050910712636e-13\\
13.3152385277601	5.49995647115256e-14\\
13.3393520483871	1.99172569929543e-14\\
13.3634655690141	5.23025844202576e-14\\
13.387579089641	1.86578414995398e-14\\
13.411692610268	5.24285345364318e-16\\
};

\addplot[area legend, draw=blue, fill=blue, fill opacity=0.2] table[row sep=crcr]{%
11.8684272901421	2.93341895515324e-10\\
11.8925408107691	3.56859458012431e-10\\
11.916654331396	4.26021273497159e-10\\
11.940767852023	5.07478881875972e-10\\
11.96488137265	6.01316642538475e-10\\
11.9889948932769	6.99068271695332e-10\\
12.0131084139039	8.00316832283738e-10\\
12.0372219345309	9.14451421329096e-10\\
12.0613354551578	1.0369366868574e-09\\
12.0854489757848	1.14545352166014e-09\\
12.1095624964118	1.25517649379657e-09\\
12.1336760170387	1.36075315744212e-09\\
12.1577895376657	1.46263766094917e-09\\
12.1819030582927	1.55351345313364e-09\\
12.2060165789196	1.62411275470031e-09\\
12.2301300995466	1.6859051048701e-09\\
12.2542436201736	1.73381721146983e-09\\
12.2783571408005	1.76686486070538e-09\\
12.3024706614275	1.7606459222173e-09\\
12.3265841820545	1.73964514801353e-09\\
12.3506977026814	1.70180792104804e-09\\
12.3748112233084	1.6518258507432e-09\\
12.3989247439354	1.58318501438896e-09\\
12.4230382645623	1.5081496827427e-09\\
12.4471517851893	1.4249393314104e-09\\
12.4712653058163	1.31456725300514e-09\\
12.4953788264432	1.19681170978958e-09\\
12.5194923470702	1.08647735101574e-09\\
12.5436058676972	9.74463817961567e-10\\
12.5677193883241	8.59508046001038e-10\\
12.5918329089511	7.54922115122545e-10\\
12.6159464295781	6.55623038941555e-10\\
12.640059950205	5.58927420846268e-10\\
12.664173470832	4.67427927503018e-10\\
12.688286991459	3.91583426011316e-10\\
12.7124005120859	3.2701245884327e-10\\
12.7365140327129	2.69542138417796e-10\\
}
\closedcycle;

\addplot[only marks, mark=diamond*, mark options={}, mark size=2.0000pt, draw=black, fill=blue] table[row sep=crcr]{%
x	y\\
12.2971280059276	0\\
};

\addplot[only marks, mark=*, mark options={}, mark size=1.5000pt, draw=black, fill=black!20!green] table[row sep=crcr]{%
x	y\\
12.3217601212031	0\\
};

\addplot[only marks, mark=*, mark options={}, mark size=1.5000pt, draw=black, fill=red] table[row sep=crcr]{%
x	y\\
12.3975482376603	0\\
};

\end{axis}
\end{tikzpicture}%
		\caption{$p_{H_3^{\text{out}}}(h_3)$}
		\label{fig:pdf_posterior_kappa_1}
	\end{subfigure}\hfill
	\begin{subfigure}{.5\linewidth}
		\centering
		\tikzsetnextfilename{pdf_posterior_mu_1}
%
\begin{tikzpicture}

\begin{axis}[%
width=1.269\figureheight,
height=\figureheight,
at={(0\figureheight,0\figureheight)},
scale only axis,
xmin=4.58,
xmax=4.82,
xlabel={$h_4$ [GPa]},
ymin=0,
ymax=1.5e-08,
ylabel={$p_{H_4^{\text{out}}}(h_4)$},
xmajorgrids,
ymajorgrids,
legend style={legend cell align=left, align=left,
at={(0.8,0.5)},anchor=center
}
]
\addplot [color=blue, line width=1.0pt]
  table[row sep=crcr]{%
4.56805801670349	2.48936227321723e-15\\
4.57160242423061	1.3467622249587e-13\\
4.57514683175772	1.39038834561514e-13\\
4.57869123928483	1.39924119834645e-13\\
4.58223564681194	9.58962113390384e-13\\
4.58578005433905	2.38413431983119e-12\\
4.58932446186617	3.59603417859167e-12\\
4.59286886939328	5.18184050959226e-12\\
4.59641327692039	8.69965236660273e-12\\
4.5999576844475	1.6849735324887e-11\\
4.60350209197461	2.69010509809178e-11\\
4.60704649950173	4.1934650929159e-11\\
4.61059090702884	7.41721242115657e-11\\
4.61413531455595	1.24544479626299e-10\\
4.61767972208306	1.81024429259116e-10\\
4.62122412961017	2.66193436261625e-10\\
4.62476853713729	4.13669959473792e-10\\
4.6283129446644	5.85935976090582e-10\\
4.63185735219151	8.23860104191299e-10\\
4.63540175971862	1.13833692828064e-09\\
4.63894616724573	1.52102815453968e-09\\
4.64249057477285	2.02671166567377e-09\\
4.64603498229996	2.67835025837355e-09\\
4.64957938982707	3.45125248124922e-09\\
4.65312379735418	4.32564030477995e-09\\
4.65666820488129	5.29865280624312e-09\\
4.66021261240841	6.41047871229704e-09\\
4.66375701993552	7.66461587877813e-09\\
4.66730142746263	8.95586068754262e-09\\
4.67084583498974	1.02729866875143e-08\\
4.67439024251685	1.14489204198429e-08\\
4.67793465004397	1.25544163805452e-08\\
4.68147905757108	1.34303148416716e-08\\
4.68502346509819	1.41465415460474e-08\\
4.6885678726253	1.46414081638699e-08\\
4.69211228015241	1.4893421682168e-08\\
4.69565668767953	1.48662713251395e-08\\
4.69920109520664	1.46746161184497e-08\\
4.70274550273375	1.4134273031298e-08\\
4.70628991026086	1.34687769585791e-08\\
4.70983431778798	1.24390762641961e-08\\
4.71337872531509	1.14524294537115e-08\\
4.7169231328422	1.02814364475376e-08\\
4.72046754036931	9.07391548239649e-09\\
4.72401194789642	7.95791546805753e-09\\
4.72755635542354	6.90272895819449e-09\\
4.73110076295065	5.816072229543e-09\\
4.73464517047776	4.88492909071894e-09\\
4.73818957800487	4.01987019780792e-09\\
4.74173398553198	3.29344982559532e-09\\
4.7452783930591	2.62468043291477e-09\\
4.74882280058621	2.059901814533e-09\\
4.75236720811332	1.61757893486306e-09\\
4.75591161564043	1.26723686841996e-09\\
4.75945602316754	9.76814378816748e-10\\
4.76300043069466	7.35594640267315e-10\\
4.76654483822177	5.65659685191223e-10\\
4.77008924574888	4.16926164324875e-10\\
4.77363365327599	3.138112562069e-10\\
4.7771780608031	2.25376175986722e-10\\
4.78072246833022	1.6479335472253e-10\\
4.78426687585733	1.19560681027474e-10\\
4.78781128338444	8.67491392154833e-11\\
4.79135569091155	6.33290105591822e-11\\
4.79490009843866	4.51924162217979e-11\\
4.79844450596578	3.35325402931762e-11\\
4.80198891349289	2.45938933869339e-11\\
4.80553332102	2.08231247522917e-11\\
4.80907772854711	1.60938268665823e-11\\
4.81262213607422	1.05509023558344e-11\\
4.81616654360134	7.4449091233804e-12\\
4.81971095112845	5.2910084931395e-12\\
4.82325535865556	3.10200097588952e-12\\
4.82679976618267	2.78500305786109e-12\\
4.83034417370978	2.05954646172473e-12\\
4.8338885812369	8.16375412543701e-13\\
4.83743298876401	1.5178558969002e-12\\
4.84097739629112	8.40849849257119e-13\\
4.84452180381823	3.67935540378786e-13\\
4.84806621134534	1.28869502458949e-13\\
4.85161061887246	2.80265715838008e-14\\
4.85515502639957	3.16755717835669e-13\\
4.85869943392668	2.16434485493846e-13\\
4.86224384145379	6.60057958874359e-15\\
4.8657882489809	1.695511256374e-27\\
4.86933265650802	2.19334663323743e-21\\
4.87287706403513	5.38925618335988e-17\\
4.87642147156224	2.51515843160318e-14\\
4.87996587908935	2.22957179683161e-13\\
4.88351028661647	4.20881670847341e-14\\
4.88705469414358	1.84089552026339e-13\\
4.89059910167069	3.31755809410617e-13\\
4.8941435091978	4.39890473410264e-14\\
4.89768791672491	1.55651490296668e-16\\
4.90123232425203	7.05100506806004e-24\\
4.90477673177914	1.00135603472437e-18\\
4.90832113930625	2.70110193127259e-15\\
4.91186554683336	1.38391172470657e-13\\
4.91540995436047	1.34676172556534e-13\\
4.91895436188759	2.48936227310445e-15\\
};
\addlegendentry{Output pdf}

\addplot[area legend, draw=blue, fill=blue, fill opacity=0.2] table[row sep=crcr]{%
4.64603498229996	2.67835025837355e-09\\
4.64957938982707	3.45125248124922e-09\\
4.65312379735418	4.32564030477995e-09\\
4.65666820488129	5.29865280624312e-09\\
4.66021261240841	6.41047871229704e-09\\
4.66375701993552	7.66461587877813e-09\\
4.66730142746263	8.95586068754262e-09\\
4.67084583498974	1.02729866875143e-08\\
4.67439024251685	1.14489204198429e-08\\
4.67793465004397	1.25544163805452e-08\\
4.68147905757108	1.34303148416716e-08\\
4.68502346509819	1.41465415460474e-08\\
4.6885678726253	1.46414081638699e-08\\
4.69211228015241	1.4893421682168e-08\\
4.69565668767953	1.48662713251395e-08\\
4.69920109520664	1.46746161184497e-08\\
4.70274550273375	1.4134273031298e-08\\
4.70628991026086	1.34687769585791e-08\\
4.70983431778798	1.24390762641961e-08\\
4.71337872531509	1.14524294537115e-08\\
4.7169231328422	1.02814364475376e-08\\
4.72046754036931	9.07391548239649e-09\\
4.72401194789642	7.95791546805753e-09\\
4.72755635542354	6.90272895819449e-09\\
4.73110076295065	5.816072229543e-09\\
4.73464517047776	4.88492909071894e-09\\
4.73818957800487	4.01987019780792e-09\\
4.74173398553198	3.29344982559532e-09\\
4.7452783930591	2.62468043291477e-09\\
4.74882280058621	2.059901814533e-09\\
}
\closedcycle;
\addlegendentry{Output $95\%$ confidence interval}

\addplot[only marks, mark=diamond*, mark options={}, mark size=2.0000pt, draw=black, fill=blue] table[row sep=crcr]{%
x	y\\
4.69531851866888	0\\
};
\addlegendentry{Output mean value}

\addplot[only marks, mark=*, mark options={}, mark size=1.5000pt, draw=black, fill=black!20!green] table[row sep=crcr]{%
x	y\\
4.69290490675317	0\\
};
\addlegendentry{Output value}

\addplot[only marks, mark=*, mark options={}, mark size=1.5000pt, draw=black, fill=red] table[row sep=crcr]{%
x	y\\
4.67151710393869	0\\
};
\addlegendentry{Target value}

\end{axis}
\end{tikzpicture}%
		\caption{$p_{H_4^{\text{out}}}(h_4)$}
		\label{fig:pdf_posterior_mu_1}
	\end{subfigure}
	\caption{Synthetic data: probability density functions $p_{H_1^{\text{out}}}$, $p_{H_2^{\text{out}}}$, $p_{H_3^{\text{out}}}$ and $p_{H_4^{\text{out}}}$ of random variables $H_1^{\text{out}}$, $H_2^{\text{out}}$, $H_3^{\text{out}}$ and $H_4^{\text{out}}$, respectively, obtained for a given input vector $\qb^{\text{obs}}$ and for a given input uncertainty level $s = 0.01 = 1\%$, with the output $95\%$ confidence intervals $I^{\text{out}}$ (blue areas), the output mean values $\underline{\hb}^{\text{out}}$ (blue diamonds), the output values $\hb^{\text{out}}$ (green circles) and the corresponding target values $\hb^{\text{target}}$ (red circles)}
	\label{fig:pdf_posterior}
\end{figure}
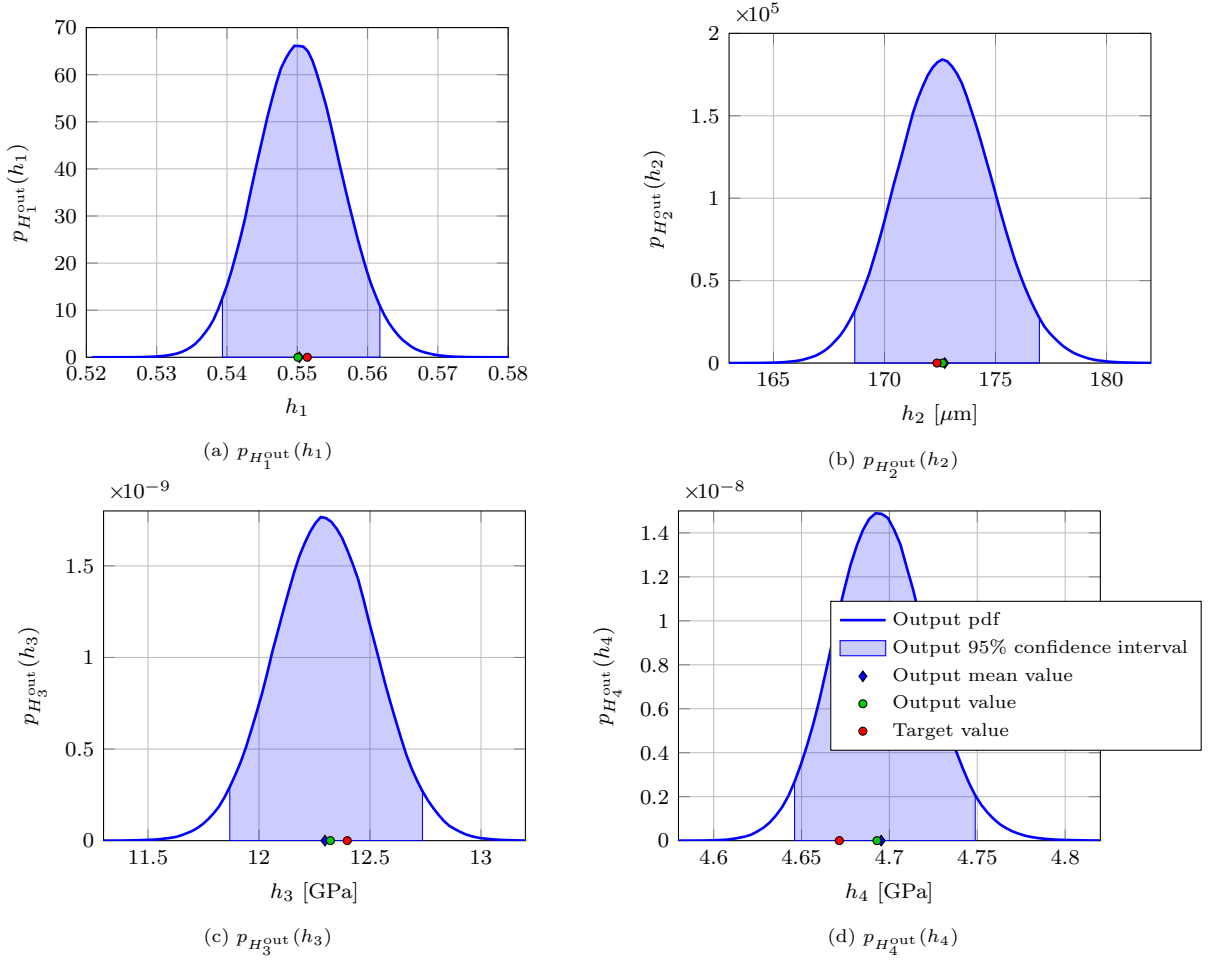

\begin{figure}[h!]
	\centering
	\begin{subfigure}{.5\linewidth}
		\centering
		\tikzsetnextfilename{pdfs_posterior_delta}
		\input{pdfs_posterior_delta}
		\caption{$p_{H_1^{\text{out}}}(h_1)$}
		\label{fig:pdfs_posterior_delta}
	\end{subfigure}\hfill
	\begin{subfigure}{.5\linewidth}
		\centering
		\tikzsetnextfilename{pdfs_posterior_lcorr}
		\input{pdfs_posterior_lcorr}
		\caption{$p_{H_2^{\text{out}}}(h_2)$}
		\label{fig:pdfs_posterior_lcorr}
	\end{subfigure}\\
	\begin{subfigure}{.45\linewidth}
		\centering
		\tikzsetnextfilename{pdfs_posterior_kappa}
		\input{pdfs_posterior_kappa}
		\caption{$p_{H_3^{\text{out}}}(h_3)$}
		\label{fig:pdfs_posterior_kappa}
	\end{subfigure}\hfill
	\begin{subfigure}{.55\linewidth}
		\centering
		\tikzsetnextfilename{pdfs_posterior_mu}
		\input{pdfs_posterior_mu}
		\caption{$p_{H_4^{\text{out}}}(h_4)$}
		\label{fig:pdfs_posterior_mu}
	\end{subfigure}
	\caption{Synthetic data: probability density functions $p_{H_1^{\text{out}}}$, $p_{H_2^{\text{out}}}$, $p_{H_3^{\text{out}}}$ and $p_{H_4^{\text{out}}}$ of random variables $H_1^{\text{out}}$, $H_2^{\text{out}}$, $H_3^{\text{out}}$ and $H_4^{\text{out}}$, respectively, obtained for a given input vector $\qb^{\text{obs}}$ and for different values of input uncertainty level $s \in \set{0.01,0.02,0.03,0.04,0.05}$, with the output mean values $\underline{\hb}^{\text{out}}$ (colored diamonds), the output values $\hb^{\text{out}}$ (green circles) and the corresponding target values $\hb^{\text{target}}$ (red circles)}
	\label{fig:pdfs_posterior}
\end{figure}

\section{Numerical example on real experimental data for a biological material}\label{sec:results_real_data}

We now consider a given input vector $\qb^{\text{obs}}$ of observed quantities of interest coming from experimental measurements of 2D displacement fields obtained from a single static vertical uniaxial compression test performed on a unique cubic specimen (with dimensions $1\times 1 \times 1$~cm$^3$) made of a biological tissue (beef femur cortical bone) and monitored by 2D digital image correlation (DIC) on one observed side of the cubic specimen corresponding to a 2D square domain $\Omega^{\text{macro}}$ with macroscopic dimensions $1\times 1$~cm$^2$.
Such experimental kinematic field measurements have been carried out in \cite{Ngu16} and already used in \cite{Ngu15,Zha20} for identifying the apparent elastic properties of bovine cortical bone at mesoscale.
The experimental test configuration corresponds to the numerical one described in Figure~\ref{fig:boundary_value_problem}.
The interested reader can refer to \cite{Ngu16} for technical details concerning the experimental setup of the mechanical test (specimen preparation, test bench, test procedure, optical measuring instrument, optical image acquisition system and DIC method) for obtaining the 2D displacement field measurements.
The experimental quantities of interest $\qb^{\text{obs}} = (q_1^{\text{obs}},\dots,q_9^{\text{obs}})$ have been derived from the experimental fine-scale displacement field computed on a 2D square subdomain $\Omega^{\text{meso}} \subset \Omega^{\text{macro}}$ with mesoscopic dimensions $1\times 1$~cm$^2$ (located near the center of the observed face of the cubic sample to limit edge effects) and discretized with a fine regular grid of $100 \times 100$ quadrangular elements with uniform element size $h^{\text{meso}}$ = $10~\upmu\text{m} = 10^{-5}$~m in each spatial direction. The experimental linearized strain field has been directly computed from the experimentally measured displacement field by using classical interpolation techniques and then used to compute the three first experimental quantities of interest $q_1^{\text{obs}}$, $q_2^{\text{obs}}$ and $q_3^{\text{obs}}$, where $q_1^{\text{obs}}$ corresponds to the spatial dispersion coefficient quantifying the level of spatial fluctuations of the linearized experimental strain field around its spatial average over $\Omega^{\text{meso}}$, while $q_2^{\text{obs}}$ and $q_3^{\text{obs}}$ correspond to the two characteristic lengths along the two spatial directions $x_1$ and $x_2$ characterizing the spatial fluctuations of the linearized experimental strain field around its spatial average over $\Omega^{\text{meso}}$ and numerically computed using a usual signal processing method.
The effective compliance matrix $[S^{\text{eff}}_{\text{obs}}] \in \Mbb_3^+(\Rbb)$ has experimentally been identified in previous works \cite{Ngu15,Zha20} by solving a classical inverse problem at coarse scale (macroscale) using experimental coarse-scale displacement field measurements at macroscale. More precisely, since the observed face of the cubic sample corresponds to a plane of isotropy of the material, $[S^{\text{eff}}_{\text{obs}}]$ is completely characterized and parameterized by the bulk modulus $\kappa$ and the shear modulus $\mu$ of the isotropic elastic material at macroscale. The optimal values of $\kappa$ and $\mu$ have been identified by minimizing the spatial average over macroscopic domain $\Omega^{\text{macro}}$ of the distance (defined with respect to the Frobenius norm) between the strain field (parameterized by $(\kappa,\mu)$) computed numerically by solving the deterministic linear elasticity boundary value problem (that models the experimental test configuration) at macroscale and the strain field measured experimentally at macroscale.
The upper Cholesky factor $[L^{\text{eff}}_{\text{obs}}]$ of $[S^{\text{eff}}_{\text{obs}}]$ has then been computed by performing the Cholesky factorization of $[S^{\text{eff}}_{\text{obs}}] = [L^{\text{eff}}_{\text{obs}}]^T [L^{\text{eff}}_{\text{obs}}]$ in order to derive the last vector-valued experimental quantity of interest $(q_4^{\text{obs}},\dots,q_9^{\text{obs}})$, with $q^{\text{obs}}_4 = \log([L^{\text{eff}}_{\text{obs}}]_{11})$, $q^{\text{obs}}_5 = [L^{\text{eff}}_{\text{obs}}]_{12}$, $q^{\text{obs}}_6 = [L^{\text{eff}}_{\text{obs}}]_{13}$, $q^{\text{obs}}_7 = \log([L^{\text{eff}}_{\text{obs}}]_{22})$, $q^{\text{obs}}_8 = [L^{\text{eff}}_{\text{obs}}]_{23}$ and $q^{\text{obs}}_9 = \log([L^{\text{eff}}_{\text{obs}}]_{33})$, as presented in Section~\ref{sec:definition_HFCMM}.

As for the previous validation example on synthetic data, the trained neural network is first used to compute the output vector $\hb^{\text{out}}$ for the experimentally observed input vector $\qb^{\text{obs}}$ without introducing uncertainties. Then, in order to quantify the robustness of the network output vector $\hb^{\text{out}}$ with respect to the uncertainties on the input vector $\qb^{\text{obs}}$, we consider the input random vector $\Qb^{\text{obs}}$ whose probabilistic model has been introduced in Section~\ref{sec:robustness} and which is parameterized by the four-dimensional vector-valued parameter $\ssb=(s_0,s_1,s_2,s^{\text{eff}})$ with $s_0 = s_1 = s_2 = s^{\text{eff}} = s$, in which $s$ is the input uncertainty level allowing the level of statistical fluctuations of $\Qb^{\text{obs}}$ around its mean value $\qb^{\text{obs}}$ to be controlled. In practice, the value of $s$ is related to the knowledge of the experimental errors and should be driven by the expertise of the experimenter. For the considered application on real bovine cortical bone data, a reasonable value for $s$ is of the order of few percents. In the following, we consider five different values for input uncertainty level $s \in \set{0.01,0.02,0.03,0.04,0.05}$ and for each of them, $N_s = 10^6$ independent realizations $\qb^{\text{obs},(1)},\dots,\qb^{\text{obs},(N_s)}$ of input random vector $\Qb^{\text{obs}}$ are generated, then presented and applied to the trained neural network in order to compute the $N_s$ corresponding independent realizations $\hb^{\text{out},(1)},\dots,\hb^{\text{out},(N_s)}$ of output random vector $\Hb^{\text{out}}$. The values of output mean vector $\hb^{\text{out}}$ and the bounds of the confidence intervals of each of the components of $\Hb^{\text{out}}$ are then computed by using classical empirical estimates.

\begin{figure}[h!]
	\centering
	\begin{subfigure}{.5\linewidth}
		\centering
		\tikzsetnextfilename{robustness_delta_bone}
%
\begin{tikzpicture}

\begin{axis}[%
width=1.269\figureheight,
height=\figureheight,
at={(0\figureheight,0\figureheight)},
scale only axis,
xmin=0,
xmax=0.05,
xlabel={$s$},
ymin=0.4,
ymax=0.85,
ylabel={$h_1^{\text{out}}$},
xmajorgrids,
ymajorgrids,
legend style={legend cell align=left, align=left
}
]

\addplot[area legend, draw=blue, fill=blue, fill opacity=0.2]
table[row sep=crcr] {%
x	y\\
0	0.610590072451957\\
0.01	0.632178163901665\\
0.02	0.660885744832946\\
0.03	0.702745607494262\\
0.04	0.760601005783185\\
0.05	0.826403347577366\\
0.05	0.437008378094832\\
0.04	0.493222895472763\\
0.03	0.540345951713182\\
0.02	0.570059737253135\\
0.01	0.591406299047722\\
0	0.610590072451957\\
}--cycle;

\addplot [color=blue, line width=1.0pt]
  table[row sep=crcr]{%
0	0.610590072451957\\
0.01	0.610087410745185\\
0.02	0.609124630454363\\
0.03	0.608849816015809\\
0.04	0.610320988289987\\
0.05	0.614096153665736\\
};

\end{axis}
\end{tikzpicture}%
		\caption{dispersion parameter $H_1^{\text{out}}$}
		\label{fig:robustness_delta_bone}
	\end{subfigure}\hfill
	\begin{subfigure}{.5\linewidth}
		\centering
		\tikzsetnextfilename{robustness_lcorr_bone}
%
\begin{tikzpicture}

\begin{axis}[%
width=1.269\figureheight,
height=\figureheight,
at={(0\figureheight,0\figureheight)},
scale only axis,
xmin=0,
xmax=0.05,
xlabel={$s$},
ymin=0,
ymax=90,
ylabel={$h_2^{\text{out}}$ [$\mu$m]},
xmajorgrids,
ymajorgrids,
legend style={legend cell align=left, align=left
}
]

\addplot[area legend, draw=blue, fill=blue, fill opacity=0.2]
table[row sep=crcr] {%
x	y\\
0	65.9062206547429\\
0.01	70.4741383214648\\
0.02	74.6752858952744\\
0.03	77.3552000007674\\
0.04	81.1280165814683\\
0.05	88.3245846539085\\
0.05	5.66477512191514\\
0.04	26.9535533872035\\
0.03	44.4719876399295\\
0.02	56.4119206318047\\
0.01	62.6396459806784\\
0	65.9062206547429\\
}--cycle;

\addplot [color=blue, line width=1.0pt]
  table[row sep=crcr]{%
0	65.9062206547429\\
0.01	65.8907628708782\\
0.02	65.4853549406661\\
0.03	64.0257567366226\\
0.04	61.4837566371804\\
0.05	58.4556095170861\\
};

\end{axis}
\end{tikzpicture}%
		\caption{spatial correlation length $H_2^{\text{out}}$}
		\label{fig:robustness_lcorr_bone}
	\end{subfigure}\\
	\begin{subfigure}{.5\linewidth}
		\centering
		\tikzsetnextfilename{robustness_kappa_bone}
%
\begin{tikzpicture}

\begin{axis}[%
width=1.269\figureheight,
height=\figureheight,
at={(0\figureheight,0\figureheight)},
scale only axis,
xmin=0,
xmax=0.05,
xlabel={$s$},
ymin=7,
ymax=16,
ylabel={$h_3^{\text{out}}$ [GPa]},
xmajorgrids,
ymajorgrids,
legend style={legend cell align=left, align=left
}
]

\addplot[area legend, draw=blue, fill=blue, fill opacity=0.2]
table[row sep=crcr] {%
x	y\\
0	10.4479618095717\\
0.01	11.0223984487387\\
0.02	11.689902789524\\
0.03	12.5737881703471\\
0.04	14.0754456700555\\
0.05	15.9801047227411\\
0.05	7.18370256705209\\
0.04	8.0047295510278\\
0.03	8.67446014698393\\
0.02	9.28618134418491\\
0.01	9.88077496799611\\
0	10.4479618095717\\
}--cycle;

\addplot [color=blue, line width=1.0pt]
  table[row sep=crcr]{%
0	10.4479618095717\\
0.01	10.459556850696\\
0.02	10.4993585974174\\
0.03	10.5910068741323\\
0.04	10.7694856323676\\
0.05	11.0260292670537\\
};

\end{axis}
\end{tikzpicture}%
		\caption{mean bulk modulus $H_3^{\text{out}}$}
		\label{fig:robustness_kappa_bone}
	\end{subfigure}\hfill
	\begin{subfigure}{.5\linewidth}
		\centering
		\tikzsetnextfilename{robustness_mu_bone}
%
\begin{tikzpicture}

\begin{axis}[%
width=1.269\figureheight,
height=\figureheight,
at={(0\figureheight,0\figureheight)},
scale only axis,
xmin=0,
xmax=0.05,
xlabel={$s$},
ymin=4.4,
ymax=5.6,
ylabel={$h_4^{\text{out}}$ [GPa]},
xmajorgrids,
ymajorgrids,
legend style={legend cell align=left, align=left
},
legend pos=north east
]

\addplot[area legend, draw=blue, fill=blue, fill opacity=0.2]
table[row sep=crcr] {%
x	y\\
0	4.59840650118907\\
0.01	4.65580471880419\\
0.02	4.78117086795744\\
0.03	5.06619743789562\\
0.04	5.39056489461216\\
0.05	5.59722400383443\\
0.05	4.40073994154104\\
0.04	4.43443438257094\\
0.03	4.4701510797109\\
0.02	4.51074377927221\\
0.01	4.55292698249516\\
0	4.59840650118907\\
}--cycle;
\addlegendentry{Output $95\%$ confidence interval}

\addplot [color=blue, line width=1.0pt]
  table[row sep=crcr]{%
0	4.59840650118907\\
0.01	4.60209566557382\\
0.02	4.61870226846785\\
0.03	4.65711385186122\\
0.04	4.71393468016881\\
0.05	4.77820120386092\\
};
\addlegendentry{Output mean value}

\end{axis}
\end{tikzpicture}%
		\caption{mean shear modulus $H_4^{\text{out}}$}
		\label{fig:robustness_mu_bone}
	\end{subfigure}
	\caption{Real experimental data: evolutions of the output mean values $\underline{\hb}^{\text{out}}$ (blue curve) and the $95\%$ confidence intervals $I^{\text{out}}$ (blue areas) of random variables $H_1^{\text{out}}$, $H_2^{\text{out}}$, $H_3^{\text{out}}$ and $H_4^{\text{out}}$, respectively, with respect to input uncertainty level $s$, for a given input vector $\qb^{\text{obs}}$}
	\label{fig:robustness_bone}
\end{figure}
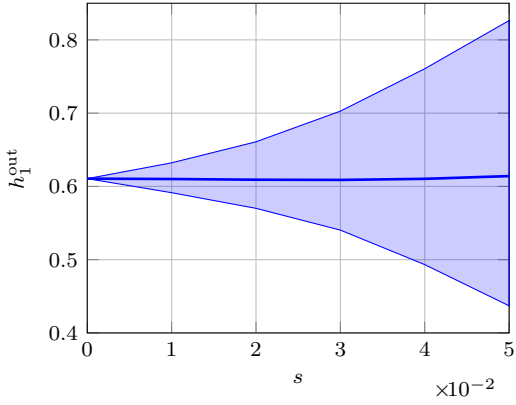
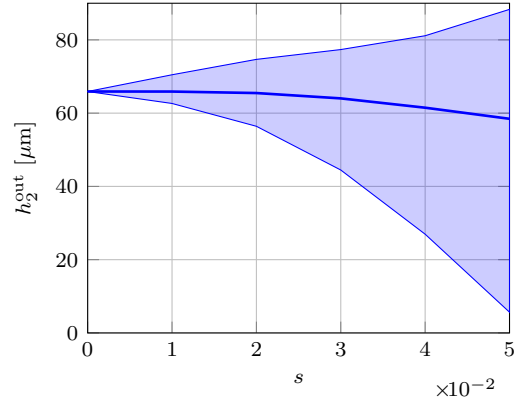
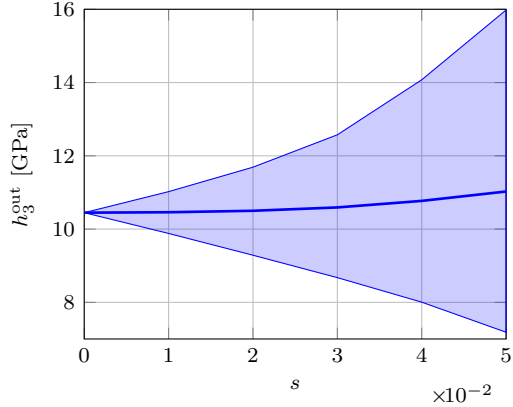
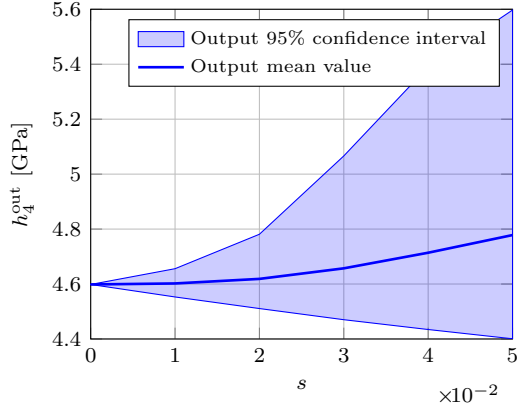

Table~\ref{tab:identification_bone} reports the values of output vector $\hb^{\text{out}}$ (corresponding to an input uncertainty level $s=0$) and the ones of output mean vector $\underline{\hb}^{\text{out}}$ as well as the bounds of the output $95\%$ confidence intervals of $\Hb^{\text{out}}$ for the different values of input uncertainty level $s \in \set{0.01,0.02,0.03,0.04,0.05}$. As a complement, Figure~\ref{fig:robustness_bone} represents the graphs of the mean output values $\underline{h}_1^{\text{out}}$, $\underline{h}_2^{\text{out}}$, $\underline{h}_3^{\text{out}}$ and $\underline{h}_4^{\text{out}}$ and the corresponding confidence regions (with a probability level $95\%$) of $H_1^{\text{out}}$, $H_2^{\text{out}}$, $H_3^{\text{out}}$ and $H_4^{\text{out}}$, respectively, with respect to the input uncertainty level $s$. We observe that the mean value $\underline{h}_1^{\text{out}}$ of $H_1^{\text{out}}$ is relatively insensitive to the input uncertainty level $s$, while the mean values $\underline{h}_2^{\text{out}}$, $\underline{h}_3^{\text{out}}$ and $\underline{h}_4^{\text{out}}$ of $H_2^{\text{out}}$, $H_3^{\text{out}}$ and $H_4^{\text{out}}$ slightly vary and deviate from the respective output values $h_2^{\text{out}}$, $h_3^{\text{out}}$ and $h_4^{\text{out}}$ (obtained by using the trained neural network with an input uncertainty level $s=0$) as the input uncertainty level $s$ increases. It should be noted that the values of the output vector $\hb^{\text{out}} = (0.6106,65.906,10.448,4.598)$ in $[-]\times[\mu\text{m}]\times[\text{GPa}]\times[\text{GPa}]$ obtained for the input vector $\qb^{\text{obs}}$ (without considering any input uncertainties) are close to the values of the optimal vector $\hb^{\text{opt}} = (0.533,61.111,10.500,4.667)$ in $([-],[\mu\text{m}],[\text{GPa}],[\text{GPa}])$ obtained in the previous work \cite{Zha20} by solving a computationally expensive multi-objective optimization problem using a fixed-point iterative algorithm with the same experimental measurements as those used in the present work. The identified values obtained with the previous method in \cite{Zha20} result from a compromise between computational efficiency and numerical accuracy and are therefore less accurate than the ones obtained with the ANN-based identification method proposed in this work. The network output values are then in agreement with the identified values already published in the literature for this type of biological tissue (bovine cortical bone).

\begin{table}[h!]
\caption{Real experimental data: output vector $\hb^{\text{out}}$, output mean vector $\underline{\hb}^{\text{out}}$ and output $95\%$ confidence interval $I^{\text{out}}$ obtained for a given input vector $\qb^{\text{obs}}$ and for different values of input uncertainty level $s$}
\label{tab:identification_bone}
\centering
\begin{tabular}{|cc|c|c|c|c|} \hline
 & & $h_1$ & $h_2$ [$\mu$m] & $h_3$ [GPa] & $h_4$ [GPa] \\ \hline
Output vector $\hb^{\text{out}}$ & $s=0\%$ & $0.6106$ & $65.906$ & $10.448$ & $4.598$ \\ \cline{2-6}
\multirow{5}{*}{Output mean vector $\underline{\hb}^{\text{out}}$} & $s=1\%$ & $0.6101$ & $65.891$ & $10.460$ & $4.602$ \\
 & $s=2\%$ & $0.6091$ & $65.485$ & $10.499$ & $4.619$ \\
 & $s=3\%$ & $0.6089$ & $64.026$ & $10.591$ & $4.657$ \\
 & $s=4\%$ & $0.6103$ & $61.484$ & $10.770$ & $4.714$ \\
 & $s=5\%$ & $0.6141$ & $58.456$ & $11.026$ & $4.778$ \\ \hline
\multirow{4}{*}{Output $95\%$} & $s=1\%$ & $\intervalcc{0.5914}{0.6322}$ & $\intervalcc{62.640}{70.474}$ & $\intervalcc{9.881}{11.022}$ & $\intervalcc{4.553}{4.656}$ \\
\multirow{4}{*}{confidence interval $I^{\text{out}}$} & $s=2\%$ & $\intervalcc{0.5701}{0.6609}$ & $\intervalcc{56.412}{74.675}$ & $\intervalcc{9.286}{11.690}$ & $\intervalcc{4.511}{4.781}$ \\
 & $s=3\%$ & $\intervalcc{0.5403}{0.7027}$ & $\intervalcc{44.472}{77.355}$ & $\intervalcc{8.674}{12.574}$ & $\intervalcc{4.470}{5.066}$ \\
 & $s=4\%$ & $\intervalcc{0.4932}{0.7606}$ & $\intervalcc{26.954}{81.128}$ & $\intervalcc{8.005}{14.075}$ & $\intervalcc{4.434}{5.391}$ \\
 & $s=5\%$ & $\intervalcc{0.4370}{0.8264}$ & $\intervalcc{5.665}{88.325}$ & $\intervalcc{7.184}{15.980}$ & $\intervalcc{4.401}{5.597}$ \\ \hline
\end{tabular}
\end{table}

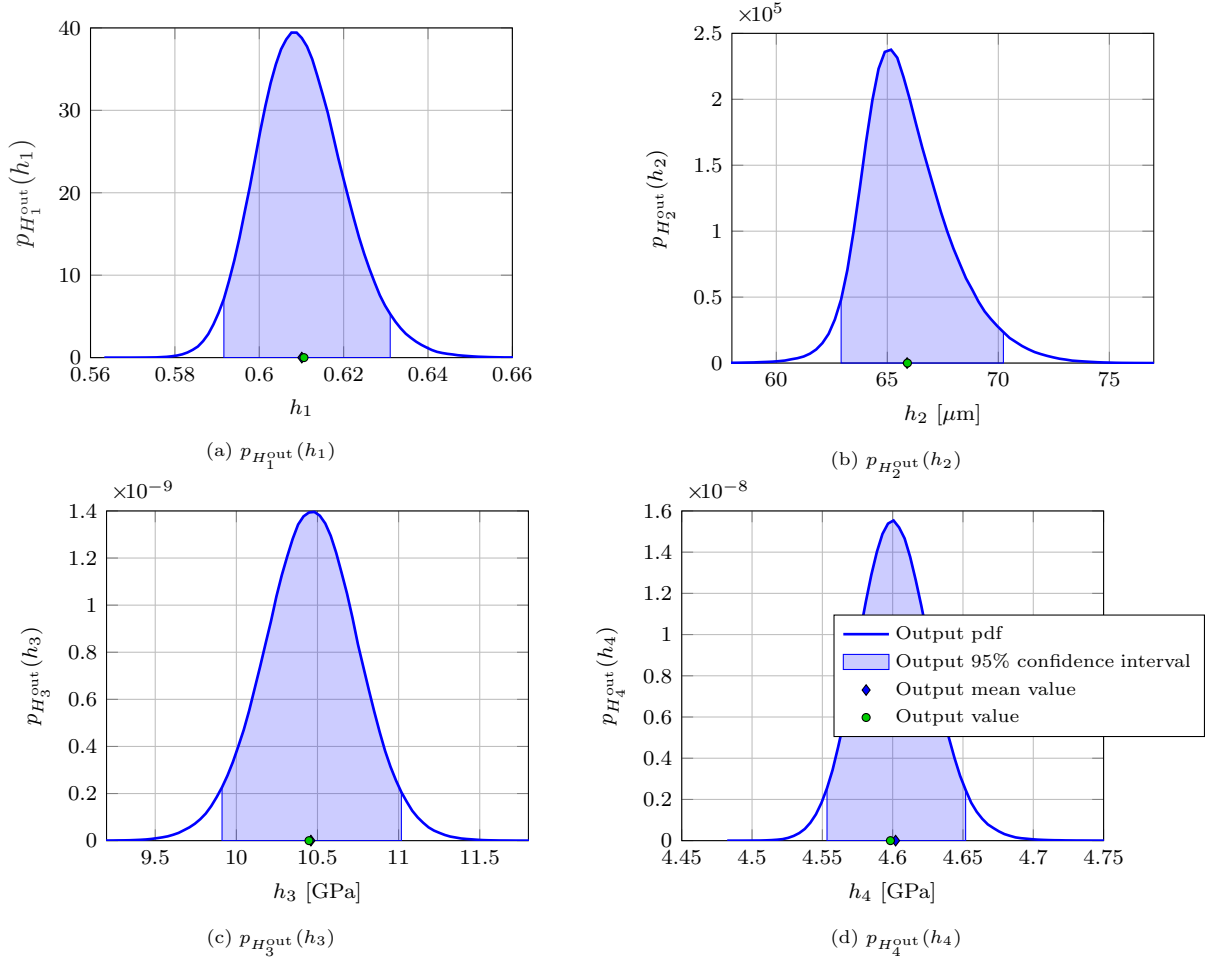
\begin{figure}[h!]
	\centering
	\begin{subfigure}{.5\linewidth}
		\centering
		\tikzsetnextfilename{pdf_posterior_delta_bone_1}
%
\begin{tikzpicture}

\begin{axis}[%
width=1.269\figureheight,
height=\figureheight,
at={(0\figureheight,0\figureheight)},
scale only axis,
xmin=0.56,
xmax=0.66,
xlabel={$h_1$},
ymin=0,
ymax=40,
ylabel style={font=\color{white!15!black}},
ylabel={$p_{H_1^{\text{out}}}(h_1)$},
xmajorgrids,
ymajorgrids,
legend style={legend cell align=left, align=left
}
]
\addplot [color=blue, line width=1.0pt]
  table[row sep=crcr]{%
0.563270594695931	6.52669378108612e-06\\
0.56450281188983	0.00029331970105629\\
0.565735029083729	0.000695846959407241\\
0.566967246277628	0.00129277006350195\\
0.568199463471527	0.000942432993169802\\
0.569431680665426	0.00113957231656726\\
0.570663897859325	0.00405939365374428\\
0.571896115053223	0.00789829979374458\\
0.573128332247122	0.0114117746442408\\
0.574360549441021	0.0198056907905442\\
0.57559276663492	0.0277063533437297\\
0.576824983828819	0.0614067034636852\\
0.578057201022718	0.0955479721469012\\
0.579289418216617	0.162813527085003\\
0.580521635410515	0.254891298791046\\
0.581753852604414	0.398743315996023\\
0.582986069798313	0.623594963587245\\
0.584218286992212	0.942538853097063\\
0.585450504186111	1.34695139941571\\
0.58668272138001	2.00204134157445\\
0.587914938573909	2.855938833072\\
0.589147155767808	4.03842507102958\\
0.590379372961707	5.45644188856263\\
0.591611590155605	7.12222797369638\\
0.592843807349504	9.3389687162368\\
0.594076024543403	11.7788753059398\\
0.595308241737302	14.5360457086156\\
0.596540458931201	17.6683360589812\\
0.5977726761251	20.7382959385868\\
0.599004893318999	24.1068181443127\\
0.600237110512897	27.4837468306134\\
0.601469327706796	30.5732900174908\\
0.602701544900695	33.3201905696435\\
0.603933762094594	35.581781374642\\
0.605165979288493	37.4101339112702\\
0.606398196482392	38.6076061695572\\
0.607630413676291	39.4348435941275\\
0.608862630870189	39.4298596216914\\
0.610094848064088	38.7699499617288\\
0.611327065257987	37.7415855367372\\
0.612559282451886	36.2026118146964\\
0.613791499645785	34.1377736855034\\
0.615023716839684	31.9517262212938\\
0.616255934033583	29.6784948387906\\
0.617488151227481	26.9628403130554\\
0.61872036842138	24.2237297534336\\
0.619952585615279	21.7195329842312\\
0.621184802809178	19.2753851358816\\
0.622417020003077	16.8784796219996\\
0.623649237196976	14.5996597224252\\
0.624881454390875	12.5848388539149\\
0.626113671584773	10.7732658196397\\
0.627345888778672	9.12276336086135\\
0.628578105972571	7.56580363104257\\
0.62981032316647	6.31872078483693\\
0.631042540360369	5.28636790796506\\
0.632274757554268	4.35828502671275\\
0.633506974748167	3.56666247351926\\
0.634739191942066	2.89283667935676\\
0.635971409135965	2.38756330605726\\
0.637203626329863	1.9270443863857\\
0.638435843523762	1.54593384203817\\
0.639668060717661	1.22428580223759\\
0.64090027791156	0.924821560522418\\
0.642132495105459	0.699012019052255\\
0.643364712299358	0.562410547049837\\
0.644596929493257	0.457765451306624\\
0.645829146687155	0.384426057918944\\
0.647061363881054	0.29069675389895\\
0.648293581074953	0.236125876324739\\
0.649525798268852	0.188977229596975\\
0.650758015462751	0.143651014011878\\
0.65199023265665	0.106699827968915\\
0.653222449850549	0.085089424182415\\
0.654454667044447	0.0625270943729015\\
0.655686884238346	0.0474234242407529\\
0.656919101432245	0.0346252519607399\\
0.658151318626144	0.0235037599909321\\
0.659383535820043	0.0236879280519614\\
0.660615753013942	0.0203835026952703\\
0.661847970207841	0.0113536041933526\\
0.663080187401739	0.0117575402698039\\
0.664312404595638	0.0090927976289167\\
0.665544621789537	0.00759219203804171\\
0.666776838983436	0.00398873782704288\\
0.668009056177335	0.00242523233343744\\
0.669241273371234	0.00189944892943217\\
0.670473490565133	0.0019756598652319\\
0.671705707759031	0.00165241471191838\\
0.67293792495293	0.000909842002594409\\
0.674170142146829	0.000488451850992769\\
0.675402359340728	0.000403427066424923\\
0.676634576534627	0.00039806844434897\\
0.677866793728526	0.00013571030407122\\
0.679099010922425	0.000619664817537926\\
0.680331228116324	0.00060003204727291\\
0.681563445310222	0.000308271195248884\\
0.682795662504121	0.000487895834212646\\
0.68402787969802	0.000290855267661371\\
0.685260096891919	6.52553204367371e-06\\
};

\addplot[area legend, draw=blue, fill=blue, fill opacity=0.2] table[row sep=crcr]{%
0.591611590155605	7.12222797369638\\
0.592843807349504	9.3389687162368\\
0.594076024543403	11.7788753059398\\
0.595308241737302	14.5360457086156\\
0.596540458931201	17.6683360589812\\
0.5977726761251	20.7382959385868\\
0.599004893318999	24.1068181443127\\
0.600237110512897	27.4837468306134\\
0.601469327706796	30.5732900174908\\
0.602701544900695	33.3201905696435\\
0.603933762094594	35.581781374642\\
0.605165979288493	37.4101339112702\\
0.606398196482392	38.6076061695572\\
0.607630413676291	39.4348435941275\\
0.608862630870189	39.4298596216914\\
0.610094848064088	38.7699499617288\\
0.611327065257987	37.7415855367372\\
0.612559282451886	36.2026118146964\\
0.613791499645785	34.1377736855034\\
0.615023716839684	31.9517262212938\\
0.616255934033583	29.6784948387906\\
0.617488151227481	26.9628403130554\\
0.61872036842138	24.2237297534336\\
0.619952585615279	21.7195329842312\\
0.621184802809178	19.2753851358816\\
0.622417020003077	16.8784796219996\\
0.623649237196976	14.5996597224252\\
0.624881454390875	12.5848388539149\\
0.626113671584773	10.7732658196397\\
0.627345888778672	9.12276336086135\\
0.628578105972571	7.56580363104257\\
0.62981032316647	6.31872078483693\\
0.631042540360369	5.28636790796506\\
}
\closedcycle;

\addplot[only marks, mark=diamond*, mark options={}, mark size=2.0000pt, draw=black, fill=blue] table[row sep=crcr]{%
x	y\\
0.610087410745185	0\\
};

\addplot[only marks, mark=*, mark options={}, mark size=1.5000pt, draw=black, fill=black!20!green] table[row sep=crcr]{%
x	y\\
0.610590072451957	0\\
};

\end{axis}
\end{tikzpicture}%
		\caption{$p_{H_1^{\text{out}}}(h_1)$}
		\label{fig:pdf_posterior_delta_bone_1}
	\end{subfigure}\hfill
	\begin{subfigure}{.5\linewidth}
		\centering
		\tikzsetnextfilename{pdf_posterior_lcorr_bone_1}
%
\begin{tikzpicture}

\begin{axis}[%
width=1.269\figureheight,
height=\figureheight,
at={(0\figureheight,0\figureheight)},
scale only axis,
xmin=58,
xmax=77,
xlabel={$h_2$ [$\mu$m]},
ymin=0,
ymax=250000,
ylabel={$p_{H_2^{\text{out}}}(h_2)$},
xmajorgrids,
ymajorgrids,
legend style={legend cell align=left, align=left
}
]
\addplot [color=blue, line width=1.0pt]
  table[row sep=crcr]{%
50.5509524321537	0.0373855866404604\\
50.8320561082641	2.76183636950122\\
51.1131597843746	0.737286871252125\\
51.394263460485	0.066236778179916\\
51.6753671365955	4.40024691432683\\
51.9564708127059	2.52242017783626\\
52.2375744888164	0.0105396663764038\\
52.5186781649268	3.71265560162883e-10\\
52.7997818410372	0.00021007754574051\\
53.0808855171477	0.42955289421274\\
53.3619891932581	3.17391804223696\\
53.6430928693686	0.0852508903626712\\
53.924196545479	0.658795732254542\\
54.2053002215895	6.85228374168195\\
54.4864038976999	5.94147476346168\\
54.7675075738103	13.9754247781673\\
55.0486112499208	22.6160066067675\\
55.3297149260312	16.9460251625386\\
55.6108186021417	11.2368390428634\\
55.8919222782521	24.5994877499433\\
56.1730259543626	35.6251617296123\\
56.454129630473	26.8252228486501\\
56.7352333065834	45.8789744781072\\
57.0163369826939	67.1408674738684\\
57.2974406588043	99.6342558327961\\
57.5785443349148	117.319824917956\\
57.8596480110252	162.08229637854\\
58.1407516871357	213.450145532756\\
58.4218553632461	288.427454209477\\
58.7029590393566	389.642372904742\\
58.984062715467	473.318215713184\\
59.2651663915774	682.340973762521\\
59.5462700676879	850.879454085824\\
59.8273737437983	1056.17326135758\\
60.1084774199088	1433.59263874425\\
60.3895810960192	1905.63885157267\\
60.6706847721297	2849.46950301193\\
60.9517884482401	3680.30230987004\\
61.2328921243505	4975.94855092661\\
61.513995800461	7111.28946019192\\
61.7950994765714	10289.2641648189\\
62.0762031526819	15125.7485244154\\
62.3573068287923	22154.7279552788\\
62.6384105049028	33054.5212543445\\
62.9195141810132	48238.6858098677\\
63.2006178571237	70591.9701281721\\
63.4817215332341	99956.7108332968\\
63.7628252093445	132868.406265251\\
64.043928885455	167835.530642117\\
64.3250325615654	199065.998047467\\
64.6061362376759	223166.495153847\\
64.8872399137863	235943.056917464\\
65.1683435898968	237617.746066485\\
65.4494472660072	231489.315645013\\
65.7305509421176	216830.440665452\\
66.0116546182281	199979.476909681\\
66.2927582943385	181396.85217108\\
66.573861970449	163416.705956896\\
66.8549656465594	146902.045810233\\
67.1360693226699	130520.333997238\\
67.4171729987803	114387.80826069\\
67.6982766748907	99843.4366139632\\
67.9793803510012	86838.646831789\\
68.2604840271116	75499.3367397682\\
68.5415877032221	64315.5803358786\\
68.8226913793325	55730.6490887312\\
69.103795055443	46648.9758692331\\
69.3848987315534	39131.22528295\\
69.6660024076639	33348.9611108549\\
69.9471060837743	28163.7870131765\\
70.2282097598847	23502.6293436043\\
70.5093134359952	19013.6319745096\\
70.7904171121056	15955.6799390527\\
71.0715207882161	12910.6698845167\\
71.3526244643265	10478.5962882918\\
71.633728140437	8336.44406704174\\
71.9148318165474	6802.07455591877\\
72.1959354926578	5456.17316254541\\
72.4770391687683	4303.04120208727\\
72.7581428448787	3283.91123646579\\
73.0392465209892	2623.97721573939\\
73.3203501970996	1853.25657223518\\
73.60145387321	1502.49421474758\\
73.8825575493205	1188.63559345414\\
74.1636612254309	903.625178739686\\
74.4447649015414	646.926553972096\\
74.7258685776518	456.062095994551\\
75.0069722537623	348.345172828545\\
75.2880759298727	205.527957549744\\
75.5691796059832	128.836326130126\\
75.8502832820936	94.8215421579219\\
76.131386958204	57.7882399976355\\
76.4124906343145	38.7964466403861\\
76.6935943104249	14.6952479268181\\
76.9746979865354	5.87068652055546\\
77.2558016626458	13.6309101727657\\
77.5369053387563	10.1729701190959\\
77.8180090148667	1.09160032514493\\
78.0991126909772	2.76183636950005\\
78.3802163670876	0.0373855866404668\\
};

\addplot[area legend, draw=blue, fill=blue, fill opacity=0.2] table[row sep=crcr]{%
62.9195141810132	48238.6858098677\\
63.2006178571237	70591.9701281721\\
63.4817215332341	99956.7108332968\\
63.7628252093445	132868.406265251\\
64.043928885455	167835.530642117\\
64.3250325615654	199065.998047467\\
64.6061362376759	223166.495153847\\
64.8872399137863	235943.056917464\\
65.1683435898968	237617.746066485\\
65.4494472660072	231489.315645013\\
65.7305509421176	216830.440665452\\
66.0116546182281	199979.476909681\\
66.2927582943385	181396.85217108\\
66.573861970449	163416.705956896\\
66.8549656465594	146902.045810233\\
67.1360693226699	130520.333997238\\
67.4171729987803	114387.80826069\\
67.6982766748907	99843.4366139632\\
67.9793803510012	86838.646831789\\
68.2604840271116	75499.3367397682\\
68.5415877032221	64315.5803358786\\
68.8226913793325	55730.6490887312\\
69.103795055443	46648.9758692331\\
69.3848987315534	39131.22528295\\
69.6660024076639	33348.9611108549\\
69.9471060837743	28163.7870131765\\
70.2282097598847	23502.6293436043\\
}
\closedcycle;

\addplot[only marks, mark=diamond*, mark options={}, mark size=2.0000pt, draw=black, fill=blue] table[row sep=crcr]{%
x	y\\
65.8907628708782	0\\
};

\addplot[only marks, mark=*, mark options={}, mark size=1.5000pt, draw=black, fill=black!20!green] table[row sep=crcr]{%
x	y\\
65.9062206547429	0\\
};

\end{axis}
\end{tikzpicture}%
		\caption{$p_{H_2^{\text{out}}}(h_2)$}
		\label{fig:pdf_posterior_lcorr_bone_1}
	\end{subfigure}\\
	\begin{subfigure}{.5\linewidth}
		\centering
		\tikzsetnextfilename{pdf_posterior_kappa_bone_1}
%
\begin{tikzpicture}

\begin{axis}[%
width=1.269\figureheight,
height=\figureheight,
at={(0\figureheight,0\figureheight)},
scale only axis,
xmin=9.2,
xmax=11.8,
xlabel={$h_3$ [GPa]},
ymin=0,
ymax=1.4e-09,
ylabel={$p_{H_3^{\text{out}}}(h_3)$},
xmajorgrids,
ymajorgrids,
legend style={legend cell align=left, align=left
}
]
\addplot [color=blue, line width=1.0pt]
  table[row sep=crcr]{%
8.70588167378004	2.33169131053852e-16\\
8.73937765551695	9.75766629816029e-15\\
8.77287363725385	1.8291023816396e-14\\
8.80636961899076	1.54682088507121e-15\\
8.83986560072767	3.03417926751106e-15\\
8.87336158246458	4.32134890700949e-14\\
8.90685756420148	5.57037463011821e-14\\
8.94035354593839	5.43354379774517e-14\\
8.9738495276753	2.33098448428857e-14\\
9.00734550941221	4.09990248630266e-14\\
9.04084149114911	9.20153312503189e-14\\
9.07433747288602	7.96102678730175e-14\\
9.10783345462293	2.25947547044526e-13\\
9.14132943635984	2.78163306146002e-13\\
9.17482541809674	2.88055190091965e-13\\
9.20832139983365	4.80332090218093e-13\\
9.24181738157056	7.00616287080978e-13\\
9.27531336330746	1.04146175331536e-12\\
9.30880934504437	1.24490063681781e-12\\
9.34230532678128	1.69323637039662e-12\\
9.37580130851819	2.59812676643746e-12\\
9.4092972902551	3.62594074756854e-12\\
9.442793271992	5.02111553469654e-12\\
9.47628925372891	7.2648787835698e-12\\
9.50978523546582	1.04605047671706e-11\\
9.54328121720272	1.39399782799893e-11\\
9.57677719893963	1.82649527526407e-11\\
9.61027318067654	2.48949425407352e-11\\
9.64376916241345	3.28394688231216e-11\\
9.67726514415035	4.45371083588837e-11\\
9.71076112588726	5.64660661283076e-11\\
9.74425710762417	7.17680771234199e-11\\
9.77775308936108	9.11869791665023e-11\\
9.81124907109798	1.14931406632581e-10\\
9.84474505283489	1.46352479126718e-10\\
9.8782410345718	1.8441220445753e-10\\
9.9117370163087	2.26721841457854e-10\\
9.94523299804561	2.76405139809118e-10\\
9.97872897978252	3.34121470191182e-10\\
10.0122249615194	4.00822106203752e-10\\
10.0457209432563	4.70776008865961e-10\\
10.0792169249932	5.52887623321404e-10\\
10.1127129067301	6.45613063570632e-10\\
10.1462088884671	7.38334755074607e-10\\
10.179704870204	8.35694255867548e-10\\
10.2132008519409	9.31083730126596e-10\\
10.2466968336778	1.03648821913508e-09\\
10.2801928154147	1.12762362113572e-09\\
10.3136887971516	1.20715959751807e-09\\
10.3471847788885	1.28144872878361e-09\\
10.3806807606254	1.34242860328696e-09\\
10.4141767423623	1.37716936945959e-09\\
10.4476727240992	1.39346095510939e-09\\
10.4811687058361	1.39580793422595e-09\\
10.514664687573	1.37975317294143e-09\\
10.5481606693099	1.34684470240041e-09\\
10.5816566510469	1.29449043729498e-09\\
10.6151526327838	1.22261902332323e-09\\
10.6486486145207	1.13722972972341e-09\\
10.6821445962576	1.04880541131746e-09\\
10.7156405779945	9.4680281914435e-10\\
10.7491365597314	8.34302887746811e-10\\
10.7826325414683	7.37503903811202e-10\\
10.8161285232052	6.41400718508254e-10\\
10.8496245049421	5.4868255767665e-10\\
10.883120486679	4.61711822105393e-10\\
10.9166164684159	3.81449011569781e-10\\
10.9501124501528	3.13746925087998e-10\\
10.9836084318897	2.561605998158e-10\\
11.0171044136266	2.05666333687304e-10\\
11.0506003953636	1.64742226035985e-10\\
11.0840963771005	1.26914029090294e-10\\
11.1175923588374	9.87965911401159e-11\\
11.1510883405743	7.62167992387886e-11\\
11.1845843223112	5.64720544338249e-11\\
11.2180803040481	4.27419540723927e-11\\
11.251576285785	3.30349845424206e-11\\
11.2850722675219	2.40446952800632e-11\\
11.3185682492588	1.7306743010184e-11\\
11.3520642309957	1.32787338814576e-11\\
11.3855602127326	9.17573584536944e-12\\
11.4190561944695	6.9978483893895e-12\\
11.4525521762064	5.3687210939765e-12\\
11.4860481579434	3.68245455799985e-12\\
11.5195441396803	2.33518682900801e-12\\
11.5530401214172	1.55931448285234e-12\\
11.5865361031541	1.12471136526634e-12\\
11.620032084891	6.83244267851418e-13\\
11.6535280666279	5.88102761719421e-13\\
11.6870240483648	4.46304723053809e-13\\
11.7205200301017	2.31014297662139e-13\\
11.7540160118386	1.71064273075799e-13\\
11.7875119935755	1.19291321635441e-13\\
11.8210079753124	9.54017868167117e-14\\
11.8545039570493	8.21582788273576e-14\\
11.8879999387862	5.92144653707816e-14\\
11.9214959205231	5.46765870490969e-14\\
11.9549919022601	3.2155164682267e-14\\
11.988487883997	1.00987646560892e-14\\
12.0219838657339	2.33169131053852e-16\\
};

\addplot[area legend, draw=blue, fill=blue, fill opacity=0.2] table[row sep=crcr]{%
9.9117370163087	2.26721841457854e-10\\
9.94523299804561	2.76405139809118e-10\\
9.97872897978252	3.34121470191182e-10\\
10.0122249615194	4.00822106203752e-10\\
10.0457209432563	4.70776008865961e-10\\
10.0792169249932	5.52887623321404e-10\\
10.1127129067301	6.45613063570632e-10\\
10.1462088884671	7.38334755074607e-10\\
10.179704870204	8.35694255867548e-10\\
10.2132008519409	9.31083730126596e-10\\
10.2466968336778	1.03648821913508e-09\\
10.2801928154147	1.12762362113572e-09\\
10.3136887971516	1.20715959751807e-09\\
10.3471847788885	1.28144872878361e-09\\
10.3806807606254	1.34242860328696e-09\\
10.4141767423623	1.37716936945959e-09\\
10.4476727240992	1.39346095510939e-09\\
10.4811687058361	1.39580793422595e-09\\
10.514664687573	1.37975317294143e-09\\
10.5481606693099	1.34684470240041e-09\\
10.5816566510469	1.29449043729498e-09\\
10.6151526327838	1.22261902332323e-09\\
10.6486486145207	1.13722972972341e-09\\
10.6821445962576	1.04880541131746e-09\\
10.7156405779945	9.4680281914435e-10\\
10.7491365597314	8.34302887746811e-10\\
10.7826325414683	7.37503903811202e-10\\
10.8161285232052	6.41400718508254e-10\\
10.8496245049421	5.4868255767665e-10\\
10.883120486679	4.61711822105393e-10\\
10.9166164684159	3.81449011569781e-10\\
10.9501124501528	3.13746925087998e-10\\
10.9836084318897	2.561605998158e-10\\
11.0171044136266	2.05666333687304e-10\\
}
\closedcycle;

\addplot[only marks, mark=diamond*, mark options={}, mark size=2.0000pt, draw=black, fill=blue] table[row sep=crcr]{%
x	y\\
10.459556850696	0\\
};

\addplot[only marks, mark=*, mark options={}, mark size=1.5000pt, draw=black, fill=black!20!green] table[row sep=crcr]{%
x	y\\
10.4479618095717	0\\
};

\end{axis}
\end{tikzpicture}%
		\caption{$p_{H_3^{\text{out}}}(h_3)$}
		\label{fig:pdf_posterior_kappa_bone_1}
	\end{subfigure}\hfill
	\begin{subfigure}{.5\linewidth}
		\centering
		\tikzsetnextfilename{pdf_posterior_mu_bone_1}
%
\begin{tikzpicture}

\begin{axis}[%
width=1.269\figureheight,
height=\figureheight,
at={(0\figureheight,0\figureheight)},
scale only axis,
xmin=4.45,
xmax=4.75,
xlabel={$h_4$ [GPa]},
ymin=0,
ymax=1.6e-08,
ylabel={$p_{H_4^{\text{out}}}(h_4)$},
xmajorgrids,
ymajorgrids,
legend style={legend cell align=left, align=left,
at={(0.8,0.5)},anchor=center
}
]
\addplot [color=blue, line width=1.0pt]
  table[row sep=crcr]{%
4.48233140185252	2.76109097904061e-15\\
4.48627459965573	2.67792454233698e-13\\
4.49021779745893	5.1307054204967e-13\\
4.49416099526214	1.3551635949686e-12\\
4.49810419306534	2.50735577650188e-12\\
4.50204739086855	4.77693024813844e-12\\
4.50599058867175	6.19803079499254e-12\\
4.50993378647496	1.42918150324579e-11\\
4.51387698427817	2.59668010722232e-11\\
4.51782018208137	4.52608437613734e-11\\
4.52176337988458	7.36511281696406e-11\\
4.52570657768778	1.34912420659668e-10\\
4.52964977549099	2.30907895130395e-10\\
4.53359297329419	3.68221190443828e-10\\
4.5375361710974	5.68269274395178e-10\\
4.5414793689006	8.70930599482873e-10\\
4.54542256670381	1.275991133946e-09\\
4.54936576450702	1.83395884992916e-09\\
4.55330896231022	2.54554960367834e-09\\
4.55725216011343	3.42459399716577e-09\\
4.56119535791663	4.53683470252487e-09\\
4.56513855571984	5.73528736858391e-09\\
4.56908175352304	7.1194991434864e-09\\
4.57302495132625	8.61057073406122e-09\\
4.57696814912945	1.01588014211775e-08\\
4.58091134693266	1.15914075342461e-08\\
4.58485454473587	1.29059890597538e-08\\
4.58879774253907	1.40194282305279e-08\\
4.59274094034228	1.48829388017046e-08\\
4.59668413814548	1.53859240547797e-08\\
4.60062733594869	1.55376493514742e-08\\
4.60457053375189	1.51986086298151e-08\\
4.6085137315551	1.46644716887828e-08\\
4.6124569293583	1.37670703647801e-08\\
4.61640012716151	1.26439794128591e-08\\
4.62034332496472	1.14115215561233e-08\\
4.62428652276792	1.00633490846079e-08\\
4.62822972057113	8.67169535137765e-09\\
4.63217291837433	7.28370098996141e-09\\
4.63611611617754	6.03512006855753e-09\\
4.64005931398074	4.97258098040052e-09\\
4.64400251178395	4.00566599698287e-09\\
4.64794570958715	3.14769421891604e-09\\
4.65188890739036	2.46280785794188e-09\\
4.65583210519357	1.84335009150664e-09\\
4.65977530299677	1.39948674864975e-09\\
4.66371850079998	1.05967735250652e-09\\
4.66766169860318	7.88403679798676e-10\\
4.67160489640639	6.12770948005583e-10\\
4.67554809420959	4.4940467557482e-10\\
4.6794912920128	3.17710847933279e-10\\
4.683434489816	2.33148185227202e-10\\
4.68737768761921	1.57303441142707e-10\\
4.69132088542242	1.18028184201761e-10\\
4.69526408322562	9.36990136059542e-11\\
4.69920728102883	6.75424981836743e-11\\
4.70315047883203	4.85424196201313e-11\\
4.70709367663524	3.64048685045357e-11\\
4.71103687443844	2.69595048887604e-11\\
4.71498007224165	2.01066205102013e-11\\
4.71892327004485	1.7507230056545e-11\\
4.72286646784806	1.16720822413737e-11\\
4.72680966565127	9.32540093518866e-12\\
4.73075286345447	6.56401276566473e-12\\
4.73469606125768	6.33112470994531e-12\\
4.73863925906088	5.056211244479e-12\\
4.74258245686409	3.70973438914457e-12\\
4.74652565466729	3.34770176723538e-12\\
4.7504688524705	2.54146002132815e-12\\
4.7544120502737	1.31453780700178e-12\\
4.75835524807691	1.54692251489293e-12\\
4.76229844588012	1.28898576328586e-12\\
4.76624164368332	2.42226629126411e-12\\
4.77018484148653	1.27818251344809e-12\\
4.77412803928973	4.33013106678397e-13\\
4.77807123709294	2.90209627984285e-13\\
4.78201443489614	5.05783672265668e-13\\
4.78595763269935	2.94325904948708e-13\\
4.78990083050255	1.50850772873438e-13\\
4.79384402830576	1.28873150000085e-15\\
4.79778722610897	3.69885112268428e-20\\
4.80173042391217	9.96135500996905e-16\\
4.80567362171538	1.58998417822365e-13\\
4.80961681951858	3.83835011615898e-13\\
4.81356001732179	2.14617124935337e-13\\
4.81750321512499	1.39229203769035e-13\\
4.8214464129282	3.5966908400741e-13\\
4.82538961073141	6.48400983556425e-14\\
4.82933280853461	4.25133943291877e-13\\
4.83327600633782	3.93161297721785e-14\\
4.83721920414102	1.87351648585908e-13\\
4.84116240194423	6.05453610606388e-14\\
4.84510559974743	1.03042803260638e-16\\
4.84904879755064	1.58460370810064e-29\\
4.85299199535384	4.3998622525174e-22\\
4.85693519315705	6.4338499723386e-17\\
4.86087839096026	4.96757704903151e-14\\
4.86482158876346	2.67350993325542e-13\\
4.86876478656667	1.84462528596757e-13\\
4.87270798436987	2.57434894759866e-15\\
};
\addlegendentry{Output pdf}

\addplot[area legend, draw=blue, fill=blue, fill opacity=0.2] table[row sep=crcr]{%
4.55330896231022	2.54554960367834e-09\\
4.55725216011343	3.42459399716577e-09\\
4.56119535791663	4.53683470252487e-09\\
4.56513855571984	5.73528736858391e-09\\
4.56908175352304	7.1194991434864e-09\\
4.57302495132625	8.61057073406122e-09\\
4.57696814912945	1.01588014211775e-08\\
4.58091134693266	1.15914075342461e-08\\
4.58485454473587	1.29059890597538e-08\\
4.58879774253907	1.40194282305279e-08\\
4.59274094034228	1.48829388017046e-08\\
4.59668413814548	1.53859240547797e-08\\
4.60062733594869	1.55376493514742e-08\\
4.60457053375189	1.51986086298151e-08\\
4.6085137315551	1.46644716887828e-08\\
4.6124569293583	1.37670703647801e-08\\
4.61640012716151	1.26439794128591e-08\\
4.62034332496472	1.14115215561233e-08\\
4.62428652276792	1.00633490846079e-08\\
4.62822972057113	8.67169535137765e-09\\
4.63217291837433	7.28370098996141e-09\\
4.63611611617754	6.03512006855753e-09\\
4.64005931398074	4.97258098040052e-09\\
4.64400251178395	4.00566599698287e-09\\
4.64794570958715	3.14769421891604e-09\\
4.65188890739036	2.46280785794188e-09\\
}
\closedcycle;
\addlegendentry{Output $95\%$ confidence interval}

\addplot[only marks, mark=diamond*, mark options={}, mark size=2.0000pt, draw=black, fill=blue] table[row sep=crcr]{%
x	y\\
4.60209566557382	0\\
};
\addlegendentry{Output mean value}

\addplot[only marks, mark=*, mark options={}, mark size=1.5000pt, draw=black, fill=black!20!green] table[row sep=crcr]{%
x	y\\
4.59840650118907	0\\
};
\addlegendentry{Output value}

\end{axis}
\end{tikzpicture}%
		\caption{$p_{H_4^{\text{out}}}(h_4)$}
		\label{fig:pdf_posterior_mu_bone_1}
	\end{subfigure}
	\caption{Real experimental data: probability density functions $p_{H_1^{\text{out}}}$, $p_{H_2^{\text{out}}}$, $p_{H_3^{\text{out}}}$ and $p_{H_4^{\text{out}}}$ of random variables $H_1^{\text{out}}$, $H_2^{\text{out}}$, $H_3^{\text{out}}$ and $H_4^{\text{out}}$, respectively, obtained for a given input vector $\qb^{\text{obs}}$ and for a given input uncertainty level $s = 0.01 = 1\%$, with the output $95\%$ confidence intervals $I^{\text{out}}$ (blue areas), the output mean values $\underline{\hb}^{\text{out}}$ (blue diamonds) and the output values $\hb^{\text{out}}$ (green circles)}
	\label{fig:pdf_posterior_bone}
\end{figure}

\begin{figure}[h!]
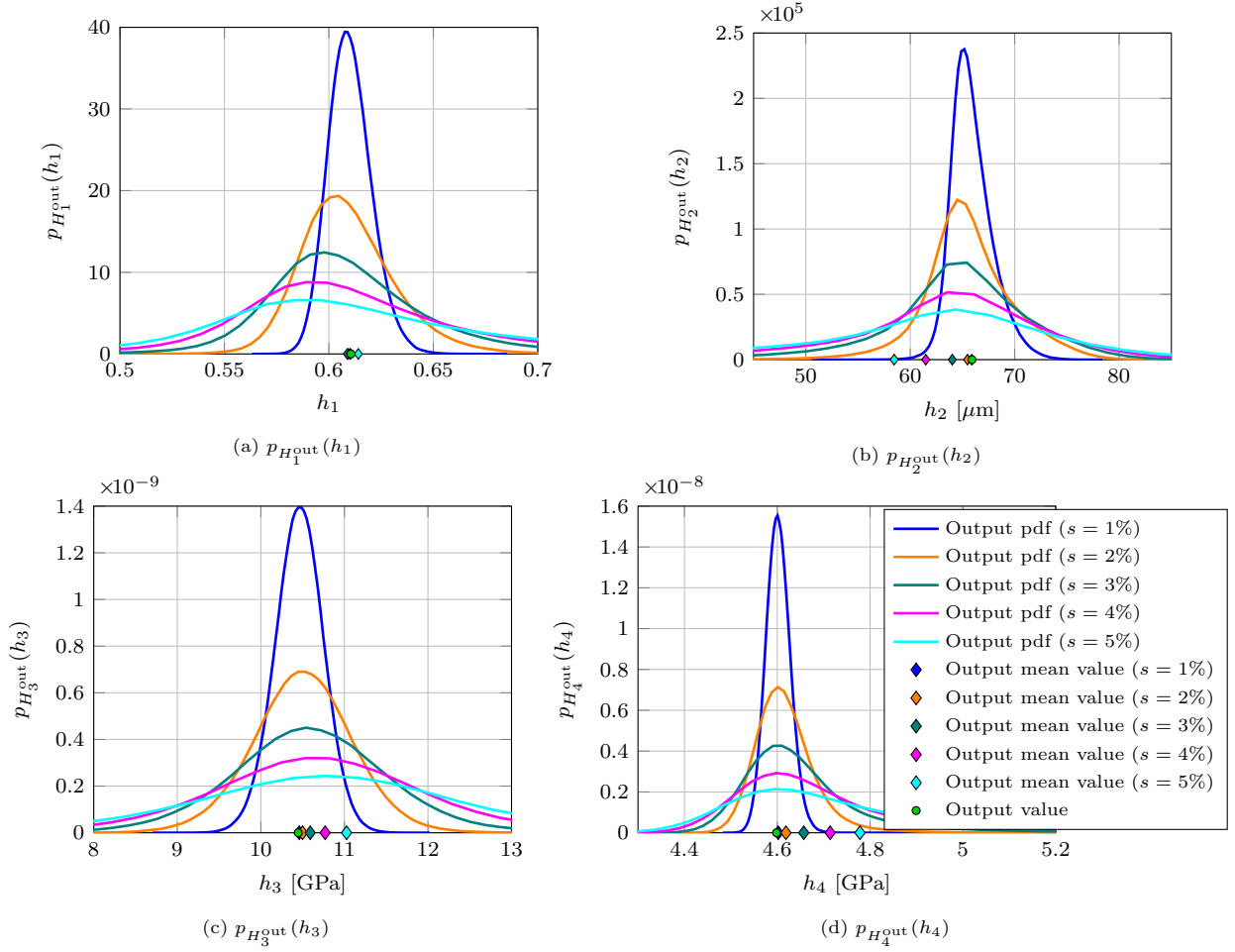

	\centering
	\begin{subfigure}{.5\linewidth}
		\centering
		\tikzsetnextfilename{pdfs_posterior_delta_bone}
		\input{pdfs_posterior_delta_bone}
		\caption{$p_{H_1^{\text{out}}}(h_1)$}
		\label{fig:pdfs_posterior_delta_bone}
	\end{subfigure}\hfill
	\begin{subfigure}{.5\linewidth}
		\centering
		\tikzsetnextfilename{pdfs_posterior_lcorr_bone}
		\input{pdfs_posterior_lcorr_bone}
		\caption{$p_{H_2^{\text{out}}}(h_2)$}
		\label{fig:pdfs_posterior_lcorr_bone}
	\end{subfigure}\\
	\begin{subfigure}{.45\linewidth}
		\centering
		\tikzsetnextfilename{pdfs_posterior_kappa_bone}
		\input{pdfs_posterior_kappa_bone}
		\caption{$p_{H_3^{\text{out}}}(h_3)$}
		\label{fig:pdfs_posterior_kappa_bone}
	\end{subfigure}\hfill
	\begin{subfigure}{.55\linewidth}
		\centering
		\tikzsetnextfilename{pdfs_posterior_mu_bone}
		\input{pdfs_posterior_mu_bone}
		\caption{$p_{H_4^{\text{out}}}(h_4)$}
		\label{fig:pdfs_posterior_mu_bone}
	\end{subfigure}
	\caption{Real experimental data: probability density functions $p_{H_1^{\text{out}}}$, $p_{H_2^{\text{out}}}$, $p_{H_3^{\text{out}}}$ and $p_{H_4^{\text{out}}}$ of random variables $H_1^{\text{out}}$, $H_2^{\text{out}}$, $H_3^{\text{out}}$ and $H_4^{\text{out}}$, respectively, obtained for a given input vector $\qb^{\text{obs}}$ and for different values of input uncertainty level $s \in \set{0.01,0.02,0.03,0.04,0.05}$, with the output mean values $\underline{\hb}^{\text{out}}$ (colored diamonds) and the output values $\hb^{\text{out}}$ (green circles)}
	\label{fig:pdfs_posterior_bone}
\end{figure}

The marginal probability density functions $p_{H_1^{\text{out}}}$, $p_{H_2^{\text{out}}}$, $p_{H_3^{\text{out}}}$ and $p_{H_4^{\text{out}}}$ of random variables $H_1^{\text{out}}$, $H_2^{\text{out}}$, $H_3^{\text{out}}$ and $H_4^{\text{out}}$, respectively, estimated by using the kernel density estimation method with the $N_s = 10^6$ independent realizations $\hb^{\text{out},(1)},\dots,\hb^{\text{out},(N_s)}$ of $\Hb^{\text{out}} = (H_1^{\text{out}},H_2^{\text{out}},H_3^{\text{out}},H_4^{\text{out}})$ are represented in Figure~\ref{fig:pdf_posterior_bone} for a given input uncertainty level $s = 0.01 = 1\%$ and in Figure~\ref{fig:pdfs_posterior_bone} for an input uncertainty level $s$ varying from $0.01=1\%$ to $0.05=5\%$. We observe similar trends as for the previous validation example. Despite the large scattering of the network outputs for the highest input uncertainties level, the output mean values $\underline{h}_1^{\text{out}}$, $\underline{h}_2^{\text{out}}$, $\underline{h}_3^{\text{out}}$ and $\underline{h}_4^{\text{out}}$ are still relatively close to the corresponding output values $h_1^{\text{out}}$, $h_2^{\text{out}}$, $h_3^{\text{out}}$ and $h_4^{\text{out}}$ (obtained without considering input uncertainties), thus showing the capability of the neural network-based identification method to efficiently computing robust output predictions with respect to the input uncertainties. Finally, such a real-world application has demonstrated the potential of the proposed identification method for solving the challenging statistical inverse problem related to the statistical identification of a stochastic model of the random compliance field (in high stochastic dimension) at mesoscale for a heterogeneous anisotropic elastic microstructure, by making use of a trained artificial neural network.

\section{Conclusions}
\label{sec:conclusion}

In this paper, a neural network-based identification method has been presented for solving the statistical inverse problem related to the statistical identification of the hyperparameters of a prior stochastic model of the random compliance elasticity field characterizing the apparent elastic properties of heterogeneous materials with complex random microstructure. Such a challenging statistical inverse problem has been formulated as a function approximation problem and solved by using an artificial neural network trained from a large numerical database. A first (initial) database has been generated using forward numerical simulations of mechanical computational models introduced within the framework of linear elasticity theory under 2D plane stress assumption. A second (processed) database has been derived by conditioning the input data contained in the initial database with respect to the target data. A sensitivity analysis of the target data with respect to the input data contained in each of the two databases has been performed. Two- and three-layer feedforward neural networks have been trained with each of the initial and processed databases and optimized by considering different network configurations in order to construct fine-tuned trained models. Numerical results show that the neural networks trained with the processed database exhibit much better performances in terms of mean squared error, linear regression analysis and probability distribution between network outputs and targets than the ones trained with the initial database. The conditioning of the initial database turns out to be an essential step in obtaining an efficient trained neural network for solving the underlying statistical inverse problem. An \emph{ad hoc} probabilistic model of the input random vector has been finally proposed in order to take into account experimental errors on the network input and to perform a robustness analysis of the network output with respect to the input uncertainties level. The challenging problem related to the identification of the input uncertainties level would deserve an in-depth analysis and should be part of a forthcoming work. The proposed neural-network based identification method has been successfully applied to synthetic data and then carried out on real experimental data coming from experimental measurements on a beef cortical bone specimen. Although the proposed method has been developed for a simple 2D plane stress linear elasticity problem, it could be easily extended to more complicated 3D physical problems encountered in computational mechanics and engineering sciences. Finally, instead of using classical feedforward static (or series) neural networks, other neural network architectures may be considered to increase the network training speed and improve the neural network performance, such as multilayer cascade-forward neural networks (that may include additional feedforward connections), dynamic (or recurrent) neural networks (with feedback (or recurrent) connections and/or tapped delay lines) and directed acyclic graph (DAG) neural networks (that may include skipped layers or layers connected and operating in parallel), thus allowing for various neural network configurations and topologies to learn either static or dynamic (time-dependent) series relationships depending on the problem to be solved.

\section*{Acknowlegements}

The authors gratefully acknowledge Christian Soize, Professor at Universit\'e Gustave Eiffel, Laboratoire MSME, for helpful discussions and valuable suggestions.

\bibliographystyle{elsarticle-num}
\bibliography{Biblio}

\begin{thebibliography}{10}
\expandafter\ifx\csname url\endcsname\relax
  \def\url#1{\texttt{#1}}\fi
\expandafter\ifx\csname urlprefix\endcsname\relax\def\urlprefix{URL }\fi
\expandafter\ifx\csname href\endcsname\relax
  \def\href#1#2{#2} \def\path#1{#1}\fi

\bibitem{Soi06}
C.~Soize, \href{https://doi.org/10.1016/j.cma.2004.12.014}{{Non-Gaussian
  positive-definite matrix-valued random fields for elliptic stochastic partial
  differential operators}}, Computer Methods in Applied Mechanics and
  Engineering 195~(1--3) (2006) 26--64.
\newblock \href {https://doi.org/10.1016/j.cma.2004.12.014}
  {\path{doi:10.1016/j.cma.2004.12.014}}.
\newline\urlprefix\url{https://doi.org/10.1016/j.cma.2004.12.014}

\bibitem{Soi08a}
C.~Soize,
  \href{https://doi.org/10.1016/j.probengmech.2007.12.019}{{Tensor-valued
  random fields for meso-scale stochastic model of anisotropic elastic
  microstructure and probabilistic analysis of representative volume element
  size}}, Probabilistic Engineering Mechanics 23~(2--3) (2008) 307--323, 5th
  International Conference on Computational Stochastic Mechanics.
\newblock \href {https://doi.org/10.1016/j.probengmech.2007.12.019}
  {\path{doi:10.1016/j.probengmech.2007.12.019}}.
\newline\urlprefix\url{https://doi.org/10.1016/j.probengmech.2007.12.019}

\bibitem{Ngu15}
M.-T. Nguyen, C.~Desceliers, C.~Soize, J.-M. Allain, H.~Gharbi,
  \href{https://doi.org/10.1615/IntJMultCompEng.2015011435}{{Multiscale
  identification of the random elasticity field at mesoscale of a heterogeneous
  microstructure using multiscale experimental observations}}, International
  Journal for Multiscale Computational Engineering 13~(4) (2015) 281--295.
\newblock \href {https://doi.org/10.1615/IntJMultCompEng.2015011435}
  {\path{doi:10.1615/IntJMultCompEng.2015011435}}.
\newline\urlprefix\url{https://doi.org/10.1615/IntJMultCompEng.2015011435}

\bibitem{Zha20}
T.~Zhang, F.~Pled, C.~Desceliers,
  \href{https://www.mdpi.com/1996-1944/13/12/2826}{{Robust Multiscale
  Identification of Apparent Elastic Properties at Mesoscale for Random
  Heterogeneous Materials with Multiscale Field Measurements}}, Materials
  13~(12) (2020).
\newblock \href {https://doi.org/10.3390/ma13122826}
  {\path{doi:10.3390/ma13122826}}.
\newline\urlprefix\url{https://www.mdpi.com/1996-1944/13/12/2826}

\bibitem{Des06}
C.~Desceliers, R.~Ghanem, C.~Soize,
  \href{https://doi.org/10.1002/nme.1576}{{Maximum likelihood estimation of
  stochastic chaos representations from experimental data}}, International
  Journal for Numerical Methods in Engineering 66~(6) (2006) 978--1001.
\newblock \href {https://doi.org/10.1002/nme.1576}
  {\path{doi:10.1002/nme.1576}}.
\newline\urlprefix\url{https://doi.org/10.1002/nme.1576}

\bibitem{Gha06}
R.~G. Ghanem, A.~Doostan, \href{https://doi.org/10.1016/j.jcp.2006.01.037}{{On
  the construction and analysis of stochastic models: Characterization and
  propagation of the errors associated with limited data}}, Journal of
  Computational Physics 217~(1) (2006) 63--81.
\newblock \href {https://doi.org/10.1016/j.jcp.2006.01.037}
  {\path{doi:10.1016/j.jcp.2006.01.037}}.
\newline\urlprefix\url{https://doi.org/10.1016/j.jcp.2006.01.037}

\bibitem{Des07}
C.~Desceliers, C.~Soize, R.~Ghanem,
  \href{https://doi.org/10.1007/s00466-006-0072-7}{{Identification of Chaos
  Representations of Elastic Properties of Random Media Using Experimental
  Vibration Tests}}, Computational Mechanics 39~(6) (2007) 831--838.
\newblock \href {https://doi.org/10.1007/s00466-006-0072-7}
  {\path{doi:10.1007/s00466-006-0072-7}}.
\newline\urlprefix\url{https://doi.org/10.1007/s00466-006-0072-7}

\bibitem{Mar07}
{Youssef M. Marzouk and Habib N. Najm and Larry A. Rahn},
  \href{https://doi.org/10.1016/j.jcp.2006.10.010}{Stochastic spectral methods
  for efficient bayesian solution of inverse problems}, Journal of
  Computational Physics 224~(2) (2007) 560--586.
\newblock \href {https://doi.org/10.1016/j.jcp.2006.10.010}
  {\path{doi:10.1016/j.jcp.2006.10.010}}.
\newline\urlprefix\url{https://doi.org/10.1016/j.jcp.2006.10.010}

\bibitem{Arn08}
M.~Arnst, D.~Clouteau, M.~Bonnet,
  \href{https://doi.org/10.1016/j.cma.2007.08.011}{{Inversion of probabilistic
  structural models using measured transfer functions}}, Computer Methods in
  Applied Mechanics and Engineering 197~(6) (2008) 589--608.
\newblock \href {https://doi.org/10.1016/j.cma.2007.08.011}
  {\path{doi:10.1016/j.cma.2007.08.011}}.
\newline\urlprefix\url{https://doi.org/10.1016/j.cma.2007.08.011}

\bibitem{Das08}
S.~Das, R.~Ghanem, J.~C. Spall, {Asymptotic Sampling Distribution for
  Polynomial Chaos Representation of Data: A Maximum Entropy and Fisher
  information approach}, in: Proceedings of the 45th IEEE Conference on
  Decision and Control, 2006, pp. 4139--4144.
\newblock \href {https://doi.org/10.1109/CDC.2006.377613}
  {\path{doi:10.1109/CDC.2006.377613}}.

\bibitem{Das09b}
S.~Das, R.~Ghanem, S.~Finette,
  \href{https://doi.org/10.1016/j.jcp.2009.08.025}{{Polynomial chaos
  representation of spatio-temporal random fields from experimental
  measurements}}, Journal of Computational Physics 228~(23) (2009) 8726--8751.
\newblock \href {https://doi.org/10.1016/j.jcp.2009.08.025}
  {\path{doi:10.1016/j.jcp.2009.08.025}}.
\newline\urlprefix\url{https://doi.org/10.1016/j.jcp.2009.08.025}

\bibitem{Des09}
C.~Desceliers, C.~Soize, Q.~Grimal, M.~Talmant, S.~Naili,
  \href{https://doi.org/10.1121/1.3087428}{{Determination of the random
  anisotropic elasticity layer using transient wave propagation in a
  fluid-solid multilayer: Model and experiments}}, The Journal of the
  Acoustical Society of America 125~(4) (2009) 2027--2034.
\newblock \href {http://arxiv.org/abs/https://doi.org/10.1121/1.3087428}
  {\path{arXiv:https://doi.org/10.1121/1.3087428}}, \href
  {https://doi.org/10.1121/1.3087428} {\path{doi:10.1121/1.3087428}}.
\newline\urlprefix\url{https://doi.org/10.1121/1.3087428}

\bibitem{Gui09}
J.~Guilleminot, C.~Soize, D.~Kondo,
  \href{https://doi.org/10.1016/j.mechmat.2009.08.004}{{Mesoscale probabilistic
  models for the elasticity tensor of fiber reinforced composites: Experimental
  identification and numerical aspects}}, Mechanics of Materials 41~(12) (2009)
  1309--1322.
\newblock \href {https://doi.org/10.1016/j.mechmat.2009.08.004}
  {\path{doi:10.1016/j.mechmat.2009.08.004}}.
\newline\urlprefix\url{https://doi.org/10.1016/j.mechmat.2009.08.004}

\bibitem{Ma09}
X.~Ma, N.~Zabaras, \href{http://stacks.iop.org/0266-5611/25/i=3/a=035013}{{An
  efficient Bayesian inference approach to inverse problems based on an
  adaptive sparse grid collocation method}}, Inverse Problems 25~(3) (2009)
  035013.
\newline\urlprefix\url{http://stacks.iop.org/0266-5611/25/i=3/a=035013}

\bibitem{Mar09a}
Y.~M. Marzouk, H.~N. Najm,
  \href{https://doi.org/10.1016/j.jcp.2008.11.024}{{Dimensionality reduction
  and polynomial chaos acceleration of Bayesian inference in inverse
  problems}}, Journal of Computational Physics 228~(6) (2009) 1862--1902.
\newblock \href {https://doi.org/10.1016/j.jcp.2008.11.024}
  {\path{doi:10.1016/j.jcp.2008.11.024}}.
\newline\urlprefix\url{https://doi.org/10.1016/j.jcp.2008.11.024}

\bibitem{Arn10}
M.~Arnst, R.~Ghanem, C.~Soize,
  \href{https://doi.org/10.1016/j.jcp.2009.12.033}{{Identification of Bayesian
  posteriors for coefficients of chaos expansions}}, Journal of Computational
  Physics 229~(9) (2010) 3134--3154.
\newblock \href {https://doi.org/10.1016/j.jcp.2009.12.033}
  {\path{doi:10.1016/j.jcp.2009.12.033}}.
\newline\urlprefix\url{https://doi.org/10.1016/j.jcp.2009.12.033}

\bibitem{Das10}
S.~Das, J.~C. Spall, R.~Ghanem,
  \href{https://doi.org/10.1016/j.csda.2009.09.018}{{Efficient Monte Carlo
  computation of Fisher information matrix using prior information}},
  Computational Statistics \& Data Analysis 54~(2) (2010) 272--289.
\newblock \href {https://doi.org/10.1016/j.csda.2009.09.018}
  {\path{doi:10.1016/j.csda.2009.09.018}}.
\newline\urlprefix\url{https://doi.org/10.1016/j.csda.2009.09.018}

\bibitem{Ta10}
Q.-A. Ta, D.~Clouteau, R.~Cottereau,
  \href{http://www.tandfonline.com/doi/abs/10.3166/ejcm.19.241-253}{{Modeling
  of random anisotropic elastic media and impact on wave propagation}},
  European Journal of Computational Mechanics 19~(1-3) (2010) 241--253.
\newblock \href
  {http://arxiv.org/abs/http://www.tandfonline.com/doi/pdf/10.3166/ejcm.19.241-253}
  {\path{arXiv:http://www.tandfonline.com/doi/pdf/10.3166/ejcm.19.241-253}},
  \href {https://doi.org/10.3166/ejcm.19.241-253}
  {\path{doi:10.3166/ejcm.19.241-253}}.
\newline\urlprefix\url{http://www.tandfonline.com/doi/abs/10.3166/ejcm.19.241-253}

\bibitem{Soi10}
C.~Soize, \href{https://doi.org/10.1016/j.cma.2010.03.013}{{Identification of
  high-dimension polynomial chaos expansions with random coefficients for
  non-Gaussian tensor-valued random fields using partial and limited
  experimental data}}, Computer Methods in Applied Mechanics and Engineering
  199~(33--36) (2010) 2150--2164.
\newblock \href {https://doi.org/10.1016/j.cma.2010.03.013}
  {\path{doi:10.1016/j.cma.2010.03.013}}.
\newline\urlprefix\url{https://doi.org/10.1016/j.cma.2010.03.013}

\bibitem{Soi11a}
C.~Soize, \href{https://doi.org/10.1016/j.cma.2011.07.005}{{A computational
  inverse method for identification of non-Gaussian random fields using the
  Bayesian approach in very high dimension}}, Computer Methods in Applied
  Mechanics and Engineering 200~(45--46) (2011) 3083--3099.
\newblock \href {https://doi.org/10.1016/j.cma.2011.07.005}
  {\path{doi:10.1016/j.cma.2011.07.005}}.
\newline\urlprefix\url{https://doi.org/10.1016/j.cma.2011.07.005}

\bibitem{Cot11}
R.~Cottereau, D.~Clouteau, H.~B. Dhia, C.~Zaccardi,
  \href{https://doi.org/10.1016/j.cma.2011.07.010}{{A stochastic-deterministic
  coupling method for continuum mechanics}}, Computer Methods in Applied
  Mechanics and Engineering 200~(47-48) (2011) 3280--3288.
\newblock \href {https://doi.org/10.1016/j.cma.2011.07.010}
  {\path{doi:10.1016/j.cma.2011.07.010}}.
\newline\urlprefix\url{https://doi.org/10.1016/j.cma.2011.07.010}

\bibitem{Des12}
C.~Desceliers, C.~Soize, S.~Naili, G.~Haiat,
  \href{https://doi.org/10.1016/j.ymssp.2012.03.008}{{Probabilistic model of
  the human cortical bone with mechanical alterations in ultrasonic range}},
  Mechanical Systems and Signal Processing 32 (2012) 170--177, uncertainties in
  Structural Dynamics.
\newblock \href {https://doi.org/10.1016/j.ymssp.2012.03.008}
  {\path{doi:10.1016/j.ymssp.2012.03.008}}.
\newline\urlprefix\url{https://doi.org/10.1016/j.ymssp.2012.03.008}

\bibitem{Per12}
G.~Perrin, C.~Soize, D.~Duhamel, C.~Funfschilling,
  \href{https://doi.org/10.1137/11084950X}{{Identification of Polynomial Chaos
  Representations in High Dimension from a Set of Realizations}}, SIAM Journal
  on Scientific Computing 34~(6) (2012) A2917--A2945.
\newblock \href {http://arxiv.org/abs/https://doi.org/10.1137/11084950X}
  {\path{arXiv:https://doi.org/10.1137/11084950X}}, \href
  {https://doi.org/10.1137/11084950X} {\path{doi:10.1137/11084950X}}.
\newline\urlprefix\url{https://doi.org/10.1137/11084950X}

\bibitem{Clo13}
D.~Clouteau, R.~Cottereau, G.~Lombaert,
  \href{https://doi.org/10.1016/j.jsv.2012.10.011}{{Dynamics of structures
  coupled with elastic media---A review of numerical models and methods}},
  Journal of Sound and Vibration 332~(10) (2013) 2415--2436.
\newblock \href {https://doi.org/10.1016/j.jsv.2012.10.011}
  {\path{doi:10.1016/j.jsv.2012.10.011}}.
\newline\urlprefix\url{https://doi.org/10.1016/j.jsv.2012.10.011}

\bibitem{Hay94}
S.~Haykin, {Neural Networks: A Comprehensive Foundation}, 1st Edition, Prentice
  Hall PTR, Upper Saddle River, NJ, USA, 1994.

\bibitem{Hag96}
M.~T. Hagan, H.~B. Demuth, M.~H. Beale, {Neural Network Design}, PWS Publishing
  Co., Boston, MA, USA, 1996.

\bibitem{Dem14}
H.~B. Demuth, M.~H. Beale, O.~De~Jess, M.~T. Hagan,
  \href{https://books.google.fr/books?id=4EW9oQEACAAJ}{{Neural Network
  Design}}, 2nd Edition, Martin Hagan, USA, 2014.
\newline\urlprefix\url{https://books.google.fr/books?id=4EW9oQEACAAJ}

\bibitem{Bow97}
A.~W. Bowman, A.~Azzalini, {Applied Smoothing Techniques for Data Analysis},
  Oxford University Press, Oxford, 1997.

\bibitem{Hor12}
I.~Horov{\'a}, J.~Kol{\'a}{\v c}ek, J.~Zelinka,
  \href{http://www.worldscientific.com/worldscibooks/10.1142/8468#t=aboutBook}{{Kernel
  Smoothing in MATLAB: Theory and Practice of Kernel Smoothing}}, World
  Scientific Publishing Co. Pte. Ltd., Singapore, 2012.
\newline\urlprefix\url{http://www.worldscientific.com/worldscibooks/10.1142/8468#t=aboutBook}

\bibitem{Giv13}
G.~H. Givens, J.~A. Hoeting, {Computational Statistics}, 2nd Edition, John
  Wiley \& Sons, Hoboken, New Jersey, 2013.

\bibitem{Sco15}
D.~W. Scott, \href{https://doi.org/10.1002/9781118575574}{{Multivariate Density
  Estimation: Theory, Practice, and Visualization}}, 2nd Edition, John Wiley \&
  Sons, Inc., 2015.
\newblock \href {https://doi.org/10.1002/9781118575574}
  {\path{doi:10.1002/9781118575574}}.
\newline\urlprefix\url{https://doi.org/10.1002/9781118575574}

\bibitem{Soi17a}
C.~Soize, \href{https://doi.org/10.1007/978-3-319-54339-0}{{Uncertainty
  Quantification: An Accelerated Course with Advanced Applications in
  Computational Engineering}}, 1st Edition, Vol.~47 of Interdisciplinary
  Applied Mathematics, Springer International Publishing, 2017.
\newblock \href {https://doi.org/10.1007/978-3-319-54339-0}
  {\path{doi:10.1007/978-3-319-54339-0}}.
\newline\urlprefix\url{https://doi.org/10.1007/978-3-319-54339-0}

\bibitem{Jay57a}
E.~T. Jaynes,
  \href{http://link.aps.org/doi/10.1103/PhysRev.106.620}{{Information Theory
  and Statistical Mechanics}}, Phys. Rev. 106 (1957) 620--630.
\newblock \href {https://doi.org/10.1103/PhysRev.106.620}
  {\path{doi:10.1103/PhysRev.106.620}}.
\newline\urlprefix\url{http://link.aps.org/doi/10.1103/PhysRev.106.620}

\bibitem{Jay57b}
E.~T. Jaynes,
  \href{http://link.aps.org/doi/10.1103/PhysRev.108.171}{{Information Theory
  and Statistical Mechanics. II}}, Phys. Rev. 108 (1957) 171--190.
\newblock \href {https://doi.org/10.1103/PhysRev.108.171}
  {\path{doi:10.1103/PhysRev.108.171}}.
\newline\urlprefix\url{http://link.aps.org/doi/10.1103/PhysRev.108.171}

\bibitem{Sob90}
K.~Sobezyk, J.~Tr{\c e}bicki,
  \href{https://doi.org/10.1016/0266-8920(90)90001-Z}{{Maximum entropy
  principle in stochastic dynamics}}, Probabilistic Engineering Mechanics 5~(3)
  (1990) 102--110.
\newblock \href {https://doi.org/10.1016/0266-8920(90)90001-Z}
  {\path{doi:10.1016/0266-8920(90)90001-Z}}.
\newline\urlprefix\url{https://doi.org/10.1016/0266-8920(90)90001-Z}

\bibitem{Kap92}
J.~N. Kapur, H.~K. Kesavan,
  \href{https://doi.org/10.1007/978-94-011-2430-0_1}{{Entropy Optimization
  Principles and Their Applications}}, Springer Netherlands, Dordrecht, 1992,
  pp. 3--20.
\newblock \href {https://doi.org/10.1007/978-94-011-2430-0_1}
  {\path{doi:10.1007/978-94-011-2430-0_1}}.
\newline\urlprefix\url{https://doi.org/10.1007/978-94-011-2430-0_1}

\bibitem{Jum00}
G.~Jumarie, \href{https://doi.org/10.1007/978-94-015-9496-7}{{Maximum Entropy,
  Information Without Probability and Complex Fractals: Classical and Quantum
  Approach}}, Vol. 112 of Fundamental Theories of Physics, Springer Science \&
  Business Media, Dordrecht, 2000.
\newblock \href {https://doi.org/10.1007/978-94-015-9496-7}
  {\path{doi:10.1007/978-94-015-9496-7}}.
\newline\urlprefix\url{https://doi.org/10.1007/978-94-015-9496-7}

\bibitem{Jay03}
E.~T. Jaynes, {Probability Theory: The Logic of Science}, Cambridge university
  press, 2003.

\bibitem{Cov06}
T.~M. Cover, J.~A. Thomas, {Elements of Information Theory}, A
  Wiley-Interscience publication, Wiley, New York, NY, USA, 2006.

\bibitem{Hug87}
T.~J.~R. Hughes, {The finite element method : linear static and dynamic finite
  element analysis}, Prentice Hall, Englewood Cliffs, New Jersey, 1987.

\bibitem{Zie05}
O.~C. Zienkiewicz, R.~L. Taylor, J.~Z. Zhu, {The Finite Element Method: Its
  Basis and Fundamentals}, Butterworth-Heinemann, 2005.

\bibitem{Zha19}
T.~Zhang, \href{https://tel.archives-ouvertes.fr/tel-02506242}{{Multiscale
  statistical inverse problem for the identification of random fields of
  elastic properties (in french)}}, phdthesis, {Universit{\'e} Paris-Est} (Dec.
  2019).
\newline\urlprefix\url{https://tel.archives-ouvertes.fr/tel-02506242}

\bibitem{Nem93}
S.~Nemat-Nasser, M.~Hori, {Micromechanics: Overall Properties of Heterogeneous
  Materials}, Vol.~37 of North-Holland Series in Applied Mathematics and
  Mechanics, North-Holland, Amsterdam, The Netherlands, 1993.

\bibitem{Bor01}
M.~Bornert, T.~Bretheau, P.~Gilormini, {Homog{\'e}n{\'e}isation en
  m{\'e}canique des mat{\'e}riaux 1. Mat{\'e}riaux al{\'e}atoires
  {\'e}lastiques et milieux p{\'e}riodiques}, Herm\`es Science publications,
  Paris, 2001.

\bibitem{Tor02}
S.~Torquato, \href{https://doi.org/10.1007/978-1-4757-6355-3}{{Random
  Heterogeneous Materials: Microstructure and Macroscopic Properties}},
  Vol.~16, Springer-Verlag, New York, NY, USA, 2002.
\newblock \href {https://doi.org/10.1007/978-1-4757-6355-3}
  {\path{doi:10.1007/978-1-4757-6355-3}}.
\newline\urlprefix\url{https://doi.org/10.1007/978-1-4757-6355-3}

\bibitem{Zao02}
A.~Zaoui,
  \href{https://doi.org/10.1061/(ASCE)0733-9399(2002)128:8(808)}{{Continuum
  Micromechanics: Survey}}, Journal of Engineering Mechanics 128~(8) (2002)
  808--816.
\newblock \href
  {http://arxiv.org/abs/https://ascelibrary.org/doi/pdf/10.1061/(ASCE)0733-9399(2002)128:8(808)}
  {\path{arXiv:https://ascelibrary.org/doi/pdf/10.1061/(ASCE)0733-9399(2002)128:8(808)}},
  \href {https://doi.org/10.1061/(ASCE)0733-9399(2002)128:8(808)}
  {\path{doi:10.1061/(ASCE)0733-9399(2002)128:8(808)}}.
\newline\urlprefix\url{https://doi.org/10.1061/(ASCE)0733-9399(2002)128:8(808)}

\bibitem{Bou04}
A.~Bourgeat, A.~Piatnitski,
  \href{https://doi.org/10.1016/j.anihpb.2003.07.003}{{Approximations of
  effective coefficients in stochastic homogenization}}, Annales de l'Institut
  Henri Poincare (B) Probability and Statistics 40~(2) (2004) 153--165.
\newblock \href {https://doi.org/10.1016/j.anihpb.2003.07.003}
  {\path{doi:10.1016/j.anihpb.2003.07.003}}.
\newline\urlprefix\url{https://doi.org/10.1016/j.anihpb.2003.07.003}

\bibitem{Ngu16}
M.-T. Nguyen, J.-M. Allain, H.~Gharbi, C.~Desceliers, C.~Soize,
  \href{https://doi.org/10.1016/j.jmbbm.2016.06.011}{{Experimental multiscale
  measurements for the mechanical identification of a cortical bone by digital
  image correlation}}, Journal of the Mechanical Behavior of Biomedical
  Materials 63 (2016) 125--133.
\newblock \href {https://doi.org/10.1016/j.jmbbm.2016.06.011}
  {\path{doi:10.1016/j.jmbbm.2016.06.011}}.
\newline\urlprefix\url{https://doi.org/10.1016/j.jmbbm.2016.06.011}

\bibitem{Shi71}
M.~Shinozuka, \href{https://doi.org/10.1121/1.1912338}{{Simulation of
  Multivariate and Multidimensional Random Processes}}, The Journal of the
  Acoustical Society of America 49~(1B) (1971) 357--368.
\newblock \href {http://arxiv.org/abs/https://doi.org/10.1121/1.1912338}
  {\path{arXiv:https://doi.org/10.1121/1.1912338}}, \href
  {https://doi.org/10.1121/1.1912338} {\path{doi:10.1121/1.1912338}}.
\newline\urlprefix\url{https://doi.org/10.1121/1.1912338}

\bibitem{Shi72a}
M.~Shinozuka, Y.~K. Wen, \href{https://doi.org/10.2514/3.50064}{{Monte Carlo
  Solution of Nonlinear Vibrations}}, AIAA Journal 10~(1) (1972) 37--40.
\newblock \href {https://doi.org/10.2514/3.50064} {\path{doi:10.2514/3.50064}}.
\newline\urlprefix\url{https://doi.org/10.2514/3.50064}

\bibitem{Shi72b}
M.~Shinozuka, C.-M. Jan,
  \href{https://doi.org/10.1016/0022-460X(72)90600-1}{{Digital simulation of
  random processes and its applications}}, Journal of Sound and Vibration
  25~(1) (1972) 111--128.
\newblock \href {https://doi.org/10.1016/0022-460X(72)90600-1}
  {\path{doi:10.1016/0022-460X(72)90600-1}}.
\newline\urlprefix\url{https://doi.org/10.1016/0022-460X(72)90600-1}

\bibitem{Poi89}
F.~Poirion, C.~Soize,
  \href{https://hal-upec-upem.archives-ouvertes.fr/hal-00770316}{{Numerical
  simulation of homogeneous and inhomogeneous Gaussian stochastic vector
  fields}}, {La Recherche Aerospatiale (English edition)} 1~(-) (1989) 41--61.
\newline\urlprefix\url{https://hal-upec-upem.archives-ouvertes.fr/hal-00770316}

\bibitem{Poi95}
F.~Poirion, C.~Soize,
  \href{https://doi.org/10.1007/3-540-60214-3_50}{{Numerical methods and
  mathematical aspects for simulation of homogeneous and non homogeneous
  gaussian vector fields}}, in: P.~Kr{\'e}e, W.~Wedig (Eds.), Probabilistic
  Methods in Applied Physics, Springer Berlin Heidelberg, Berlin, Heidelberg,
  1995, pp. 17--53.
\newblock \href {https://doi.org/10.1007/3-540-60214-3_50}
  {\path{doi:10.1007/3-540-60214-3_50}}.
\newline\urlprefix\url{https://doi.org/10.1007/3-540-60214-3_50}

\bibitem{Sil86}
B.~W. Silverman, {Density Estimation for Statistics and Data Analysis}, Chapman
  and Hall, London, 1986.

\bibitem{Rob04}
C.~Robert, G.~Casella,
  \href{https://dx.doi.org/10.1007/978-1-4757-4145-2}{{Monte Carlo Statistical
  Methods}}, 2nd Edition, Springer Texts in Statistics, Springer-Verlag, New
  York, NY, USA, 2004.
\newblock \href {https://doi.org/10.1007/978-1-4757-4145-2}
  {\path{doi:10.1007/978-1-4757-4145-2}}.
\newline\urlprefix\url{https://dx.doi.org/10.1007/978-1-4757-4145-2}

\bibitem{Kai05}
J.~Kaipio, E.~Somersalo, \href{https://doi.org/10.1007/b138659}{{Statistical
  and Computational Inverse Problems}}, 1st Edition, Vol. 160 of Applied
  Mathematical Sciences, Springer-Verlag, New York, NY, USA, 2005.
\newblock \href {https://doi.org/10.1007/b138659} {\path{doi:10.1007/b138659}}.
\newline\urlprefix\url{https://doi.org/10.1007/b138659}

\bibitem{Spa05a}
J.~C. Spall, \href{https://dx.doi.rg/10.1002/0471722138}{{Introduction to
  Stochastic Search and Optimization: Estimation, Simulation, and Control}},
  Vol.~65, John Wiley \& Sons, 2005.
\newblock \href {https://doi.org/10.1002/0471722138}
  {\path{doi:10.1002/0471722138}}.
\newline\urlprefix\url{https://dx.doi.rg/10.1002/0471722138}

\bibitem{Cyb89}
G.~Cybenko, \href{https://doi.org/10.1007/BF02551274}{{Approximation by
  superpositions of a sigmoidal function}}, Mathematics of Control, Signals and
  Systems 2~(4) (1989) 303--314.
\newblock \href {https://doi.org/10.1007/BF02551274}
  {\path{doi:10.1007/BF02551274}}.
\newline\urlprefix\url{https://doi.org/10.1007/BF02551274}

\bibitem{Hor89}
K.~Hornik, M.~Stinchcombe, H.~White,
  \href{https://doi.org/10.1016/0893-6080(89)90020-8}{{Multilayer feedforward
  networks are universal approximators}}, Neural Networks 2~(5) (1989)
  359--366.
\newblock \href {https://doi.org/10.1016/0893-6080(89)90020-8}
  {\path{doi:10.1016/0893-6080(89)90020-8}}.
\newline\urlprefix\url{https://doi.org/10.1016/0893-6080(89)90020-8}

\bibitem{Hor91}
K.~Hornik, \href{https://doi.org/10.1016/0893-6080(91)90009-T}{{Approximation
  capabilities of multilayer feedforward networks}}, Neural Networks 4~(2)
  (1991) 251--257.
\newblock \href {https://doi.org/10.1016/0893-6080(91)90009-T}
  {\path{doi:10.1016/0893-6080(91)90009-T}}.
\newline\urlprefix\url{https://doi.org/10.1016/0893-6080(91)90009-T}

\bibitem{Les93}
M.~Leshno, V.~Y. Lin, A.~Pinkus, S.~Schocken,
  \href{https://doi.org/10.1016/S0893-6080(05)80131-5}{{Multilayer feedforward
  networks with a nonpolynomial activation function can approximate any
  function}}, Neural Networks 6~(6) (1993) 861--867.
\newblock \href {https://doi.org/10.1016/S0893-6080(05)80131-5}
  {\path{doi:10.1016/S0893-6080(05)80131-5}}.
\newline\urlprefix\url{https://doi.org/10.1016/S0893-6080(05)80131-5}

\bibitem{Bar93}
A.~R. {Barron}, \href{https://doi.org/10.1109/18.256500}{{Universal
  approximation bounds for superpositions of a sigmoidal function}}, IEEE
  Transactions on Information Theory 39~(3) (1993) 930--945.
\newblock \href {https://doi.org/10.1109/18.256500}
  {\path{doi:10.1109/18.256500}}.
\newline\urlprefix\url{https://doi.org/10.1109/18.256500}

\bibitem{LeCun15}
Y.~LeCun, Y.~Bengio, G.~Hinton,
  \href{https://doi.org/10.1038/nature14539}{{Deep learning}}, Nature
  521~(7553) (2015) 436--444.
\newblock \href {https://doi.org/10.1038/nature14539}
  {\path{doi:10.1038/nature14539}}.
\newline\urlprefix\url{https://doi.org/10.1038/nature14539}

\bibitem{Goo16}
I.~Goodfellow, Y.~Bengio, A.~Courville, {Deep Learning}, The MIT Press,
  Cambridge, MA, USA, 2016, \url{http://www.deeplearningbook.org}.

\bibitem{Vogl88}
T.~P. Vogl, J.~K. Mangis, A.~K. Rigler, W.~T. Zink, D.~L. Alkon,
  \href{https://doi.org/10.1007/BF00332914}{{Accelerating the convergence of
  the back-propagation method}}, Biological Cybernetics 59~(4) (1988) 257--263.
\newblock \href {https://doi.org/10.1007/BF00332914}
  {\path{doi:10.1007/BF00332914}}.
\newline\urlprefix\url{https://doi.org/10.1007/BF00332914}

\bibitem{Ngu90}
D.~{Nguyen}, B.~{Widrow},
  \href{https://doi.org/10.1109/IJCNN.1990.137819}{{Improving the learning
  speed of 2-layer neural networks by choosing initial values of the adaptive
  weights}}, in: 1990 IJCNN International Joint Conference on Neural Networks,
  Vol.~3, 1990, pp. 21--26.
\newblock \href {https://doi.org/10.1109/IJCNN.1990.137819}
  {\path{doi:10.1109/IJCNN.1990.137819}}.
\newline\urlprefix\url{https://doi.org/10.1109/IJCNN.1990.137819}

\bibitem{Bea92}
M.~H. Beale, M.~T. Hagan, H.~B. Demuth, Neural network toolbox user's guide,
  The MathWorks Inc (1992).

\bibitem{Soi05a}
C.~Soize, \href{https://doi.org/10.1016/j.cma.2004.06.038}{{Random matrix
  theory for modeling uncertainties in computational mechanics}}, Computer
  Methods in Applied Mechanics and Engineering 194~(12--16) (2005) 1333--1366,
  special Issue on Computational Methods in Stochastic Mechanics and
  Reliability Analysis.
\newblock \href {https://doi.org/10.1016/j.cma.2004.06.038}
  {\path{doi:10.1016/j.cma.2004.06.038}}.
\newline\urlprefix\url{https://doi.org/10.1016/j.cma.2004.06.038}

\bibitem{Soi16a}
C.~Soize, \href{https://doi.org/10.1007/978-3-319-11259-6_5-1}{{Random Matrix
  Models and Nonparametric Method for Uncertainty Quantification}}, Springer
  International Publishing, Cham, 2016, pp. 1--69.
\newblock \href {https://doi.org/10.1007/978-3-319-11259-6_5-1}
  {\path{doi:10.1007/978-3-319-11259-6_5-1}}.
\newline\urlprefix\url{https://doi.org/10.1007/978-3-319-11259-6_5-1}

\bibitem{Law95}
C.~Lawson, R.~Hanson, \href{https://doi.org/10.1137/1.9781611971217}{{Solving
  Least Squares Problems}}, Society for Industrial and Applied Mathematics,
  1995.
\newblock \href
  {http://arxiv.org/abs/https://epubs.siam.org/doi/pdf/10.1137/1.9781611971217}
  {\path{arXiv:https://epubs.siam.org/doi/pdf/10.1137/1.9781611971217}}, \href
  {https://doi.org/10.1137/1.9781611971217}
  {\path{doi:10.1137/1.9781611971217}}.
\newline\urlprefix\url{https://doi.org/10.1137/1.9781611971217}

\bibitem{Ser80}
R.~Serfling, \href{https://dx.doi.org/10.1002/9780470316481}{{Approximation
  Theorems of Mathematical Statistics}}, Wiley, New York, NY, USA, 1980.
\newblock \href {https://doi.org/10.1002/9780470316481}
  {\path{doi:10.1002/9780470316481}}.
\newline\urlprefix\url{https://dx.doi.org/10.1002/9780470316481}

\bibitem{Pap02}
A.~Papoulis, S.~U. Pillai, {Probability, Random Variables, and Stochastic
  Processes}, 4th Edition, McGraw-Hill Higher Education, New York, NY, USA,
  2002.

\end{thebibliography}

\end{document}